\documentclass{article}

    \usepackage[preprint]{neurips_2025}

\usepackage[utf8]{inputenc} 
\usepackage[T1]{fontenc}    
\usepackage{hyperref}       
\usepackage{url}            
\usepackage{booktabs}       
\usepackage{amsfonts}       
\usepackage{nicefrac}       
\usepackage{microtype}      
\usepackage{xcolor}         

\usepackage{graphicx}
\usepackage{amsmath}
\usepackage{cleveref}
\usepackage{wrapfig}
\usepackage{microtype}
\usepackage{tabularx}
\usepackage{array}
\usepackage[table]{xcolor}
\setcitestyle{numbers}
\usepackage{floatrow}
\usepackage{bbding}
\newfloatcommand{capbtabbox}{table}[][\FBwidth]

\def\eg{\emph{e.g.}} 
\def\ie{\emph{i.e.}} 
 
\def\etc{\emph{etc.}}

\def\etal{\emph{et al.}}

\definecolor{red}{RGB}{230,10,30}

\definecolor{taskrfortyfive}{RGB}{50,50,93}

\definecolor{taskrninety}{RGB}{87,86,100}

\title{Probing Intrinsic Medical Task Relationships: \\ A Contrastive Learning Perspective}

\author{%
  Jonas Muth \\
  \And
  Zdravko Marinov \\
  \And
  Simon Rei{\ss}~\Envelope \\
  \AND
  \\
  Karlsruhe Institute of Technology \\
  Karlsruhe, Germany\\
  \Envelope~\texttt{simon.reiss@kit.edu} \\
}

\begin{document}

\maketitle

\begin{abstract}

\noindent While much of the medical computer vision community has focused on advancing performance for specific tasks, the underlying relationships between tasks, \ie, how they relate, overlap, or differ on a representational level, remain largely unexplored.
Our work explores these intrinsic relationships between medical vision tasks, specifically, we investigate 30 tasks, such as semantic tasks (\eg, segmentation and detection), image generative tasks (\eg, denoising, inpainting, or colorization), and image transformation tasks (\eg, geometric transformations).
Our goal is to probe whether a data-driven representation space can capture an underlying structure of tasks across a variety of 39 datasets from wildly different medical imaging modalities, including computed tomography, magnetic resonance, electron microscopy, X-ray ultrasound and more. By revealing how tasks relate to one another, we aim to provide insights into their fundamental properties and interconnectedness. To this end, we introduce Task-Contrastive Learning (TaCo), a contrastive learning framework designed to embed tasks into a shared representation space. Through TaCo, we map these heterogeneous tasks from different modalities into a joint space and analyze their properties: identifying which tasks are distinctly represented, which blend together, and how iterative alterations to tasks are reflected in the embedding space.
Our work provides a foundation for understanding the intrinsic structure of medical vision tasks, offering a deeper understanding of task similarities and their interconnected properties in embedding spaces.

\end{abstract}
\section{Introduction}
\label{sec:intro}

In recent years, long-standing computer vision tasks have been very successfully addressed through deep neural networks.
This in turn -- driven by their impressive results -- has reinforced the collection and annotation of new datasets in new domains for new tasks to be addressed through deep learning.
In the bygone decade, the tasks were addressed individually by training new neural networks again and again.
These specialized models could shine with strong performance, but such individual solutions entail that for each new task, the training pipeline has to be followed: data collection, data annotation and model training.

\begin{figure}
  \begin{center}
    \includegraphics[width=0.75\textwidth]{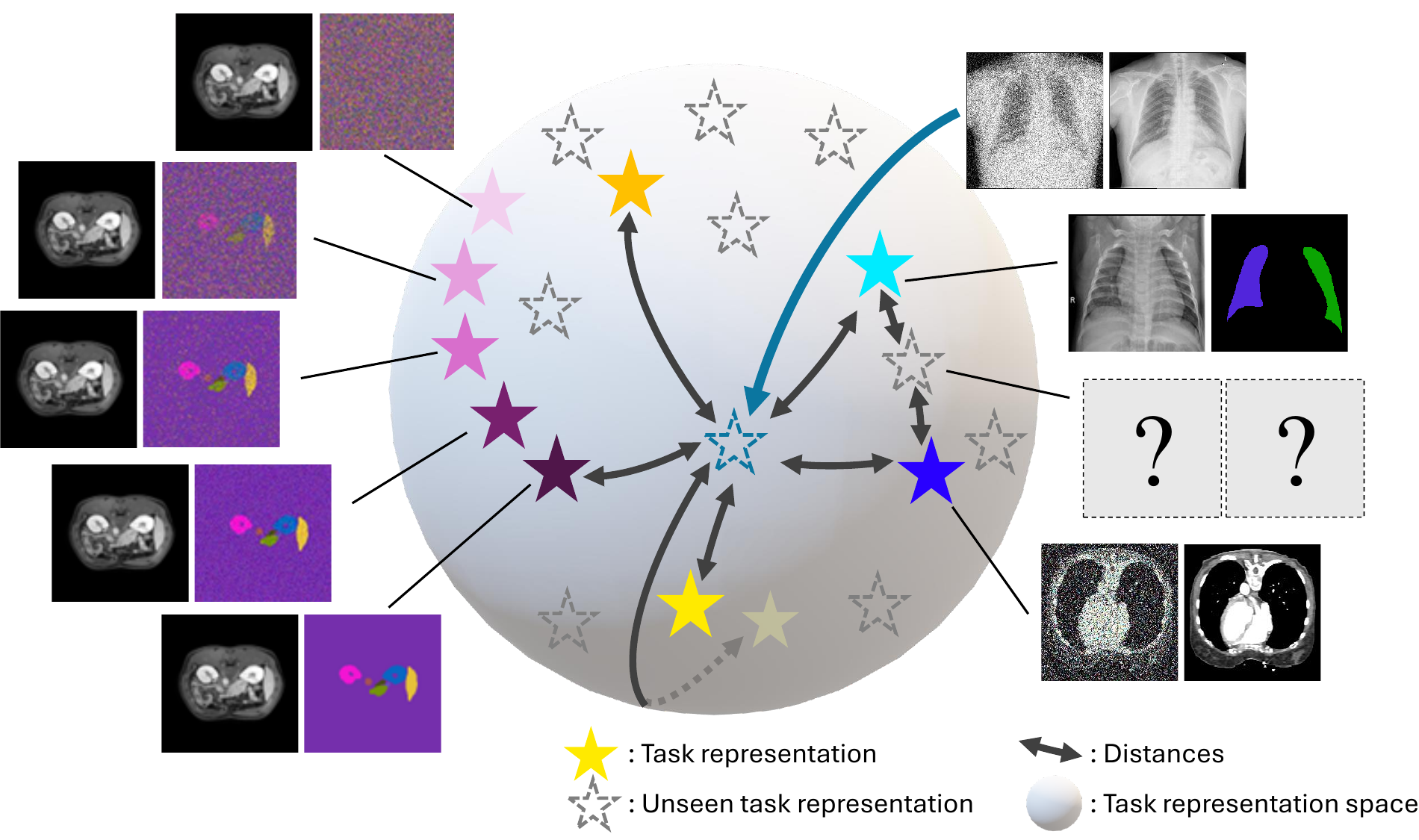}
  \end{center}
  \caption{We aim to train a model, which is able to embed medical computer vision tasks into a joint representation space. This space allows us to explore how tasks relate to one another: for example, we can analyze the proximity of unseen tasks to known tasks (right) or study how iterative task alterations (\eg, increasing noise) are reflected in the learned embedding space (left).
    }
    \label{fig:intro_figure}
\end{figure}

More recently, research on methods that are designed to adapt to new contexts and new imaging domains, as well as methods which directly address a large amount of tasks at once are being sought after~\cite{bar2022visual,lu2022unified,czolbe2023neuralizer,wang2023images,wang2023seggpt,zou2024segment,bai2024sequential,guo2024data,gu2024analogist}.
These works ask questions such as: How can a unified model be trained for many tasks? How can models adapt to new tasks?
We aim to ask the question behind these questions: How can tasks be represented and how do tasks relate to each other within these representations?
While the question of relations between tasks has been asked before for tasks in the natural image domain~\cite{zamir2018taskonomy}, we are interested in investigating it directly from the network perspective in the medical domain, \ie, how can task representations be learned with a neural network, and how does this learned task representation space relate different tasks to each other?
In~\Cref{fig:intro_figure}, we visualize this idea of a task embedding space for medical images.

How can we learn a data-driven task representation?
Learning representations automatically from image data is often addressed by self-supervised training strategies, \eg, networks are trained with masked auto-encoding~\cite{he2022masked}, rotation prediction~\cite{gidaris2018unsupervised}, solving image jigsaw puzzles~\cite{noroozi2016unsupervised}, or colorization~\cite{zhang2016colorful}.
A different pathway to learn representations of data is contrastive learning~\cite{chopra2005learning,oord2018representation,chen2020simple,chen2020improved,yeh2022decoupled}, which defines pairs of samples which are depict similar image content as positives, and pairings where image content differs as negatives.
In representation learning for images, these positives are most commonly defined by augmenting the same image twice and viewing this pairing as positive, as it contains matching image patterns.

In this work, we utilize contrastive learning as a mechanism to train a network towards embedding input-image and task-output pairs into an embedding space.
To our knowledge, we are the first to propose such a data-driven pathway to learn a task representation space that captures the intrinsic relationships between tasks.
In order to evaluate whether the learned task representation space captures relational properties of different tasks, we extensively evaluate the learned representation space against pretrained model architectures, and showcase how our~\emph{Task-Contrastive Learning (TaCo)} strategy captures task properties much more conveniently.
To this end, we evaluate nearest neighbor classification accuracy to distinguish between medical computer vision tasks.
On top, we explore learned distances between different tasks when embedding them into this joint space.
There, we observe that our proposed \emph{TaCo} shines in identifying similarities and differences among tasks, while baseline methods encode tasks heavily on an individual image instance basis.

\noindent Our contributions summarize to:
\begin{itemize}
    \item We introduce \emph{TaCo}, a data-driven framework for learning task representations, and apply it to explore the intrinsic relationships among 30 tasks across 39 diverse medical imaging datasets.
    \item We analyze task similarities within the learned representation space using nearest neighbor classification and study how iterative task alterations are reflected in the embedding space, revealing distinct patterns and relationships captured by \emph{TaCo} representations, but missed by baseline methods.
    \item We provide novel insights into intrinsic task relationships through analyses of the task embedding space, including dimensionality reduction and visualizations of task similarities and clusters.
\end{itemize}
\section{Related work}
\label{sec:rel_work}
\subsection{Task Taxonomy and Task Representations}
Our work focuses on creating vector representations for medical visual tasks.
A related approach that explores representations for visual tasks in the natural image domain is Task2Vec~\cite{achille2019task2vec}, which encodes classification tasks using Fisher Information Matrices and gradients of a pre-trained network with respect to task-specific losses.
They further show, that capturing similarities among task representations can be utilized to draw conclusions about the transferability of feature extractors between tasks.
In contrast, we aim at finding a pathway towards learning task representations in a data-driven fashion, and further, we focus on more complex visual tasks such as semantic segmentation, detection, denoising, \etc

Exploring the relationships between higher-level visual tasks is done in Taskonomy~\cite{zamir2018taskonomy}.
There, task-specific models of similar architecture are trained on these tasks and successively, the transfer of their representations towards other tasks is quantified.
This investigation led to a taxonomy of tasks based on the transfer efficacy.
Our approach differs in that we do not train task-specific networks but we ask the question, whether a single network can embed all tasks into a joint representation space: we consider both image-input and task-output as input to our network. 

Hojel \etal~\cite{hojel2025finding}, examine a visual in-context learning model, which was trained on multiple vision tasks.
In the activations of this model they search whether they relate to specific vision tasks.
Indeed, they are able to extract a conjunction of activations, which can serve as a task vector, yet, their exploration is limited to four tasks from the natural image domain.
In our work, we have the goal to learn task representations directly without a search and we consider a larger number of medical vision tasks.



\subsection{Image Representation Learning}

In recent years, learning representations from unlabeled images has been flourishing with works exploring it through defining proxy tasks~\cite{zhang2016colorful,gidaris2018unsupervised}, \eg, rotating an image and predicting the rotation with a neural network or converting an image to gray-scale and predicting the original colors.
This relates to our work as we also use such tasks, although for a different purpose: we utilize them to increase the number of tasks we have access to for training a neural network for representing task information. 
More such tasks, where the supervision comes from the image itself include  denoising~\cite{brempong2022denoising}, generative modeling~\cite{donahue2016adversarial}, inpainting~\cite{pathak2016context} or jigsaw puzzles~\cite{noroozi2016unsupervised}.

A different pathway to learn representations is contrastive training~\cite{oord2018representation}.
There, images are identified, which should lead to similar representations, so called positive pairs.
To obtain such positives, images can be paired with an augmented version of themselves~\cite{chen2020simple} as their semantic content will match.
In our work, we build upon supervised contrastive learning~\cite{khosla2020supervised}, which enables the utilization of a connecting factor, \eg, a label, to define positives and negatives.
Specifically, we utilize it to learn task representations by defining positives based on whether task instances encode the same task.
\section{Methods}
\label{sec:methods}
Next, we define our notation of visual tasks and networks, then we introduce our solution to task-representation learning: \emph{Task-Contrastive learning (TaCo)}.
We couple \emph{TaCo} successively with the idea of~\emph{failure tasks} for capturing distances between visual tasks in a more nuanced fashion.

\subsection{Notation and Setting}
\label{sec:setting}
We aim at learning a model $\theta(\cdot)$ which can process visual tasks $\{\mathcal{T}_1, \dots, \mathcal{T}_n\}$, more precisely instances $t \in \mathcal{T}_i$ from these visual tasks, where $t$ is a tuple of a visual input $in$ and a task output $out$, which are both encoded as images $in, out \in \mathbb{R}^{3 \times W \times H}$.
While classical task-specific models would aim at learning the mapping function from $in$ to $out$, we aim at jointly embedding them into a $D$-dimensional representation space through $\theta(t) \in \mathbb{R}^D$.
Note, that we view visual tasks on different datasets as individual tasks $\mathcal{T}_i$, \eg, when X-ray images serve as visual input and detected lungs as task output, a different visual task would be to detect lungs in computed tomography images. \\

To investigate whether the learned task embedding function $\theta(\cdot)$ produces suitable representations in $ \mathbb{R}^D$ of visual tasks, we explore the following properties:

(1) Seen visual task nearest neighbor classification: Can we predict for a new task instance $t^{new}$, based on nearest neighbors in the learned task representation space, which seen visual task $\mathcal{T}_i$ it is associated to?

(2) Unseen visual task nearest neighbor classification: Can we predict for a new task instance $t^{new}$, based on nearest neighbors in the learned task representation space, which unseen visual task $\mathcal{T}_i^{unseen}$ it is associated to? Here, $\mathcal{T}_i^{unseen}$ refers to a visual task not present in the training.

(3) Iterative task adaptation: How do the distances between the mean representation of visual tasks $\mathcal{T}_i$ and $\mathcal{T}_j$ behave, when iteratively altering one of them? \\


\noindent Representation spaces can also be investigated qualitatively by their visualization through dimensionality reduction techniques, which is our final exploration.


\subsection{Task Representation Learning}
Next, we introduce a way to train a task representation model $\theta(\cdot)$, parameterized by a ResNet~\cite{he2016deep}, with the change that we concatenate the input $in$ and output $out$ of a visual task $t$ and input it as $6$ channel image.
Similar to previous work on contrastive learning~\cite{chen2020simple}, we use a multi-layer perceptron $\phi(\cdot)$ projection head on top of the last layer.

\subsubsection{Task-Contrastive Learning}
In image-based self-supervised learning the contrastive loss formulation, which is based on positives and negatives, \eg, in SimCLR~\cite{chen2020simple}, is well explored:
\begin{equation}
        L^{self} = - \sum_{i \in I} \log \frac{\text{exp}(\langle z_i, \hat{z}_i \rangle / \tau)}{\sum_{a \in I \backslash \{i\}} \text{exp}(\langle z_i, z_a \rangle / \tau)}\enspace,
        \label{eq:contrastive}
\end{equation}

where $z_l$ is a high dimensional representation of the input data when processed by $\phi(\theta(\cdot))$, $\hat{z}_l$ indicates a representation of an augmented version of the input.
$I$ indicates all instances in a batch, including their augmented versions, $\tau \in \mathbb{R}^+$ is a temperature scaling parameter.

Out-of-the box, this contrastive formulation can not be used to integrate information which conceptually connects multiple instances, which we would like to do for multiple instances from the same visual task $\mathcal{T}_i$.
To connect instances that share a property, \eg, the class, supervised contrastive learning~\cite{khosla2020supervised} was proposed:
\begin{equation}
    L^{sup} = \sum_{i \in I} \frac{-1}{|P_i|} \sum_{p\in P_i} \log \frac{\text{exp}(\langle z_i, z_p \rangle / \tau)}{\sum\limits_{a \in I \backslash \{i\}} \text{exp}(\langle z_i, z_a\rangle / \tau)}\enspace,
    \label{eq:supervised_contrast}
\end{equation}
where $P_i$ is the set of all indices in $I$ that share the class with sample $i$ at indices other than $i$.

As our task representation learner $\theta(\cdot)$ embeds visual task instances into a representation space, we can use~\Cref{eq:supervised_contrast} such that representations of instances $t_j$, $t_k$ drawn from the same task $\mathcal{T}_i$ serve as positives.
For a task instance $t_j \in \mathcal{T}_i$, the set $P_j$ therefore contains all task instances $t_k, k \neq i$ in the batch, that also originate from $\mathcal{T}_i$.

To make sure that positive task-pairings are present in each batch, \emph{data augmentation} is replaced by a sampling mechanism, where a batch is \emph{task augmented} by sampling a matching task instance $t_k$ which stems from the same visual task $\mathcal{T}_i$ for each task instance $t_j$, which doubles the batch size.

As the number of task-instances $|\mathcal{T}_i|$ in different visual tasks may vary greatly, we introduce a \emph{balanced task sampling}, where we under-sample visual tasks with a high amount of task instances and over-sample visual tasks with a lower amount of task instances.
This is important for the learned representations to not only capture tasks with a high number of instances.
For this, each instance $t_j \in \mathcal{T}_i$ is assigned a success probability to be drawn of $1 / (|\mathcal{T}_i| \cdot \sum^n_{k=1} |\mathcal{T}_k|)$ in a multinomial distribution at each iteration.

\begin{wrapfigure}{r}{0.5\textwidth}
  \begin{center}
    \includegraphics[width=\textwidth,page=2]{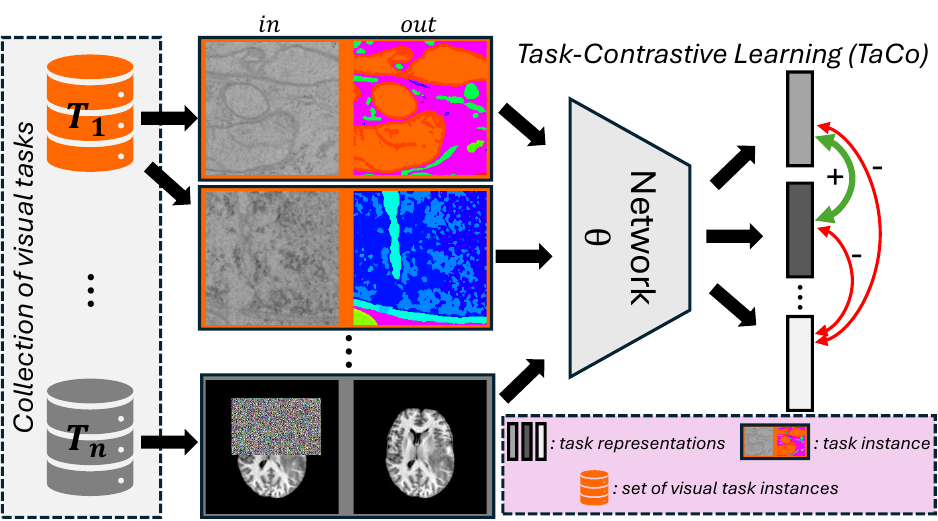}
  \end{center}
  \caption{Overview of our \emph{Task-Contrastive Learning}. We sample input-output image pairs from visual tasks and embed them with a neural network into vectors, which is tuned with a contrastive loss.}
\label{fig:overview_figure}
\end{wrapfigure}

With these task-based changes, minimizing the loss in~\Cref{eq:supervised_contrast} leads to representations of instances from the same task getting more and more aligned to each other, while representations of different tasks are pushed apart.
We term our data-driven task representation learning strategy \emph{\textbf{Ta}sk-\textbf{Co}ntrastive Learning (TaCo)} and display it in~\Cref{fig:overview_figure}.

\subsubsection{Task-Contrastive Failure Tasks}
In practice, the number $n$ of tasks $\{\mathcal{T}_0, \dots, \mathcal{T}_n\}$ may be a limiting factor to learn distinctive task representations.
Further, a network might resort towards only capturing the most distinctive properties of the input data, \ie, the dataset which the image $in$ stems from or the properties of the task-output $out$, but not their correct relation to each other.
To explore these factors and their relation to $\theta(\cdot)$ merely learning obvious task biases, we propose a mechanism to integrate inconsistent pairings, which we call \emph{failure tasks}.

Such \emph{failure tasks} are gathered by a mixing scheme, where we mix the task-input $in_j$ and the task output $out_j$ from a task instance $t_j \in \mathcal{T}_k$ with the task instance $t_l$ of the same task $\mathcal{T}_k$.
More specifically, we setup the failure task instances $\hat{t}_{j,l} = (in_j, out_l)$ and $\hat{t}_{l, j} = (in_l, out_j)$ which mix both tasks.
We group these \emph{failure tasks} into $\hat{\mathcal{T}}$ and treat it as the $n+1^{th}$ visual task.
This leads to instances, where the task input and output are mismatched, but the general bias in the data is still present.
As such, the bias alone can not be leveraged to disambiguate $t_j$ from $\hat{t}_{j,l}$, but the consistency between $in_j$ and $out_j$ needs to be captured.


\section{Data, Baselines and Inference}

\subsection{Datasets}
\label{sec:datasets}
As we require a high number of tasks to train a task-representation learner, we train our method on $35$ medical datasets, which we term as seen datasets and we hold out four additional unseen datasets for testing the generalization capability to new data.
The seen datasets span over the medical imaging modalities of microscopy, electron microscopy, magnetic resonance, PET, CT, endoscopy, optical coherence tomography, X-ray, dermoscopy and ultrasound.
The unseen datasets cover magnetic resonance, X-ray, dermoscopy and ultrasound.
The datasets are sourced from the MedSAM dataset collection~\cite{ma2024segment}, where we use a subset of the provided images and add the Openorganelle dataset family~\cite{heinrich2021whole}.
In total, we have $291,165$ images with pixel-wise segmentation labels.
We split each dataset into $70\%$ training, $15\%$ validation and $15\%$ testing data.
For visual examples, please refer to the supplemental material.


\subsection{Visual Tasks}
\label{sec:visualtasks}
Consistent to the datasets, we also define seen tasks which are used in training and in evaluation.
Unseen tasks are only used in evaluation.
To this end, we define $30$ tasks in total, which can be coarsely categorized as \emph{semantic tasks}, \emph{image generative tasks} and \emph{image transformation tasks}.

(1) \textbf{Semantic Tasks}: These tasks involve mapping an image to semantic structures. We derive all semantic tasks from pixel-wise segmentation labels via pre-processing, for example by computing bounding boxes, edges of segments, medial axes, and more. This leads to tasks such as semantic segmentation, bounding box detection, edge detection, \etc

(2) \textbf{Image Generative Tasks}: For these tasks, the input is an image which is manipulated to miss pixel-values or has some information loss which needs to be reconstructed. The task output is the original image. Examples of these tasks are inpainting, denoising, colorization, super-resolution, \etc

(3) \textbf{Image Transformation Tasks}: Here, the image data serves as input $in$ and the output $out$ is defined by passing $in$ through a transformation $\varphi(\cdot)$, such that the visual tasks are $(in, \varphi(in))$. The transformation $\varphi(\cdot)$ can be a rotation, an inversion of pixel values, a horizontal image flip, \etc

With $30$ tasks and $39$ datasets, and by taking their combination, we have access to $1,134$ visual tasks (colorization not defined for datasets containing grayscale images, thus, slightly below $30 \cdot 39 = 1.170$) with a total of $8,450,298$ task instances based on the $291,165$ images from~\Cref{sec:datasets}.
Find example images for all visual tasks in the supplemental material.



\subsection{Implementation Details}
\label{sec:impldetails}
The network $\theta(\cdot)$ is a ResNet-50~\cite{he2016deep} with a multi-layer perceptron as projector $\phi(\cdot)$ (two layers: $2048$ neurons, ReLU, $128$ output neurons), after training this projector is omitted.
The ResNet is altered to take $6$ channels as input, we train it with a learning rate of $0.001$, SGD with momentum of $0.9$, weight decay of $0.0001$, we set $\tau=0.7$ and use a batch size of $40$.
When using \emph{failure tasks}, we train with $30$ visual task instances and $10$ failure task instances, which are added after $50,000$ iterations.
We train on two NVIDIA 2080 Ti GPUs and $6 \times 200 \times 200$ inputs.

\subsection{Baseline Methods}
To the best of our knowledge, this is the first work to explore medical task representation learning from a data-driven perspective and thus no prior works exist to compare \emph{TaCo} to.
Still, to verify its effectiveness, we construct baselines from two pre-trained models and one model trained on our data.

Our baselines work in the following fashion: We utilize each baseline model $\theta_b$ to individually extract a feature vector for both the input $in$ and output $out$ with $\theta_b(in)$ and $\theta_b(out)$, successively we embed the visual task instance as their concatenation $concat(\theta_b(in), \theta_b(out))$.

\textbf{CLIP}~\cite{Radford2021LearningTV,ilharco_gabriel_2021_5143773,cherti2023reproducible}:
This model was pre-trained on the large scale dataset LAION-5B~\cite{schuhmann2022laionb} on contrastive image-text alignment.
The CLIP embeddings have been shown to be robust towards distribution shifts in out-of-domain settings~\cite{wortsman2022robust} and thereby serve as a strong baseline for our setting.
We utilize the visual encoder, a ViT-B-32~\cite{dosovitskiy2020image}, as embedding function $\theta_b$, which yields a $512$ dimensional vector.

\textbf{ImageNet}~\cite{he2016deep}:
The second baseline is the tried and tested ResNet-50 architecture, pre-trained for image recognition on the ImageNet dataset~\cite{deng2009imagenet} to recognize $1,000$ object categories.
We utilize the $2,048$ dimensional features in the penultimate layer of the network.

\textbf{SimCLR (in-domain trained)}~\cite{chen2020simple}:
To verify, that our training strategy is well suited to learn a task representation, we compare it to the contrastive training strategy SimCLR.
This method is trained in-domain on the visual task instances by individually viewing the constituents $in$ and $out$ as images, augmenting them and successive training with \Cref{eq:contrastive}.
We use the same hyperparameters as in \emph{TaCo}.

\subsection{Visual Task Inference}
To quantitatively evaluate how well the different embeddings are suited to represent tasks, we evaluate the three settings outlined in~\Cref{sec:setting}.
In order to infer a visual task $\mathcal{T}_i$ for a task instance based on a model $\theta$ with k-nearest neighbor classification (kNN), we first embed the task instance with the model and then search the closest representation in a reference set of embedded task instances for which we have access to the visual task information.
We then retrieve the vision task of the closest reference instance.
We use the cosine distance to compute nearest neighbors.

\section{Evaluation}
\label{sec:evaluation}
In this section, we first show the results of the nearest neighbor classification to categorize new visual task instances into \emph{seen visual tasks ($805)$}, \emph{unseen visual tasks} ($25$), or the combination of both which we denote as \emph{all visual tasks} ($1,134$).
Then, we show a nearest neighbor classification evaluation, where we classify the visual task instances into the $30$ base tasks and the $39$ datasets to grasp where the differences of the method's performance come from.
An ablation study further shows the importance of \emph{balanced task sampling}.
Afterwards, we investigate the progression of distances between two vision tasks in representation space when changing one of them iteratively.
We visualize the task representation space of our method \emph{TaCo} and the CLIP baseline via t-SNE~\cite{van2008visualizing} and present a \emph{TaCo}-based task similarity matrix.

\subsection{Quantitative Results}

\textbf{Visual Task Classification:}
Our main results are displayed in~\Cref{tab:knn_combined}.
We evaluate all representations in nine settings, along the two dimensions of (1) seen tasks, unseen tasks and all tasks and (2) the dimension of seen datasets, unseen datasets and both seen and unseen datasets.
We report the results for kNN classification for $k=\{1,3,5\}$ in F1-score.
On seen datasets and seen tasks, we observe, that even without training on our medical data, CLIP and the ImageNet pre-trained ResNet can be utilized fairly well to distinguish seen visual tasks, with CLIP producing a slightly higher score of $73.6\%$ F1-score as opposed to $71.7\%$ in 1-NN classification.
Training SimCLR on the medical data leads to better 1-NN scores on seen datasets than these two baselines for seen and unseen tasks, only when considering all tasks CLIP still produces better results, indicating, that naive contrastive training is not enough to better capture task representations.
When taking a look at the results of our \emph{TaCo} strategy, we see a difference in performance of over $10\%$ F1-score on seen tasks and seen datasets.
For unseen tasks on seen datasets this margin is less pronounced, with margins of $+1.9\%, 2.3\%, 2.5\%$ for $k=\{1,3,5\}$.
Generally, considering all tasks on the seen datasets, there is a large gap of $7.2\% - 7.6\%$ between the best baseline results and \emph{TaCo}.
The better classification when unseen tasks are present shows \emph{TaCo}'s advantages over the in-domain trained SimCLR.
{
\setlength{\tabcolsep}{5.1pt}
\begin{table*}
    \centering
    \caption{Results for kNN classification of vision tasks in F1-scores. We evaluate a ResNet pre-trained on ImageNet, the multi-modal CLIP vision encoder, the contrastive baseline SimCLR trained in-domain, and our method TaCo trained without and with \emph{failure tasks}.}
    \scriptsize

    \begin{tabular}{lccc|ccc|ccc}
        \toprule
         Classification into: \emph{visual tasks} & \multicolumn{3}{c}{seen tasks} & \multicolumn{3}{c}{unseen tasks} & \multicolumn{3}{c}{all tasks} \\
          & $k = 1$ & $k = 3$ & $k = 5$ & $k = 1$ & $k = 3$ & $k = 5$ & $k = 1$ & $k = 3$ & $k = 5$ \\
         \midrule
         & \multicolumn{9}{c}{\textit{Nearest neighbor classification on seen datasets}} \\
         CLIP~\cite{cherti2023reproducible}   & $73.6\%$ & $73.8\%$ & $74.0\%$ & $82.4\%$ & $82.2\%$ & $81.7\%$ & $70.9\%$ & $71.1\%$ & $71.1\%$ \\
         ImageNet~\cite{he2016deep} & $71.7\%$ & $71.9\%$ & $71.8\%$ & $81.1\%$ & $81.0\%$ & $80.7\%$ & $69.1\%$ & $69.2\%$ & $69.1\%$ \\
         SimCLR~\cite{chen2020simple} & $75.1\%$ & $74.7\%$ & $73.9\%$ & $84.0\%$ & $83.1\%$ & $82.1\%$ & $70.6\%$ & $70.2\%$ & $69.6\%$ \\
         TaCo (ours) & $\mathbf{85.2}\%$ & $\mathbf{85.0}\%$ & $\mathbf{84.6}\%$ & $\mathbf{85.9}\%$ & $\mathbf{85.4}\%$ & $\mathbf{84.6}\%$ & $\mathbf{78.4}\%$ & $\mathbf{78.7}\%$ & $\mathbf{78.3}\%$ \\
         TaCo (ours) + \emph{failure tasks} & $81.3\%$ & $80.7\%$ & $80.3\%$ & $82.1\%$ & $81.5\%$ & $79.9\%$ & $76.2\%$ & $76.1\%$ & $75.5\%$ \\
         & \multicolumn{9}{c}{\textit{Nearest neighbor classification on unseen datasets}} \\
         CLIP~\cite{cherti2023reproducible}  & $83.6\%$ & $83.8\%$ & $84.1\%$ & $93.5\%$ & $94.7\%$ & $95.4\%$ & $80.3\%$ & $80.4\%$ & $80.7\%$ \\
         ImageNet~\cite{he2016deep} & $79.6\%$ & $79.9\%$ & $80.1\%$ & $92.4\%$ & $92.8\%$ & $93.1\%$ & $76.8\%$ & $77.1\%$ & $77.9\%$ \\
         SimCLR~\cite{chen2020simple} & $82.1\%$ & $82.5\%$ & $82.9\%$ & $93.3\%$ & $93.6\%$ & $93.5\%$ & $77.8\%$ & $77.9\%$ & $78.2\%$ \\
         TaCo (ours) & $\mathbf{92.7}\%$ & $\mathbf{92.8}\%$ & $\mathbf{92.7}\%$ & $95.7\%$ & $95.4\%$ & $95.0\%$ & $85.4\%$ & $85.5\%$ & $85.5\%$ \\
         TaCo (ours) + \emph{failure tasks} & $92.4\%$ & $92.5\%$ & $92.2\%$ & $\mathbf{97.0}\%$ & $\mathbf{96.7}\%$ & $\mathbf{96.1}\%$ & $\mathbf{87.0}\%$ & $\mathbf{87.0}\%$ & $\mathbf{86.9}\%$ \\
         & \multicolumn{9}{c}{\textit{Nearest neighbor classification on seen and unseen datasets}} \\
         CLIP~\cite{cherti2023reproducible}  & $70.2\%$ & $70.4\%$ & $70.4\%$ & $78.2\%$ & $78.8\%$ & $79.0\%$ & $67.2\%$ & $67.6\%$ & $67.8\%$ \\
         ImageNet~\cite{he2016deep} & $68.2\%$ & $68.4\%$ & $68.3\%$ & $76.9\%$ & $77.1\%$ & $77.0\%$ & $66.0\%$ & $66.3\%$ & $66.6\%$\\
         SimCLR~\cite{chen2020simple} & $71.5\%$ & $71.2\%$ & $70.5\%$ & $79.7\%$ & $79.3\%$ & $78.5\%$ & $67.5\%$ & $67.4\%$ & $67.0\%$ \\
         TaCo (ours) &$\mathbf{81.1}\%$ & $\mathbf{81.0}\%$ & $\mathbf{80.6}\%$ & $\mathbf{81.2}\%$ & $\mathbf{80.9}\%$ & $\mathbf{80.3}\%$ & $\mathbf{74.3}\%$ & $\mathbf{74.4}\%$ & $\mathbf{74.0}\%$\\
         TaCo (ours) + \emph{failure tasks} & $77.0\%$ & $76.6\%$ & $75.9\%$ & $76.9\%$ & $75.9\%$ & $74.8\%$ & $71.3\%$ & $71.3\%$ & $70.6\%$\\
         \bottomrule
    \end{tabular}
    \label{tab:knn_combined}
\end{table*}
}

The difference between \emph{TaCo} and \emph{TaCo} with \emph{failure tasks} shows when observing the evaluation on unseen datasets (rows six through ten).
There, when considering unseen tasks, the added \emph{failure tasks} lead to a better distinction and higher scores with $+1.3\%, +1.3\%, +1.1\%$ ($k=\{1,3,5\}$).
This advantageous treatment of unseen tasks also shows in a better performance in the all tasks scenario on unseen datasets.
On seen and unseen datasets (last five rows), \emph{TaCo} without failure tasks leads to the hightest scores.
This is influenced by the distribution between seen- and unseen visual tasks, in our setup the former are more numerous, as such, the advantageous treatment for unseen tasks when training with \emph{failure tasks} is weighted less.
In summary, both \emph{TaCo} variants best represent visual tasks but with different foci.

\textbf{Influence of Task- and Data Distribution:}
Dissecting the visual tasks and investigating whether the representations better capture and distinguish the input data distribution (\ie, classify the $39$ datasets) or the task information (\ie, classify the $30$ tasks) is shown in~\Cref{tab:seen_datasets}.
The three scenarios shown correspond to scenarios on the diagonal of~\Cref{tab:knn_combined}: (1) seen tasks, seen datasets (2) unseen tasks, unseen datasets and (3) all tasks, all datasets.

\begin{wraptable}{r}{0.625\textwidth}
    \caption{Nearest neighbor classification ($k=5$) in F1-Score of task and dataset classification of different representations. Insight into method's focus on tasks or datasets.}
    \centering
    \scriptsize
    \setlength{\tabcolsep}{9pt}

    \begin{tabular}{lcc}
        \toprule
         Classification into: & \multicolumn{1}{c}{tasks} & \multicolumn{1}{c}{datasets} \\
         \midrule
         \multicolumn{3}{c}{\textit{Seen  tasks on seen datasets}} \\
         CLIP~\cite{cherti2023reproducible} & $85.1\%$ & $85.6\%$ \\
         ImageNet~\cite{he2016deep} & $80.9\%$ & $86.0\%$ \\
         SimCLR~\cite{chen2020simple} & $82.7\%$ & $86.2\%$ \\
         TaCo (ours) & $\textbf{96.4}\%$ & $\textbf{89.4}\%$ \\
         TaCo (ours) + \emph{failure tasks} & $95.3\%$ & $84.4\%$ \\
         \multicolumn{3}{c}{\textit{Unseen  tasks on unseen datasets}} \\
         CLIP~\cite{cherti2023reproducible} & $94.7\%$ & $\textbf{99.9}\%$ \\
         ImageNet~\cite{he2016deep} & $92.8\%$ & $99.8\%$ \\
         SimCLR~\cite{chen2020simple} & $93.2\%$ & $99.6\%$ \\
         TaCo (ours) & $\textbf{96.9}\%$ & $98.8\%$ \\
         TaCo (ours) + \emph{failure tasks} & $96.4\%$ & $97.9\%$ \\
         \multicolumn{3}{c}{\textit{All  tasks on all datasets}} \\
         CLIP~\cite{cherti2023reproducible} & $82.6\%$ & $81.8\%$ \\
         ImageNet~\cite{he2016deep} & $78.1\%$ & $82.7\%$ \\
         SimCLR~\cite{chen2020simple} & $78.2\%$ & $82.5\%$ \\
         TaCo (ours) & $89.2\%$ & $\textbf{84.1}\%$ \\
         TaCo (ours) + \emph{failure tasks} & $\textbf{90.2}\%$ & $78.7\%$ \\
         \bottomrule
    \end{tabular}
    \label{tab:seen_datasets}
\end{wraptable}


Considering the task column, we directly see, that the gap between the two \emph{TaCo} methods and the rest is quite pronounced in all three scenarios, indicating the effect of training towards distinguishing tasks.
For the dataset column, we see, that the gap is much smaller between baselines and \emph{TaCo} models, as such the main difference in performance stems from the task rather than the dataset categorization.
In the scenario of unseen tasks on unseen datasets, all models achieve close to perfect classification, this is due to only four datasets that have to be distinguished which all stem from different imaging modalities, making it very easy.
Notably, \emph{TaCo with failure tasks} consistently performs worst on dataset classification which indicates that using failure tasks shifts the focus towards representing tasks more discriminately and representing datasets is a secondary factor.

\subsection{Ablation Study}

In our ablation study, in~\Cref{tab:ablation_study}, we investigate the effects of our design choices in \emph{TaCo}.
First, we compare standard SimCLR training on the data with our \emph{Task-Contrastive Learning}, where we see that the latter improves the performance on seen tasks, while laying less emphasis on which dataset an instance originates from, which leads to an overall lower score.
When \emph{balanced task sampling} is added, the F-1 Score improves significantly, as less emphasis is put on visual tasks with a high number of instances.
When adding \emph{failure tasks}, we observe a lower performance on seen visual tasks (column three), while on unseen visual tasks (column six) performance is increases slightly. 
Quantitatively, we see \emph{TaCo} with \emph{balanced task sampling} leads to best overall results (column nine), \emph{TaCo + failure tasks} embeds unseen tasks for unseen datasets most conveniently (column six).


{
\setlength{\tabcolsep}{1.9pt}
\begin{table*}[t]
    \centering
    \scriptsize
    \caption{Ablation for kNN ($k=5$) classification of tasks, datasets and visual tasks for seen, unseen and all visual tasks in F-1 Scores.}
    \begin{tabular}{lccc|ccc|ccc}
        \toprule
         & \multicolumn{3}{c}{\textit{\scriptsize Seen tasks and seen datasets}} & \multicolumn{3}{c}{\textit{\scriptsize Unseen tasks and unseen datasets}} & \multicolumn{3}{c}{\scriptsize \textit{Seen and unseen tasks and datasets}} \\
         Classification into: & tasks & datasets & visual tasks & tasks & datasets & visual tasks & tasks & datasets & visual tasks \\
         \midrule
         SimCLR & $82.7\%$ & $86.2\%$ & $73.9\%$ & $93.2\%$ & $\mathbf{99.6}\%$ & $93.5\%$ & $78.2\%$ & $82.5\%$ & $67.0\%$ \\
         TaCo & $89.8\%$ & $80.9\%$ & $71.9\%$ & $93.8\%$ & $98.2\%$ & $92.0\%$ & $83.8\%$ & $77.1\%$ & $63.5\%$ \\
         + \emph{balanced task sampling} & $\mathbf{96.4}\%$ & $\mathbf{89.4}\%$ & $\mathbf{84.6}\%$ & $\mathbf{96.9}\%$ & $98.8\%$ & $95.0\%$ & $89.2\%$ & $\mathbf{84.1}\%$ & $\mathbf{74.0}\%$ \\
         + \emph{failure tasks} & $95.3\%$ & $84.4\%$ & $80.3\%$ & $96.4\%$ & $97.9\%$ & $\mathbf{96.1}\%$ & $\mathbf{90.2}\%$ & $78.7\%$ & $70.6\%$ \\
         \bottomrule
    \end{tabular}
    \label{tab:ablation_study}
\end{table*}
}

\subsection{Probing the Structure of Task Contrastive Representations}

\subsubsection{Iterative Task Adaptation}
\begin{figure}[t]
    \textit{Successively altering the brightness factor from lower brightness to higher brightness}
    
    \includegraphics[width=\linewidth]{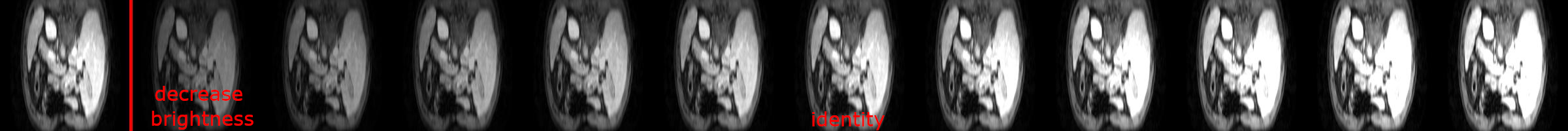}
    
    \textit{Successively rotating the image by increasing degrees}
    
    \includegraphics[width=\linewidth]{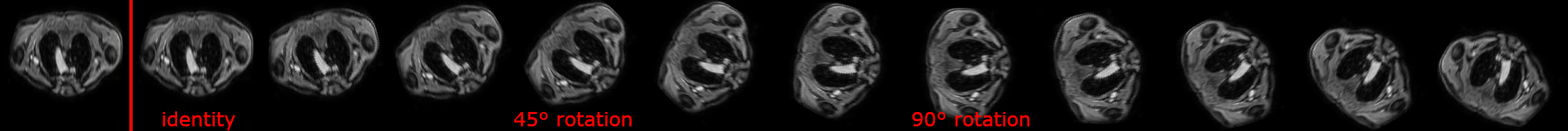}
    
    \textit{Successively adding noise to the segmentation pixel values}
    
    \includegraphics[width=\linewidth]{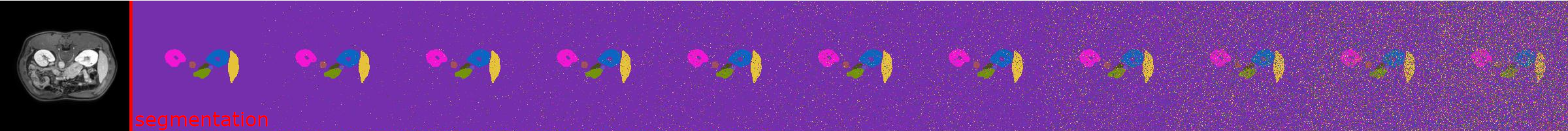}
    
    \caption{Examples for iterative task adaptation. \underline{Top}: Brightness is changed successively. \underline{Middle}: Image is successively rotated. \underline{Bottom}: Segmentation of an image is successively noised. Only subsequences are displayed here for brevity.}
    \label{fig:iterativetasks}
\end{figure}

Next, we showcase how distances between two tasks behave when one of the tasks is iteratively altered.
In~\Cref{fig:iterativetasks}, we display the scenario of iteratively changing a task, more precisely we show the iteratively changed tasks of \emph{brightness change}, \emph{rotation} and \emph{adding noise to segmentation maps}.
In the figure, the original image is displayed to the left of the red vertical lines, to the right, the iterative task adaptation is displayed.
Throughout these sequences of iteratively adapted tasks, some of the intermediate tasks are part of our training tasks, these tasks are marked with red text.
Specifically, the decrease brightness task, the identity task (input image equals output image) as well as some rotation tasks with specific degrees and the segmentation task are present in our training tasks.

\emph{Brightness change}: In the first row and first column in~\Cref{fig:iterativetasks}, the input image is displayed with the original brightness, to its right, the brightness of this image is set to half the brightness of the original, then the brightness is successively increased. The first image corresponds to how we define the task of decrease brightness from our training tasks, the $6^{th}$ image to the right of the original image has the same brightness as the original, as such it forms the identity task.

\emph{Rotation}: Here, the input image is successively rotated by $15^\circ$, which leads to different rotation tasks. The rotation by $45^\circ$ task is one of the unseen tasks in our setup, rotation by $90^\circ$ is a task from our training set.

\emph{Noisy segmentation}: The original segmentation task from training is displayed as the first image to the right of the original image in~\Cref{fig:iterativetasks}, all successive tasks are obtained by randomly changing an increasing portion of the pixels to another color from the original segmentation map.
As such, the segmentation task successively degenerates more and more.

\begin{figure*}[t]
    \centering
    \rotatebox[origin=lb]{90}{$\phantom{.}$\scriptsize distance in CLIP space}\includegraphics[trim={0 1.1cm 0 0.22cm},clip,width=0.3\textwidth, height=0.2\textwidth ]{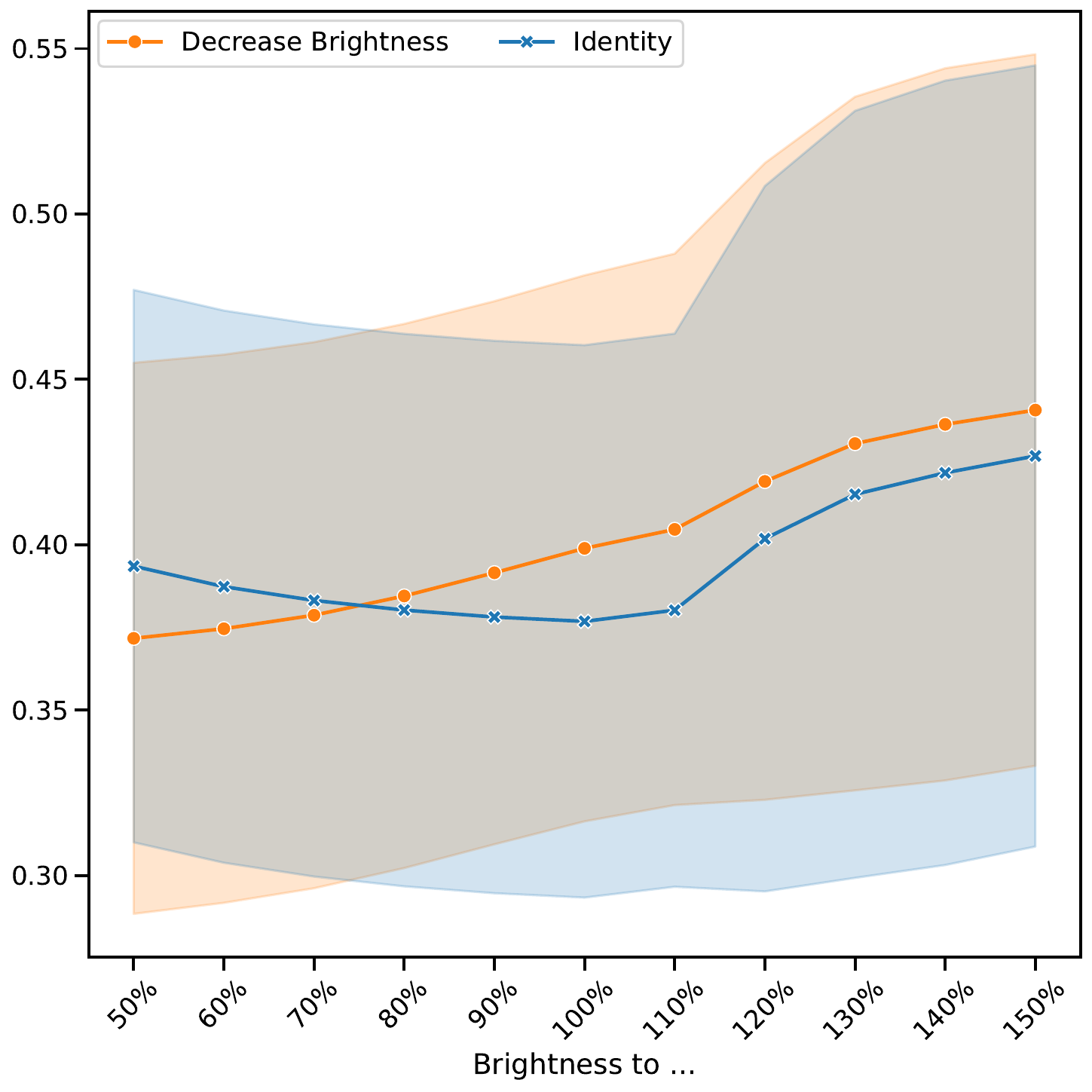}  
    \includegraphics[trim={0 1.1cm 0 0.22cm},clip,width=0.3\textwidth, height=0.2\textwidth ]{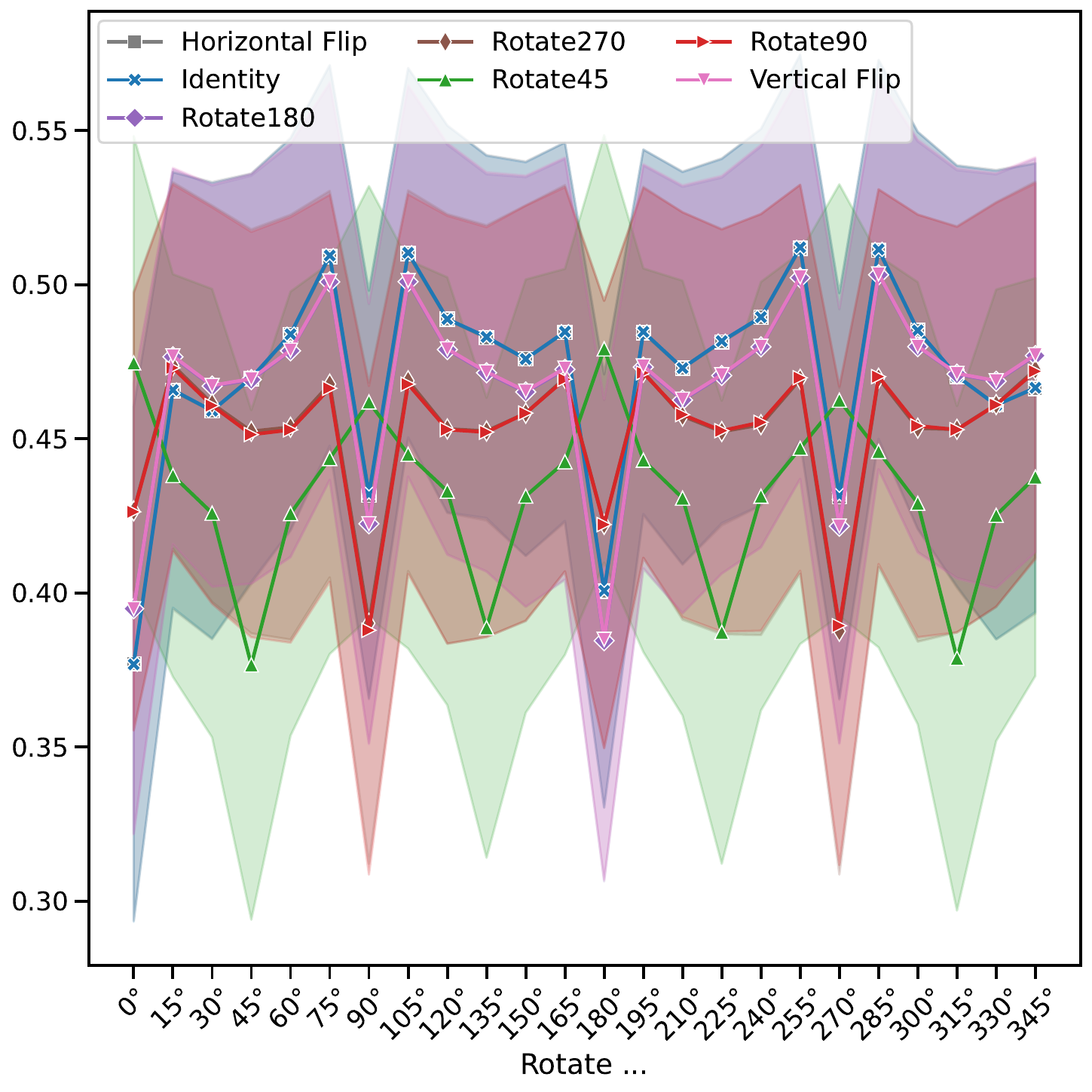}
    \includegraphics[trim={0 1.1cm 0 0.22cm},clip,width=0.3\textwidth, height=0.2\textwidth ]{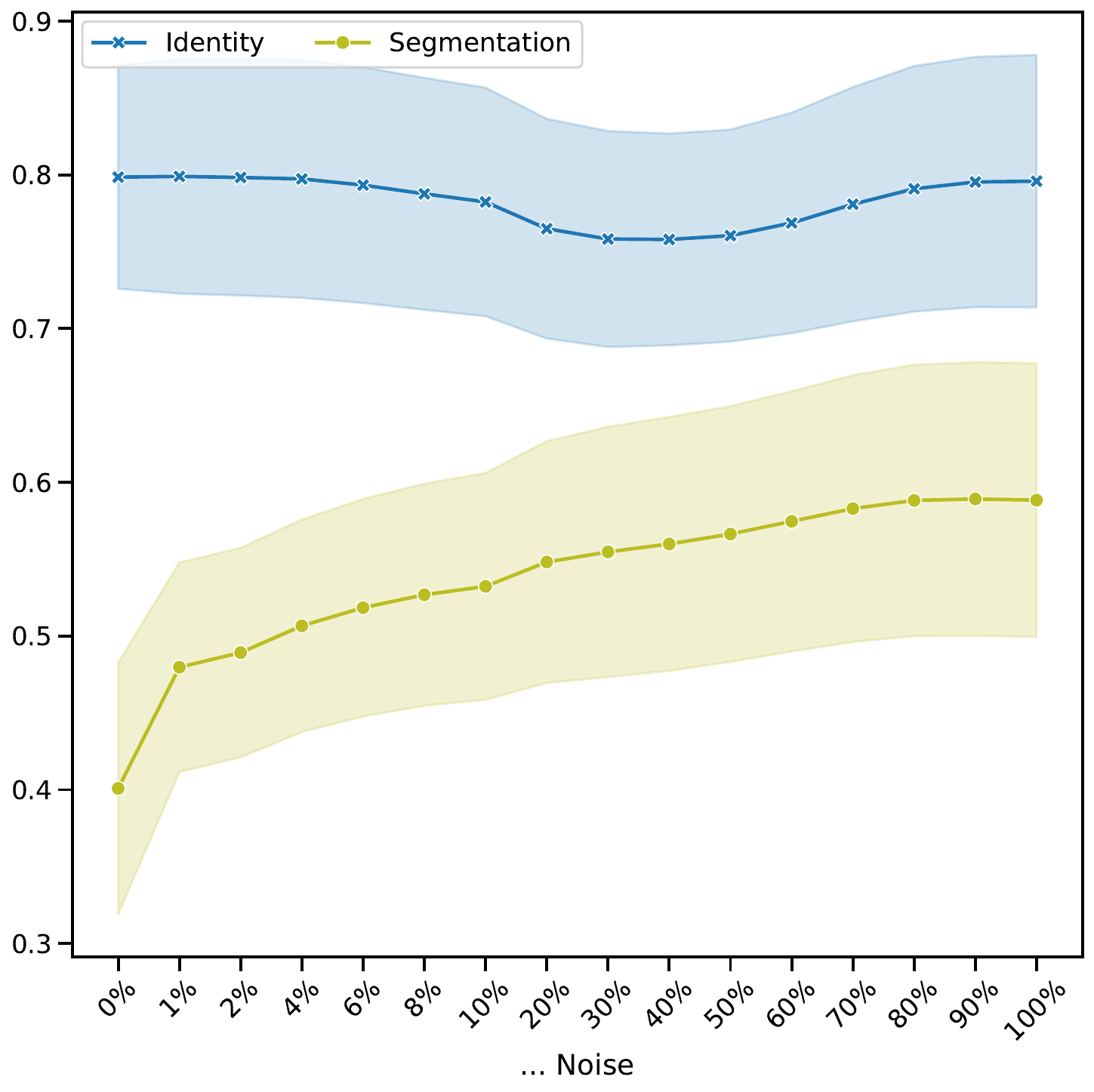}
        
    \rotatebox[origin=lb]{90}{$\phantom{.}$\scriptsize distance in \emph{TaCo} space}    \includegraphics[trim={0 1.1cm 0 0.22cm},clip,width=0.3\textwidth, height=0.2\textwidth ]{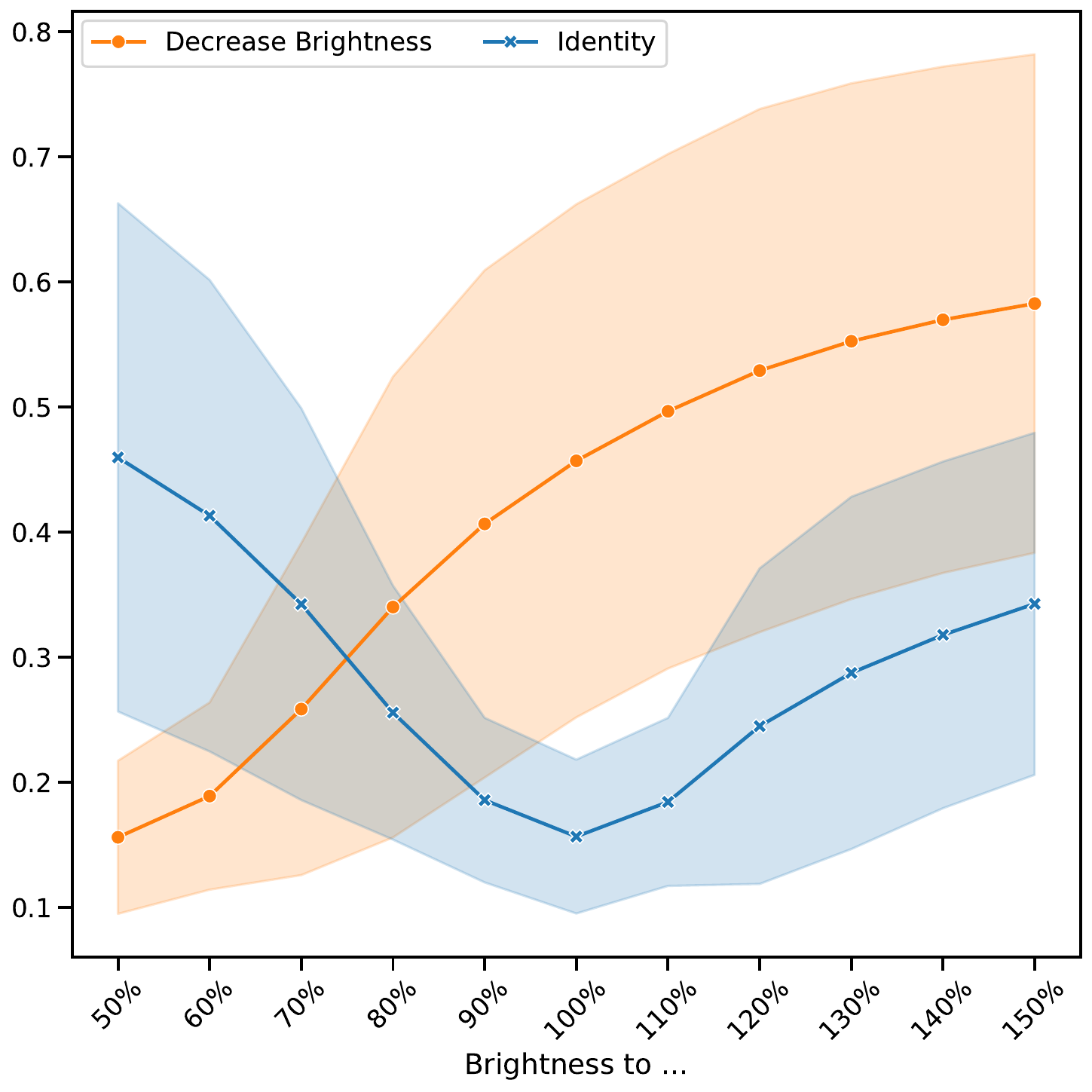}
    \includegraphics[trim={0 1.1cm 0 0.22cm},clip,width=0.3\textwidth, height=0.2\textwidth ]{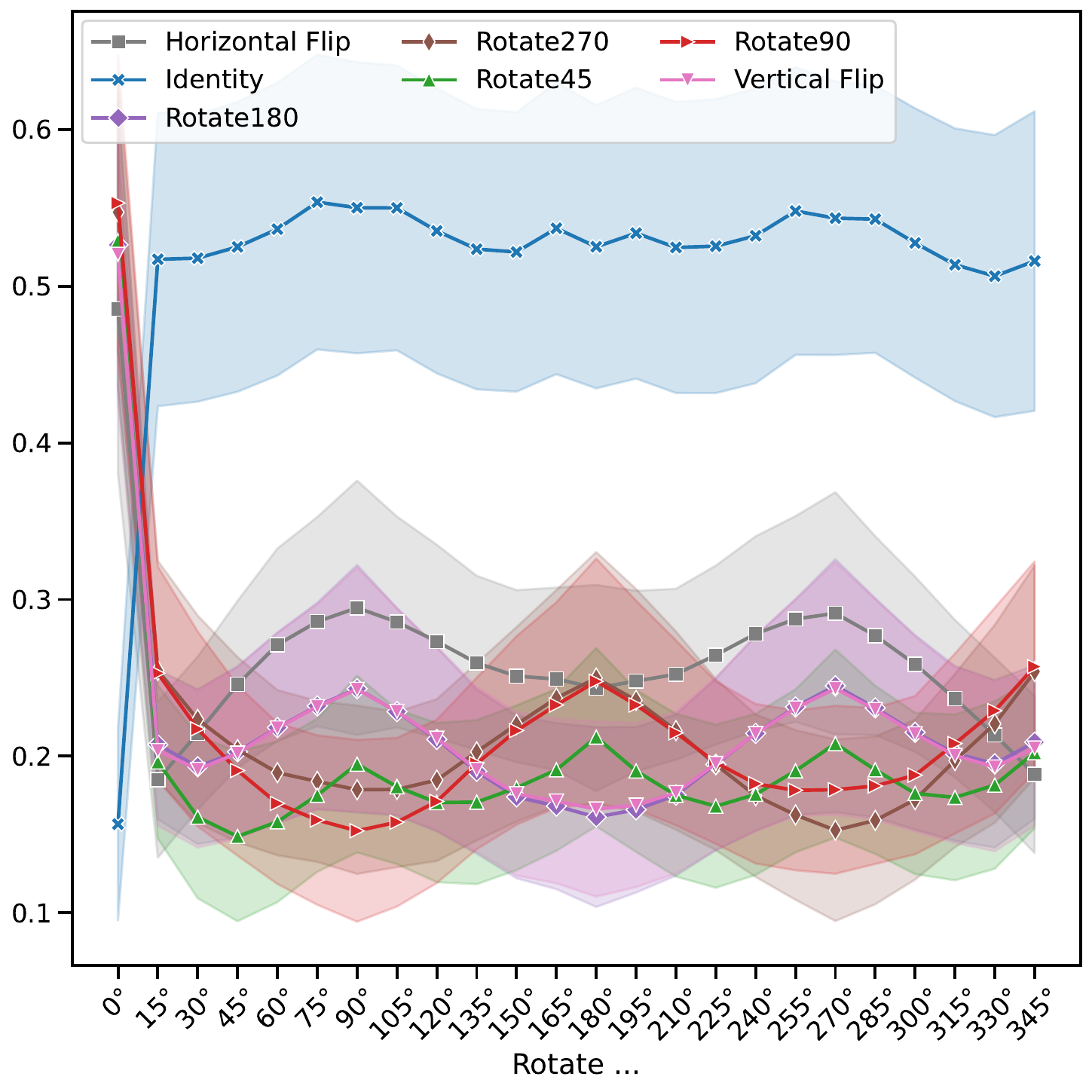}
    \includegraphics[trim={0 1.1cm 0 0.22cm},clip,width=0.3\textwidth, height=0.2\textwidth ]{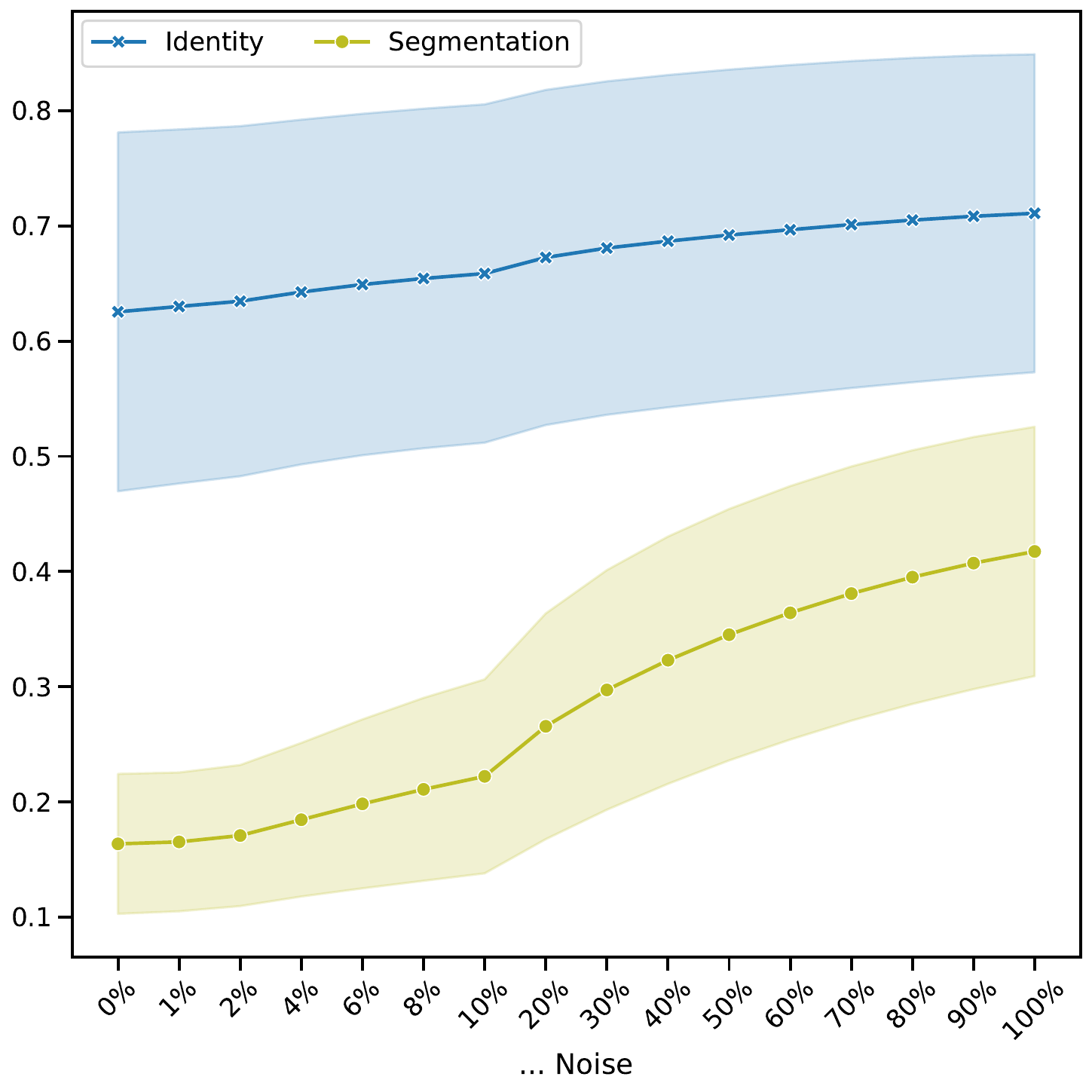}

    \rotatebox[origin=lb]{90}{$\phantom{-}$\scriptsize distance in \emph{TaCo}}\rotatebox[origin=lb]{90}{$\phantom{-}$\scriptsize \emph{+ failure task} space}    \includegraphics[trim={0 1.1cm 0 0.22cm},clip,width=0.3\textwidth, height=0.2\textwidth ]{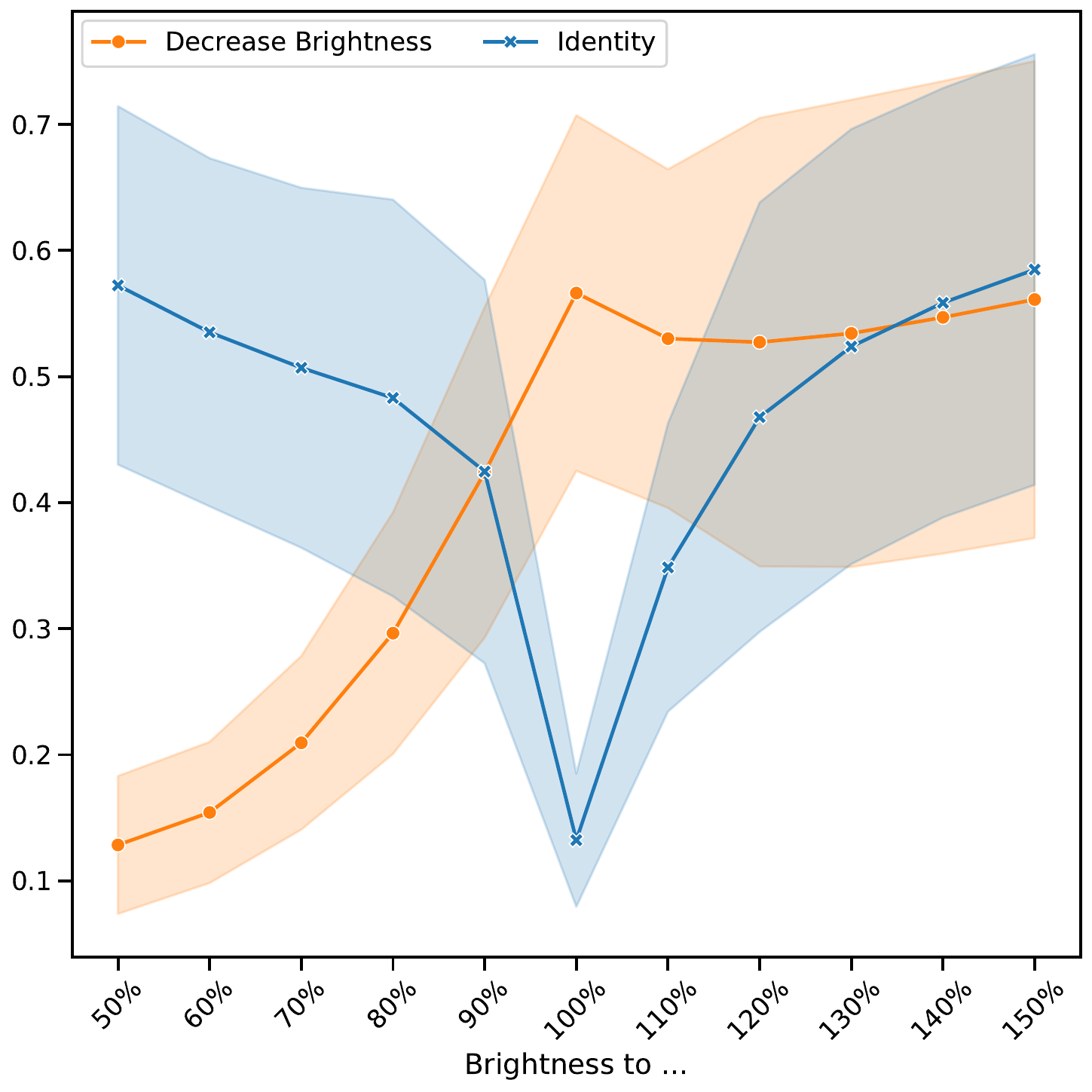}
    \includegraphics[trim={0 1.1cm 0 0.22cm},clip,width=0.3\textwidth, height=0.2\textwidth ]{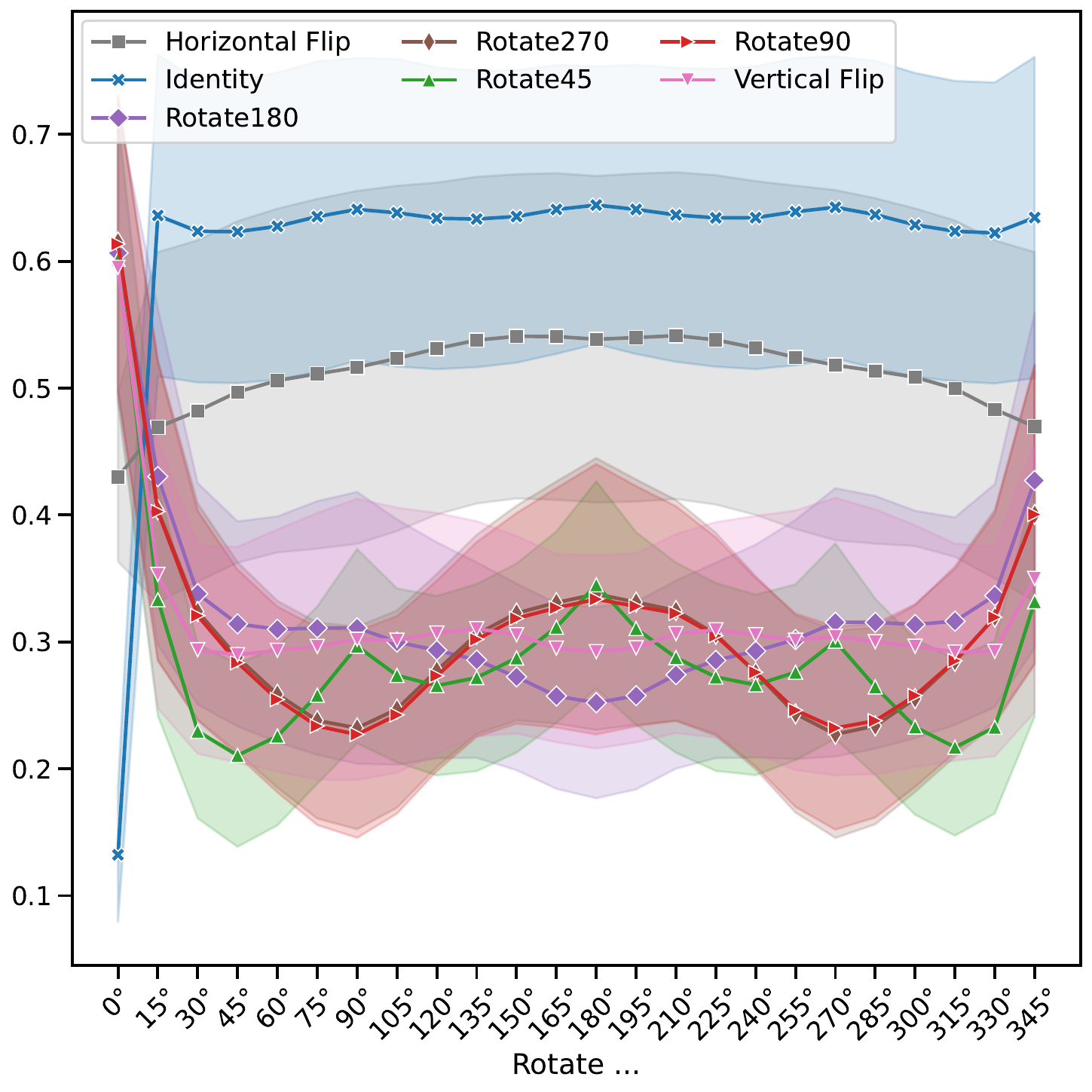}
    \includegraphics[trim={0 1.1cm 0 0.22cm},clip,width=0.3\textwidth, height=0.2\textwidth ]{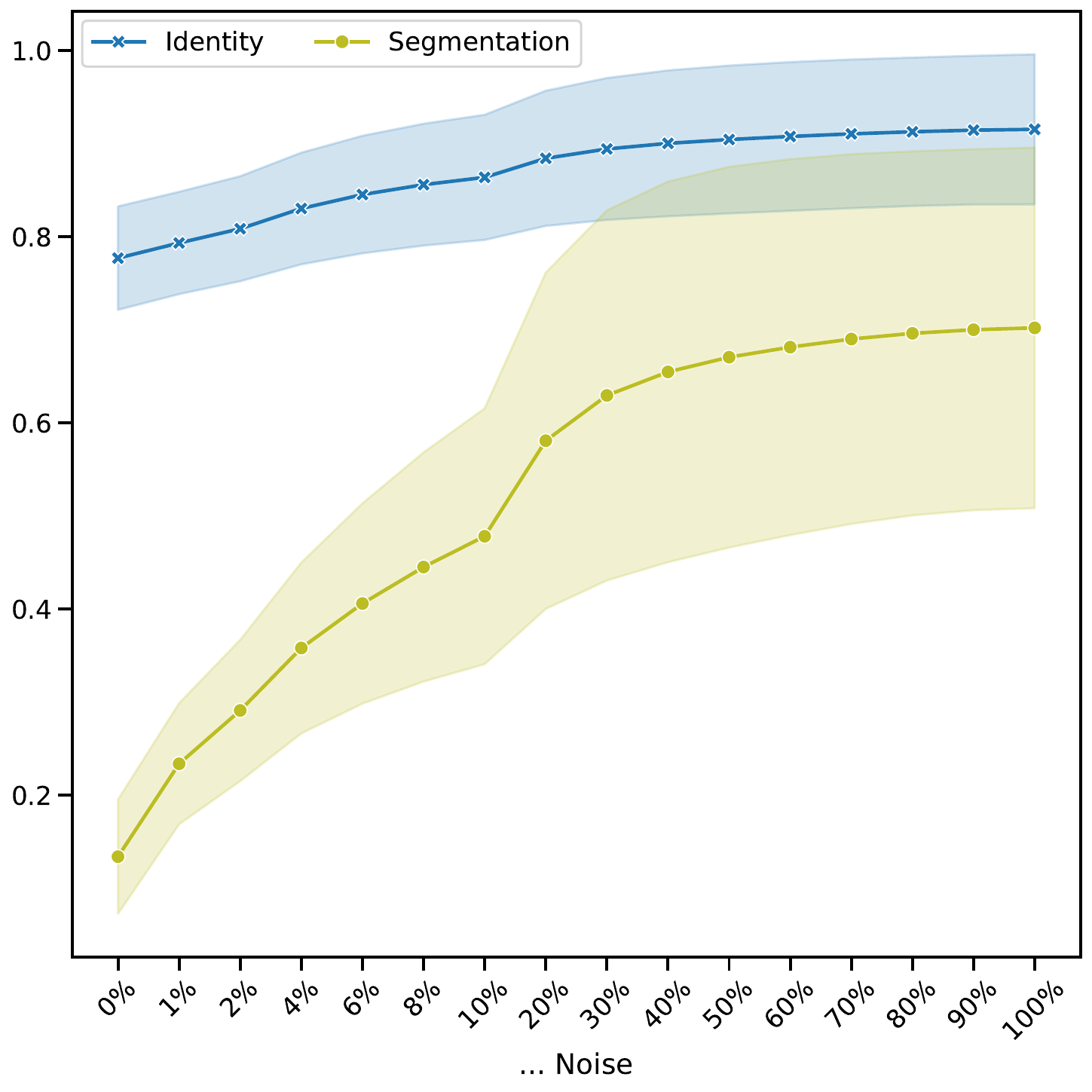}
    {
    \setlength{\tabcolsep}{30pt}
    \begin{tabular}{ccc}
          $\phantom{-}$Brightness change & $\phantom{.}$ Rotation & Noisy segmentation
    \end{tabular}
    }
    \caption{Iterative task adaptation. First row: Distances in CLIP representation space, second row: distances in \emph{TaCo} representation space. Third row: distances in~\emph{TaCo + failure tasks} representation space. Columns indicate iterative task adaptations for the tasks: changing brightness, rotating an image and adding successively more noise to a segmentation task. The x-axes describe the magnitude of task adaptation, colored areas relate to standard deviation. Note: The x-axis in the noisy segmentation plots first has a step size of $2\%$ and successively a step size of $10\%$.}
    \label{fig:iterative_task_adaptation2}
\end{figure*}

These iterative task adaptations make up the x-axes in the plots of~\Cref{fig:iterative_task_adaptation2}, where we visualize distances in task representation spaces, specifically, the CLIP space, \emph{TaCO} space and \emph{TaCO} space when trained with failure tasks.
Note, that we only display a subset of the intermediate tasks in~\Cref{fig:iterativetasks}, due to their large number.
In column one, for example, the orange line measures distances in representation space between the decrease brightness task as in our training set and the intermediate tasks when successively increasing the brightness factor (x-axis), leading to more and more different task instances.
We observe in the three plots in the first column of~\Cref{fig:iterative_task_adaptation2}, that successively increasing the brightness in the task definition leads to a higher and higher distance to the decrease brightness task from our dataset, while the distance to the identity task reaches a minimum when the brightness change is set to $100\%$, which exactly resembles the identity task.
Comparing \emph{TaCo} and CLIP, we see these trends much more pronounced for our method.
Adding failure tasks in training the task representations (row three) leads to sharper drops and peaks in the measured distances when tasks from training are reached.

A very interesting insight can be gathered from interpolating between rotation tasks in column two. For the identity task, we see a minimum at a rotation of $0^\circ$ in the \emph{TaCo} representations (row two), which is expected, as the tasks are equivalent.
With increased rotation degrees, the distance to the identity task steeply increases, but the distances to the rotation tasks from the training set, \ie, 45$^\circ$, 90$^\circ$, $180^\circ$, $270^\circ$ tasks, decrease.
The distances to these rotation tasks form minima at their own rotation degrees and -- which is notable -- at multiples of their degrees.
This shows that distinguishing those tasks from each other is hard for the networks, which is mirrored in research on vision-language models~\cite{anis2025limitations}.
Here, this phenomenon can also be partly explained by some medical data not having up and down directions such as microscopy data.

The tasks horizontal- and vertical flip produce local minima at $180^\circ$ as this rotation leads to a visually similar task.
At the same time, the distances for flipping tasks lead to maxima around $90^\circ$ and $270^\circ$ rotation, indicating that the \emph{TaCo} representations can distinguish these geometric tasks well.
Such smooth periodic behavior is predominantly seen for both our \emph{TaCo} variants, CLIP, on the other hand, shows jagged lines (column one, row one). 


These observed trends, \ie, that the progressing task distances are much more smooth and pronounced for our~\emph{TaCo} models are confirmed in column three, where the distance between the segmentation task embeddings and the more and more noisy segmentation task variants successively increase for the \emph{TaCO} representations, while for CLIP embeddings the measured distances flatten more quickly.

\begin{figure*}[ht]
    \centering
    \begin{tabular}{ccc}
          $\phantom{-----.-}$ CLIP & $\phantom{----.--}$ TaCo & $\phantom{----}$ TaCo + \emph{failure tasks}$\phantom{-----.-}$ 
    \end{tabular}
    
    \rotatebox[origin=lb]{90}{$\phantom{--.}$colored by tasks}\includegraphics[trim={0 0 8cm 0},clip,width=0.25\textwidth,height=0.25\textwidth]{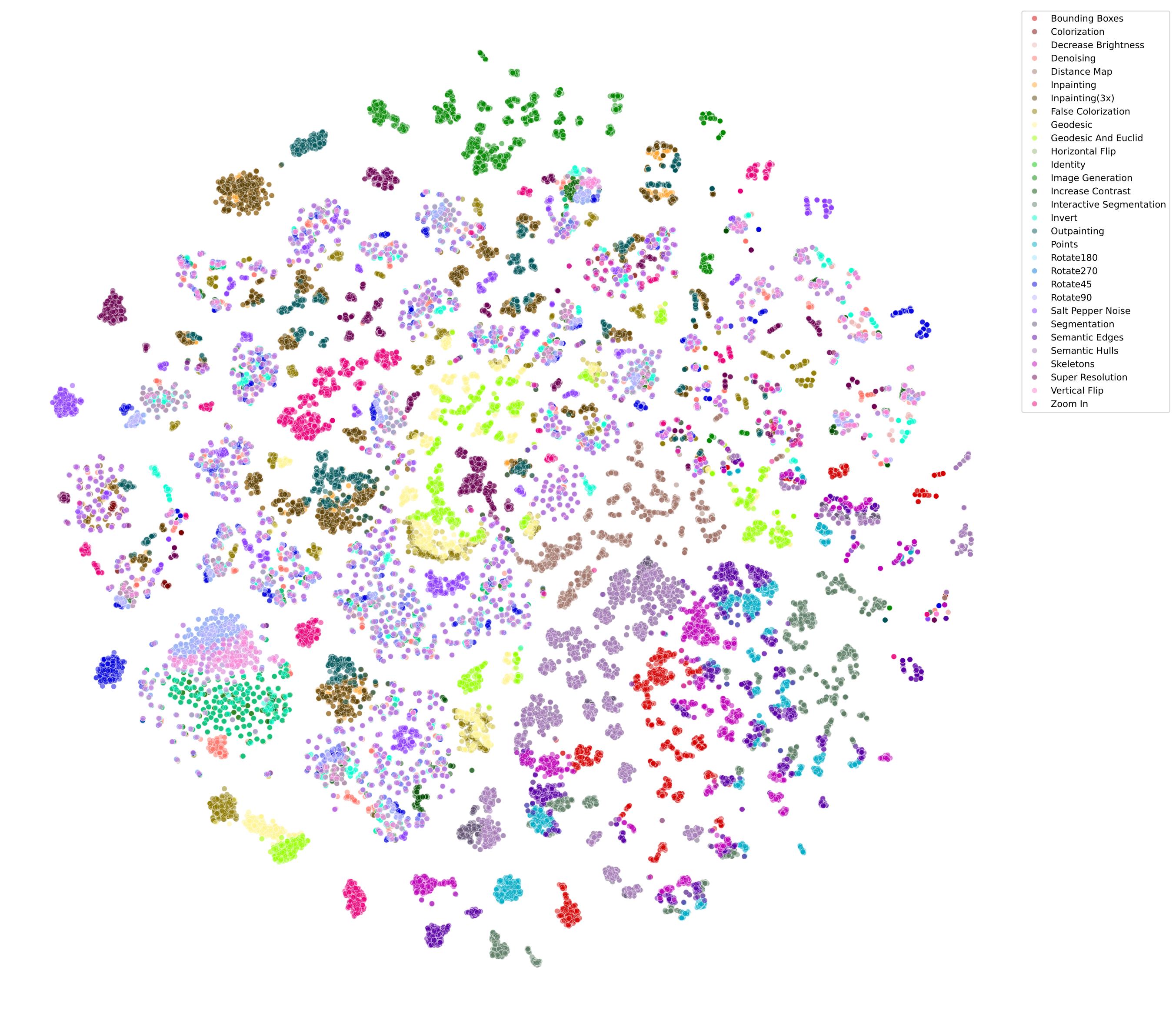}
    \includegraphics[trim={0 0 7.9cm 0},clip,width=0.25\textwidth,height=0.25\textwidth]{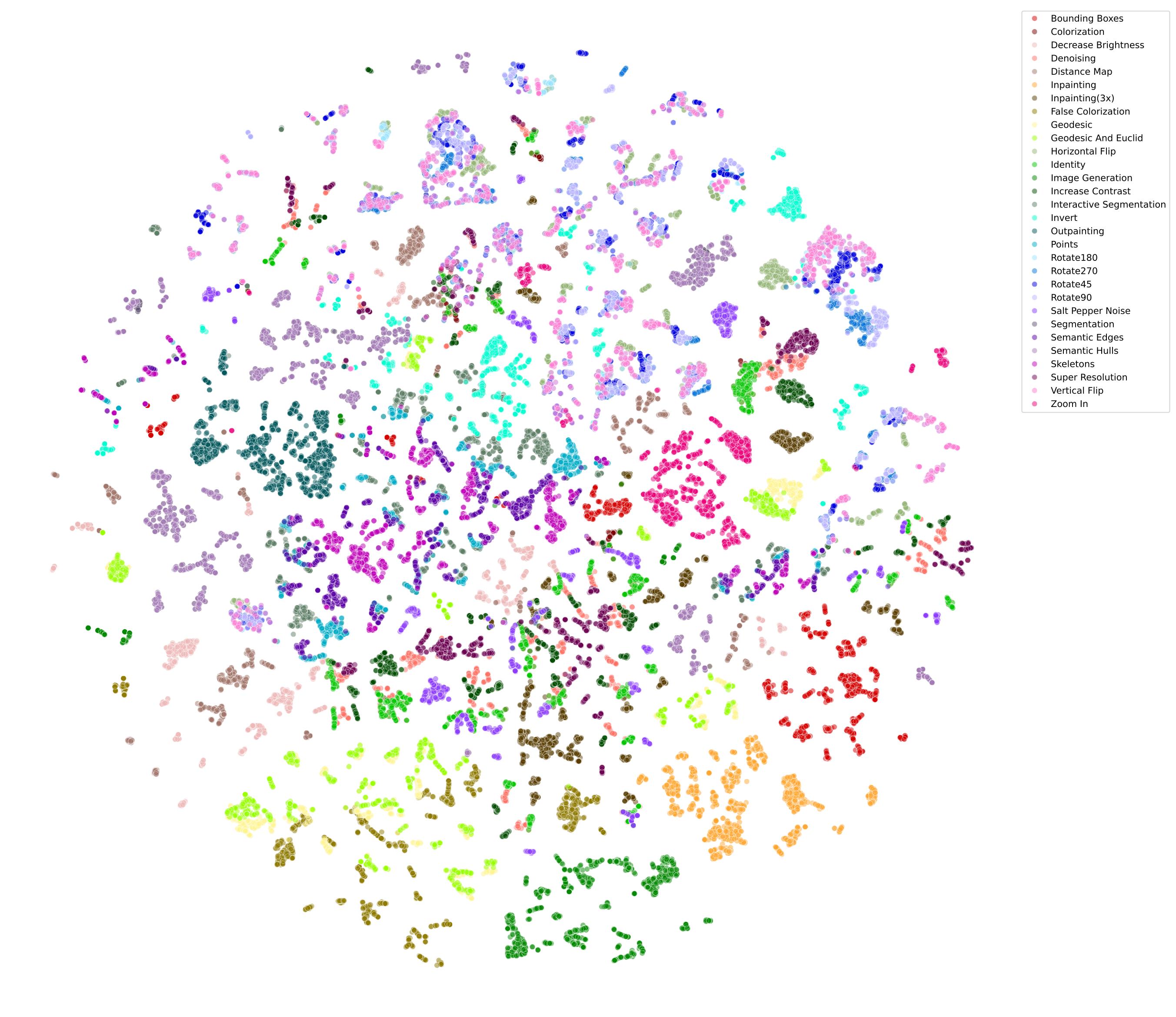}
    \includegraphics[trim={0 0 7.9cm 0},clip,width=0.25\textwidth,height=0.25\textwidth]{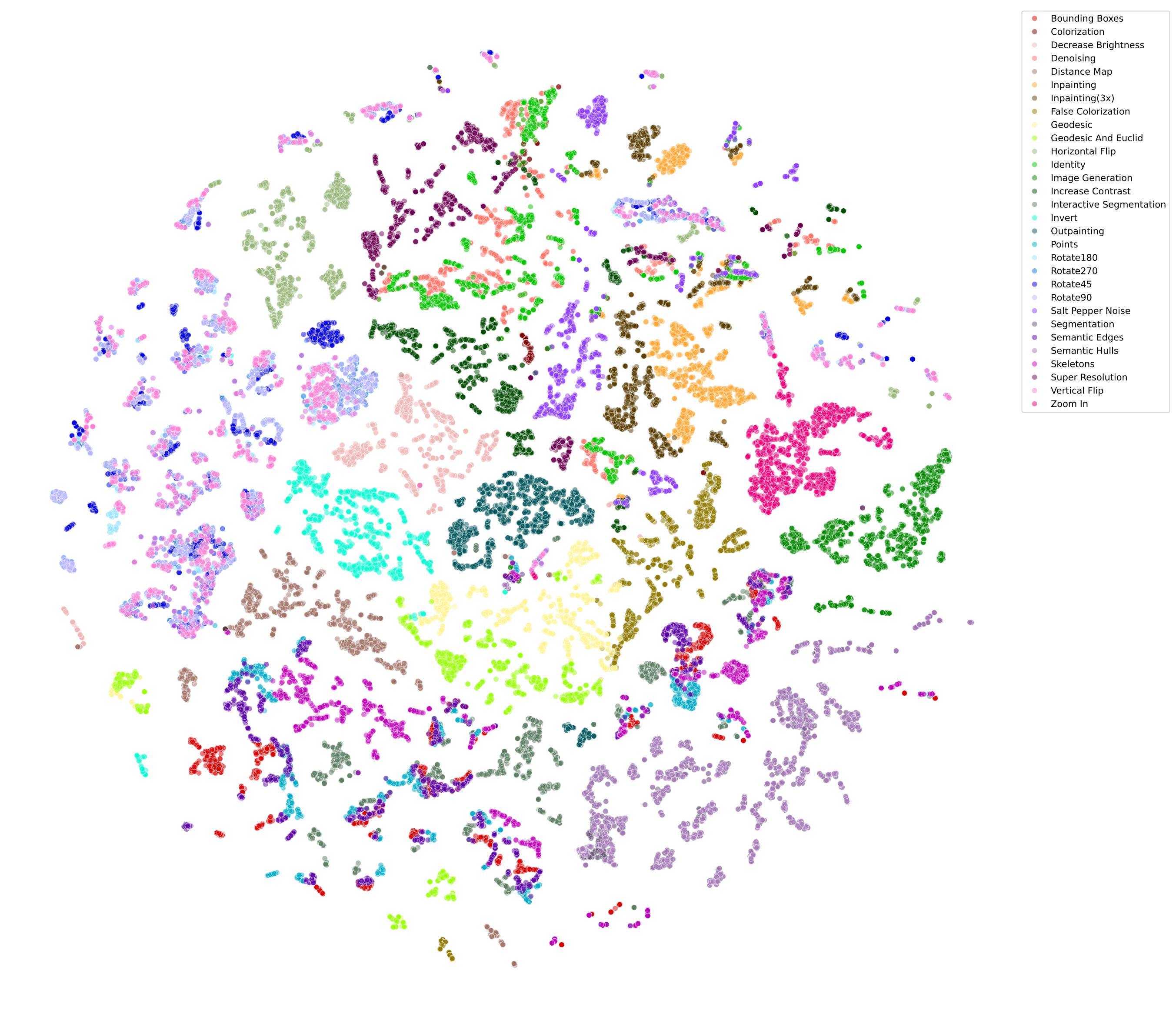}
    \includegraphics[trim={38.5cm 23cm 0 0},clip,height=0.25\textwidth]{figures/pdf_tsne_plots/SupCL/final_SupCL_sampled_negatives/all_tasks_all_datasets/tsne_classes_no_symbols_SupCL_final_SupCL_sampled_negatives_all_tasks_all_datasets_all.jpg}
        
    \rotatebox[origin=lb]{90}{$\phantom{--}$colored by datasets}\includegraphics[trim={0 0 9cm 0},clip,width=0.25\textwidth,height=0.25\textwidth]{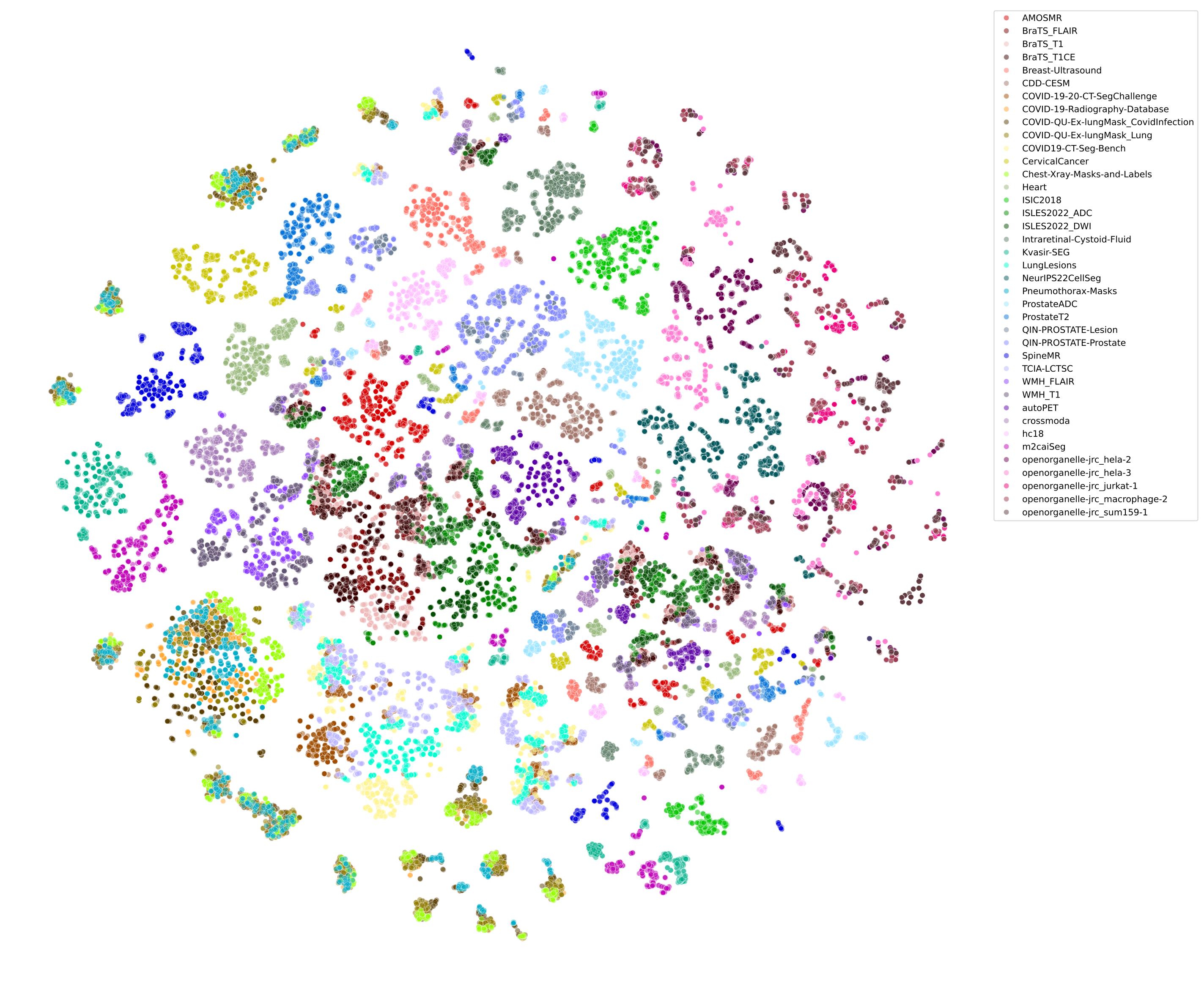}
    \includegraphics[trim={0 0 9cm 0},clip,width=0.25\textwidth,height=0.25\textwidth]{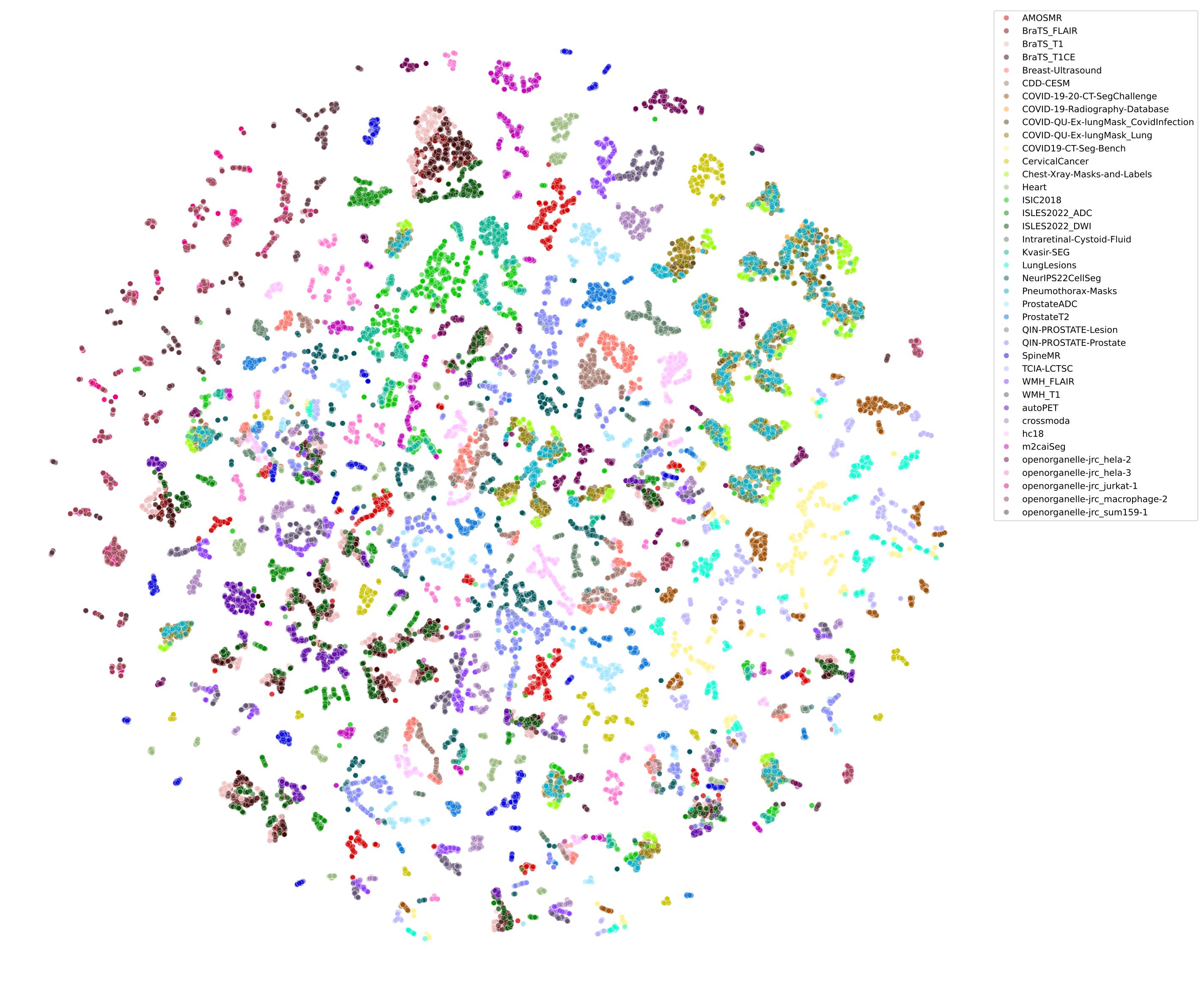}
    \includegraphics[trim={0 0 9cm 0},clip,width=0.25\textwidth,height=0.25\textwidth]{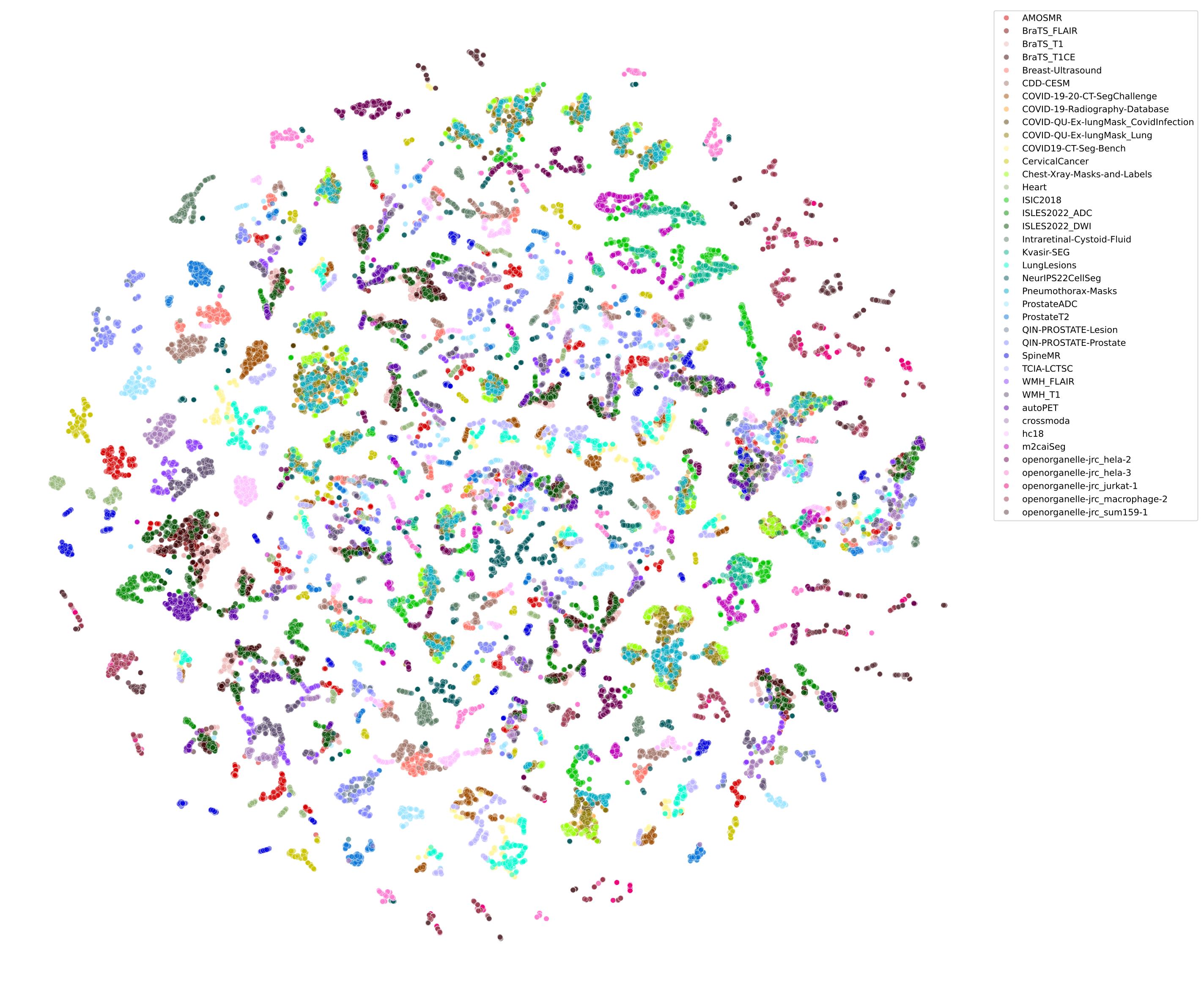}
    \includegraphics[trim={38cm 15cm 0 0},clip,height=0.25\textwidth]{figures/pdf_tsne_plots/SupCL/final_SupCL_sampled_negatives/all_tasks_all_datasets/tsne_datasets_no_symbols_SupCL_final_SupCL_sampled_negatives_all_tasks_all_datasets_all.jpg}
    \caption{Learned high dimensional task representation spaces projected via t-SNE. We showcase the embedding space of CLIP, \emph{TaCo} and \emph{TaCo + failure tasks}. Colors indicate different tasks (top row) and different datasets (bottom row), best viewed zooming in.}
    \label{fig:tsne}
\end{figure*}

\subsubsection{Task Representation Space Visualization}
\label{sec:tsnequalitative}
In~\Cref{fig:tsne}, we visualize the task representations by projecting the high-dimensional vectors into two dimensions with t-SNE~\cite{van2008visualizing}.
Colors indicate the tasks in the top row, and dataset affiliation in the second row.
While CLIP groups visual task instances sometimes into consistent task clusters (first row), it predominantly groups task instances that share the same dataset (second row).
\emph{TaCo} on the other hand produces coherent task clusters which on a secondary level discriminates between instances from different datasets, \eg, at the bottom right orange cluster of the task in-painting (row one, second t-SNE), where the task occupies a broader region, while on the more granular level disseminating datasets (row two, second column).
Lastly, we observe when adding \emph{failure tasks}, the clusters in the t-SNE visualization show larger gaps between individual clusters as compared to vanilla \emph{TaCo}.
This hints at an advantageous effect of the failure tasks for a better distinction between task clusters that form and further between the dataset sub-clusters within them.
In the task embeddings of \emph{TaCo with failure tasks} (row one, column three), we can further analyze the proximity of different tasks, \eg, geometric transformation tasks occupy the top left of the t-SNE space, with rotation- and flip tasks.
Next to it in the top portion, we see super-resolution, denoising and the identity task in close proximity.

\begin{figure*}
    \centering
    \includegraphics[width=0.4\linewidth]{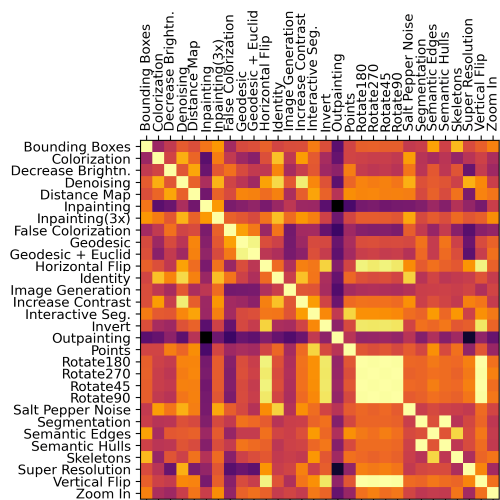}
    \includegraphics[width=0.4\linewidth]{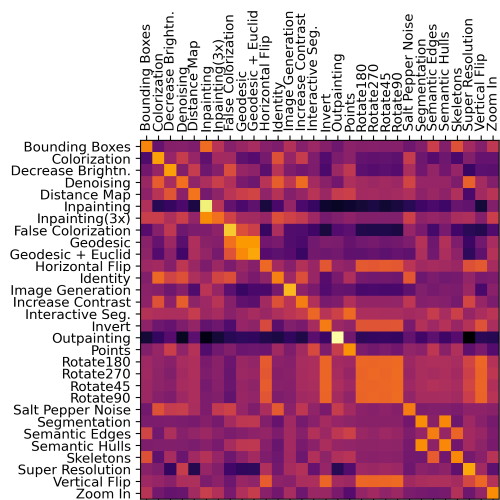}
    \caption{Task similarity matrices based on the cosine similarities between task representations of different base tasks. The right matrix is row-wise normalied.}
    \label{fig:task_similarities_both}
\end{figure*}
\subsubsection{Task Similarity Matrices}
To visualize such task similarities in more detail, we compute the cosine similarities between mean visual task representations of the $30$ tasks using the \emph{TaCo with failure tasks} model in~\Cref{fig:task_similarities_both}.
The right similarity matrix shows a row-wise normalization of the similarity values.
There, apart from confirming the similarity of geometric transformation tasks, we also see,~\eg, semantic edge detection is close to detecting boxes, central points on objects or object skeletons, indicating proximity of tasks with sparse outputs.
Further, the tasks of inpainting an image or outpainting an image, \ie, synthesizing image content at masked locations, seem to be rather distinct tasks, exhibiting an overall low similarity to all other tasks.
Tasks where the input and the output have similar image content further show larger similarity, as seen for denoising, colorization, contrast increase, denoising with salt and pepper noise or the super-resolution task.

\section{Conclusion}
\label{sec:conclusion}


In this paper, we introduced a new approach to probing the intrinsic relationships between medical vision tasks, demonstrating how task representations can be learned in a data-driven manner.
Our proposed method, Task-Contrastive Learning (TaCo), leverages task and dataset information to embed vision tasks into a shared representation space, enabling a deeper analysis of their underlying structure.
Through extensive qualitative and quantitative exploration, we uncovered how tasks relate to one another -- identifying similarities, differences, and the effects of iterative, small task alterations -- while also investigating the role of additional failure tasks in refining these representations.
Our findings provide a first step for understanding the fundamental properties of medical vision tasks, offering insights into their interconnectedness across diverse imaging modalities from computed tomography, over ultrasound to X-rays.
By focusing on the intrinsic structure of tasks, we aim to inspire further exploration into the nature of medical vision tasks.

\paragraph{Acknowledgment}
The author acknowledges support by the state of Baden-W{\"u}rttemberg
through bwHPC.
This work was performed with the help of the Large Scale Data Facility at the Karlsruhe Institute of Technology funded by the Ministry of Science, Research and the Arts Baden-W{\"u}rttemberg and by the Federal Ministry of Education and Research.
The authors gratefully acknowledge the computing time provided on the high-performance computer HoreKa by the National High-Performance Computing Center at KIT (NHR@KIT). This center is jointly supported by the Federal Ministry of Education and Research and the Ministry of Science, Research and the Arts of Baden-W{\"u}rttemberg, as part of the National High-Performance Computing (NHR) joint funding program (https://www.nhr-verein.de/en/our-partners). HoreKa is partly funded by the German Research Foundation (DFG).
This work was supported by funding from the pilot program Core-Informatics of the Helmholtz Association (HGF).

{
\bibliography{example_paper.bib}

@article{zou2024segment,
  title={Segment everything everywhere all at once},
  author={Zou, Xueyan and Yang, Jianwei and Zhang, Hao and Li, Feng and Li, Linjie and Wang, Jianfeng and Wang, Lijuan and Gao, Jianfeng and Lee, Yong Jae},
  journal={Advances in Neural Information Processing Systems},
  volume={36},
  year={2024}
}

@inproceedings{czolbe2023neuralizer,
  title={Neuralizer: General Neuroimage Analysis without Re-Training},
  author={Czolbe, Steffen and Dalca, Adrian V},
  booktitle={Proceedings of the IEEE/CVF Conference on Computer Vision and Pattern Recognition},
  pages={6217--6230},
  year={2023}
}

@article{bar2022visual,
  title={Visual prompting via image inpainting},
  author={Bar, Amir and Gandelsman, Yossi and Darrell, Trevor and Globerson, Amir and Efros, Alexei},
  journal={Advances in Neural Information Processing Systems},
  volume={35},
  pages={25005--25017},
  year={2022}
}

@inproceedings{wang2023images,
  title={Images speak in images: A generalist painter for in-context visual learning},
  author={Wang, Xinlong and Wang, Wen and Cao, Yue and Shen, Chunhua and Huang, Tiejun},
  booktitle={Proceedings of the IEEE/CVF Conference on Computer Vision and Pattern Recognition},
  pages={6830--6839},
  year={2023}
}

@article{wang2023seggpt,
  title={Seggpt: Segmenting everything in context},
  author={Wang, Xinlong and Zhang, Xiaosong and Cao, Yue and Wang, Wen and Shen, Chunhua and Huang, Tiejun},
  journal={arXiv preprint arXiv:2304.03284},
  year={2023}
}

@article{gu2024analogist,
  title={Analogist: Out-of-the-box Visual In-Context Learning with Image Diffusion Model},
  author={Gu, Zheng and Yang, Shiyuan and Liao, Jing and Huo, Jing and Gao, Yang},
  journal={ACM Transactions on Graphics (TOG)},
  volume={43},
  number={4},
  pages={1--15},
  year={2024},
  publisher={ACM New York, NY, USA}
}

@article{guo2024data,
  title={Data-efficient large vision models through sequential autoregression},
  author={Guo, Jianyuan and Hao, Zhiwei and Wang, Chengcheng and Tang, Yehui and Wu, Han and Hu, Han and Han, Kai and Xu, Chang},
  journal={arXiv preprint arXiv:2402.04841},
  year={2024}
}

@inproceedings{lu2022unified,
  title={Unified-io: A unified model for vision, language, and multi-modal tasks},
  author={Lu, Jiasen and Clark, Christopher and Zellers, Rowan and Mottaghi, Roozbeh and Kembhavi, Aniruddha},
  booktitle={The Eleventh International Conference on Learning Representations},
  year={2022}
}

@inproceedings{bai2024sequential,
  title={Sequential modeling enables scalable learning for large vision models},
  author={Bai, Yutong and Geng, Xinyang and Mangalam, Karttikeya and Bar, Amir and Yuille, Alan L and Darrell, Trevor and Malik, Jitendra and Efros, Alexei A},
  booktitle={Proceedings of the IEEE/CVF Conference on Computer Vision and Pattern Recognition},
  pages={22861--22872},
  year={2024}
}

@inproceedings{he2022masked,
  title={Masked autoencoders are scalable vision learners},
  author={He, Kaiming and Chen, Xinlei and Xie, Saining and Li, Yanghao and Doll{\'a}r, Piotr and Girshick, Ross},
  booktitle={Proceedings of the IEEE/CVF conference on computer vision and pattern recognition},
  pages={16000--16009},
  year={2022}
}

@article{gidaris2018unsupervised,
  title={Unsupervised representation learning by predicting image rotations},
  author={Gidaris, Spyros and Singh, Praveer and Komodakis, Nikos},
  journal={arXiv preprint arXiv:1803.07728},
  year={2018}
}

@inproceedings{noroozi2016unsupervised,
  title={Unsupervised learning of visual representations by solving jigsaw puzzles},
  author={Noroozi, Mehdi and Favaro, Paolo},
  booktitle={European conference on computer vision},
  pages={69--84},
  year={2016},
  organization={Springer}
}

@inproceedings{zhang2016colorful,
  title={Colorful image colorization},
  author={Zhang, Richard and Isola, Phillip and Efros, Alexei A},
  booktitle={Computer Vision--ECCV 2016: 14th European Conference, Amsterdam, The Netherlands, October 11-14, 2016, Proceedings, Part III 14},
  pages={649--666},
  year={2016},
  organization={Springer}
}

@inproceedings{chopra2005learning,
  title={Learning a similarity metric discriminatively, with application to face verification},
  author={Chopra, Sumit and Hadsell, Raia and LeCun, Yann},
  booktitle={2005 IEEE computer society conference on computer vision and pattern recognition (CVPR'05)},
  volume={1},
  pages={539--546},
  year={2005},
  organization={IEEE}
}

@inproceedings{chen2020simple,
  title={A simple framework for contrastive learning of visual representations},
  author={Chen, Ting and Kornblith, Simon and Norouzi, Mohammad and Hinton, Geoffrey},
  booktitle={International conference on machine learning},
  pages={1597--1607},
  year={2020},
  organization={PMLR}
}

@inproceedings{yeh2022decoupled,
  title={Decoupled contrastive learning},
  author={Yeh, Chun-Hsiao and Hong, Cheng-Yao and Hsu, Yen-Chi and Liu, Tyng-Luh and Chen, Yubei and LeCun, Yann},
  booktitle={European conference on computer vision},
  pages={668--684},
  year={2022},
  organization={Springer}
}

@article{chen2020improved,
  title={Improved baselines with momentum contrastive learning},
  author={Chen, Xinlei and Fan, Haoqi and Girshick, Ross and He, Kaiming},
  journal={arXiv preprint arXiv:2003.04297},
  year={2020}
}

@article{oord2018representation,
  title={Representation learning with contrastive predictive coding},
  author={Oord, Aaron van den and Li, Yazhe and Vinyals, Oriol},
  journal={arXiv preprint arXiv:1807.03748},
  year={2018}
}

@article{khosla2020supervised,
  title={Supervised contrastive learning},
  author={Khosla, Prannay and Teterwak, Piotr and Wang, Chen and Sarna, Aaron and Tian, Yonglong and Isola, Phillip and Maschinot, Aaron and Liu, Ce and Krishnan, Dilip},
  journal={Advances in neural information processing systems},
  volume={33},
  pages={18661--18673},
  year={2020}
}

@article{heinrich2021whole,
  title={Whole-cell organelle segmentation in volume electron microscopy},
  author={Heinrich, Larissa and Bennett, Davis and Ackerman, David and Park, Woohyun and Bogovic, John and Eckstein, Nils and Petruncio, Alyson and Clements, Jody and Pang, Song and Xu, C Shan and others},
  journal={Nature},
  volume={599},
  number={7883},
  pages={141--146},
  year={2021},
  publisher={Nature Publishing Group UK London}
}

@article{ma2024segment,
  title={Segment anything in medical images},
  author={Ma, Jun and He, Yuting and Li, Feifei and Han, Lin and You, Chenyu and Wang, Bo},
  journal={Nature Communications},
  volume={15},
  number={1},
  pages={654},
  year={2024},
  publisher={Nature Publishing Group UK London}
}

@misc{ilharco_gabriel_2021_5143773,
  author       = {Ilharco, Gabriel and
                  Wortsman, Mitchell and
                  Wightman, Ross and
                  Gordon, Cade and
                  Carlini, Nicholas and
                  Taori, Rohan and
                  Dave, Achal and
                  Shankar, Vaishaal and
                  Namkoong, Hongseok and
                  Miller, John and
                  Hajishirzi, Hannaneh and
                  Farhadi, Ali and
                  Schmidt, Ludwig},
  title        = {OpenCLIP},
  year         = {2021},
  publisher    = {Zenodo},
  version      = {0.1},
  doi          = {10.5281/zenodo.5143773},
  url          = {https://doi.org/10.5281/zenodo.5143773}
}

@inproceedings{cherti2023reproducible,
  title={Reproducible scaling laws for contrastive language-image learning},
  author={Cherti, Mehdi and Beaumont, Romain and Wightman, Ross and Wortsman, Mitchell and Ilharco, Gabriel and Gordon, Cade and Schuhmann, Christoph and Schmidt, Ludwig and Jitsev, Jenia},
  booktitle={Proceedings of the IEEE/CVF Conference on Computer Vision and Pattern Recognition},
  pages={2818--2829},
  year={2023}
}

@inproceedings{Radford2021LearningTV,
  title={Learning Transferable Visual Models From Natural Language Supervision},
  author={Alec Radford and Jong Wook Kim and Chris Hallacy and A. Ramesh and Gabriel Goh and Sandhini Agarwal and Girish Sastry and Amanda Askell and Pamela Mishkin and Jack Clark and Gretchen Krueger and Ilya Sutskever},
  booktitle={ICML},
  year={2021}
}

@inproceedings{schuhmann2022laionb,
  title={{LAION}-5B: An open large-scale dataset for training next generation image-text models},
  author={Christoph Schuhmann and
          Romain Beaumont and
          Richard Vencu and
          Cade W Gordon and
          Ross Wightman and
          Mehdi Cherti and
          Theo Coombes and
          Aarush Katta and
          Clayton Mullis and
          Mitchell Wortsman and
          Patrick Schramowski and
          Srivatsa R Kundurthy and
          Katherine Crowson and
          Ludwig Schmidt and
          Robert Kaczmarczyk and
          Jenia Jitsev},
  booktitle={Thirty-sixth Conference on Neural Information Processing Systems Datasets and Benchmarks Track},
  year={2022},
  url={https://openreview.net/forum?id=M3Y74vmsMcY}
}

@inproceedings{he2016deep,
  title={Deep residual learning for image recognition},
  author={He, Kaiming and Zhang, Xiangyu and Ren, Shaoqing and Sun, Jian},
  booktitle={Proceedings of the IEEE conference on computer vision and pattern recognition},
  pages={770--778},
  year={2016}
}

@inproceedings{wortsman2022robust,
  title={Robust fine-tuning of zero-shot models},
  author={Wortsman, Mitchell and Ilharco, Gabriel and Kim, Jong Wook and Li, Mike and Kornblith, Simon and Roelofs, Rebecca and Lopes, Raphael Gontijo and Hajishirzi, Hannaneh and Farhadi, Ali and Namkoong, Hongseok and others},
  booktitle={Proceedings of the IEEE/CVF conference on computer vision and pattern recognition},
  pages={7959--7971},
  year={2022}
}

@article{dosovitskiy2020image,
  title={AN IMAGE IS WORTH 16X16 WORDS: TRANSFORMERS FOR IMAGE RECOGNITION AT SCALE},
  author={Dosovitskiy, Alexey and Beyer, Lucas and Kolesnikov, Alexander and Weissenborn, Dirk and Zhai, Xiaohua and Unterthiner, Thomas and Dehghani, Mostafa and Minderer, Matthias and Heigold, Georg and Gelly, Sylvain and others},
  journal={arXiv preprint arXiv:2010.11929},
  year={2020}
}

@inproceedings{deng2009imagenet,
  title={Imagenet: A large-scale hierarchical image database},
  author={Deng, Jia and Dong, Wei and Socher, Richard and Li, Li-Jia and Li, Kai and Fei-Fei, Li},
  booktitle={2009 IEEE conference on computer vision and pattern recognition},
  pages={248--255},
  year={2009},
  organization={Ieee}
}

@article{van2008visualizing,
  title={Visualizing data using t-SNE.},
  author={Van der Maaten, Laurens and Hinton, Geoffrey},
  journal={Journal of machine learning research},
  volume={9},
  number={11},
  year={2008}
}

@inproceedings{achille2019task2vec,
  title={Task2vec: Task embedding for meta-learning},
  author={Achille, Alessandro and Lam, Michael and Tewari, Rahul and Ravichandran, Avinash and Maji, Subhransu and Fowlkes, Charless C and Soatto, Stefano and Perona, Pietro},
  booktitle={Proceedings of the IEEE/CVF international conference on computer vision},
  pages={6430--6439},
  year={2019}
}

@inproceedings{zamir2018taskonomy,
  title={Taskonomy: Disentangling task transfer learning},
  author={Zamir, Amir R and Sax, Alexander and Shen, William and Guibas, Leonidas J and Malik, Jitendra and Savarese, Silvio},
  booktitle={Proceedings of the IEEE conference on computer vision and pattern recognition},
  pages={3712--3722},
  year={2018}
}

@inproceedings{hojel2025finding,
  title={Finding visual task vectors},
  author={Hojel, Alberto and Bai, Yutong and Darrell, Trevor and Globerson, Amir and Bar, Amir},
  booktitle={European Conference on Computer Vision},
  pages={257--273},
  year={2025},
  organization={Springer}
}

@article{donahue2016adversarial,
  title={Adversarial feature learning},
  author={Donahue, Jeff and Kr{\"a}henb{\"u}hl, Philipp and Darrell, Trevor},
  journal={arXiv preprint arXiv:1605.09782},
  year={2016}
}

@inproceedings{brempong2022denoising,
  title={Denoising pretraining for semantic segmentation},
  author={Brempong, Emmanuel Asiedu and Kornblith, Simon and Chen, Ting and Parmar, Niki and Minderer, Matthias and Norouzi, Mohammad},
  booktitle={Proceedings of the IEEE/CVF conference on computer vision and pattern recognition},
  pages={4175--4186},
  year={2022}
}

@inproceedings{pathak2016context,
  title={Context encoders: Feature learning by inpainting},
  author={Pathak, Deepak and Krahenbuhl, Philipp and Donahue, Jeff and Darrell, Trevor and Efros, Alexei A},
  booktitle={Proceedings of the IEEE conference on computer vision and pattern recognition},
  pages={2536--2544},
  year={2016}
}

@incollection{NEURIPS2019_9015,
  title     = {PyTorch: An Imperative Style, High-Performance Deep Learning Library},
  author    = {Paszke, Adam and Gross, Sam and Massa, Francisco and Lerer, Adam and Bradbury, James and Chanan, Gregory and Killeen, Trevor and Lin, Zeming and Gimelshein, Natalia and Antiga, Luca and Desmaison, Alban and Kopf, Andreas and Yang, Edward and DeVito, Zachary and Raison, Martin and Tejani, Alykhan and Chilamkurthy, Sasank and Steiner, Benoit and Fang, Lu and Bai, Junjie and Chintala, Soumith},
  booktitle = {Advances in Neural Information Processing Systems 32},
  pages     = {8024--8035},
  year      = {2019},
  publisher = {Curran Associates, Inc.},
}

@inproceedings{mckinney2010data,
  title={Data structures for statistical computing in python},
  author={McKinney, Wes and others},
  booktitle={Proceedings of the 9th Python in Science Conference},
  volume={445},
  pages={51--56},
  year={2010},
  organization={Austin, TX}
}

@article{scikit-learn,
  title={Scikit-learn: Machine Learning in {P}ython},
  author={Pedregosa, F. and Varoquaux, G. and Gramfort, A. and Michel, V.
          and Thirion, B. and Grisel, O. and Blondel, M. and Prettenhofer, P.
          and Weiss, R. and Dubourg, V. and Vanderplas, J. and Passos, A. and
          Cournapeau, D. and Brucher, M. and Perrot, M. and Duchesnay, E.},
  journal={Journal of Machine Learning Research},
  volume={12},
  pages={2825--2830},
  year={2011}
}

@article{ji2022amos,
  title={AMOS: A Large-Scale Abdominal Multi-Organ Benchmark for Versatile Medical Image Segmentation},
  author={Ji, Yuanfeng and Bai, Haotian and Yang, Jie and Ge, Chongjian and Zhu, Ye and Zhang, Ruimao and Li, Zhen and Zhang, Lingyan and Ma, Wanling and Wan, Xiang and others},
  journal={arXiv preprint arXiv:2206.08023},
  year={2022}
}

@article{gatidis2022whole,
  title={A whole-body FDG-PET/CT dataset with manually annotated tumor lesions},
  author={Gatidis, Sergios and Hepp, Tobias and Fr{\"u}h, Marcel and La Foug{\`e}re, Christian and Nikolaou, Konstantin and Pfannenberg, Christina and Sch{\"o}lkopf, Bernhard and K{\"u}stner, Thomas and Cyran, Clemens and Rubin, Daniel},
  journal={Scientific Data},
  volume={9},
  number={1},
  pages={601},
  year={2022},
  publisher={Nature Publishing Group UK London}
}

@article{menze2014multimodal,
  title={The multimodal brain tumor image segmentation benchmark (BRATS)},
  author={Menze, Bjoern H and Jakab, Andras and Bauer, Stefan and Kalpathy-Cramer, Jayashree and Farahani, Keyvan and Kirby, Justin and Burren, Yuliya and Porz, Nicole and Slotboom, Johannes and Wiest, Roland and others},
  journal={IEEE transactions on medical imaging},
  volume={34},
  number={10},
  pages={1993--2024},
  year={2014},
  publisher={IEEE}
}

@article{bakas2018identifying,
  title={Identifying the best machine learning algorithms for brain tumor segmentation, progression assessment, and overall survival prediction in the BRATS challenge},
  author={Bakas, Spyridon and Reyes, Mauricio and Jakab, Andras and Bauer, Stefan and Rempfler, Markus and Crimi, Alessandro and Shinohara, Russell Takeshi and Berger, Christoph and Ha, Sung Min and Rozycki, Martin and others},
  journal={arXiv preprint arXiv:1811.02629},
  year={2018}
}

@article{bakas2017advancing,
  title={Advancing the cancer genome atlas glioma MRI collections with expert segmentation labels and radiomic features},
  author={Bakas, Spyridon and Akbari, Hamed and Sotiras, Aristeidis and Bilello, Michel and Rozycki, Martin and Kirby, Justin S and Freymann, John B and Farahani, Keyvan and Davatzikos, Christos},
  journal={Scientific data},
  volume={4},
  number={1},
  pages={1--13},
  year={2017},
  publisher={Nature Publishing Group}
}

@article{bakas10segmentation,
  title={Segmentation labels and radiomic features for the pre-operative scans of the TCGA-GBM collection (2017)},
  author={Bakas, Spyridon and Akbari, Hamed and Sotiras, Aristeidis and Bilello, Michel and Rozycki, Martin and Kirby, Justin and Freymann, John and Farahani, Keyvan and Davatzikos, Christos},
  journal={DOI: https://doi. org/10.7937 K},
  volume={9},
  year ={2017}
}

@misc{bakas2017segmentation,
  title={Segmentation labels and radiomic features for the pre-operative scans of the TCGA-LGG collection [Data Set]. The Cancer Imaging Archive},
  author={Bakas, Spyridon and Akbari, Hamed and Sotiras, Aristeidis and Bilello, Michel and Rozycki, Martin and Kirby, JS and Freymann, JB and Farahani, Keyvan and Davatzikos, Christos},
  year={2017},
  publisher={Version}
}

@article{khaled2021categorized,
  title={Categorized digital database for low energy and subtracted contrast enhanced spectral mammography images},
  author={Khaled, R and others},
  journal={The Cancer Imaging Archive},
  year={2021}
}

@article{mayr2023cervical,
  title={Cervical Cancer—Tumor Heterogeneity: Serial Functional and Molecular Imaging Across the Radiation Therapy Course in Advanced Cervical Cancer (Version 1)},
  author={Mayr, Nina and Yuh, W and Bowen, Stephen and Harkenrider, Matthew and Knopp, M and Lee, E and Leung, Eric and Lo, S and Small Jr, William and Wolfson, H},
  journal={The Cancer Imaging Archive},
  year={2023}
}

@article{jaeger2013automatic,
  title={Automatic tuberculosis screening using chest radiographs},
  author={Jaeger, Stefan and Karargyris, Alexandros and Candemir, Sema and Folio, Les and Siegelman, Jenifer and Callaghan, Fiona and Xue, Zhiyun and Palaniappan, Kannappan and Singh, Rahul K and Antani, Sameer and others},
  journal={IEEE transactions on medical imaging},
  volume={33},
  number={2},
  pages={233--245},
  year={2013},
  publisher={IEEE}
}

@article{candemir2013lung,
  title={Lung segmentation in chest radiographs using anatomical atlases with nonrigid registration},
  author={Candemir, Sema and Jaeger, Stefan and Palaniappan, Kannappan and Musco, Jonathan P and Singh, Rahul K and Xue, Zhiyun and Karargyris, Alexandros and Antani, Sameer and Thoma, George and McDonald, Clement J},
  journal={IEEE transactions on medical imaging},
  volume={33},
  number={2},
  pages={577--590},
  year={2013},
  publisher={IEEE}
}

@article{ma2021toward,
  title={Toward data-efficient learning: A benchmark for COVID-19 CT lung and infection segmentation},
  author={Ma, Jun and Wang, Yixin and An, Xingle and Ge, Cheng and Yu, Ziqi and Chen, Jianan and Zhu, Qiongjie and Dong, Guoqiang and He, Jian and He, Zhiqiang and others},
  journal={Medical physics},
  volume={48},
  number={3},
  pages={1197--1210},
  year={2021},
  publisher={Wiley Online Library}
}

@article{tahir2021covid,
  title={COVID-19 infection localization and severity grading from chest X-ray images},
  author={Tahir, Anas M and Chowdhury, Muhammad EH and Khandakar, Amith and Rahman, Tawsifur and Qiblawey, Yazan and Khurshid, Uzair and Kiranyaz, Serkan and Ibtehaz, Nabil and Rahman, M Sohel and Al-Maadeed, Somaya and others},
  journal={Computers in biology and medicine},
  volume={139},
  pages={105002},
  year={2021},
  publisher={Elsevier}
}

@article{rahman2021exploring,
  title={Exploring the effect of image enhancement techniques on COVID-19 detection using chest X-ray images},
  author={Rahman, Tawsifur and Khandakar, Amith and Qiblawey, Yazan and Tahir, Anas and Kiranyaz, Serkan and Kashem, Saad Bin Abul and Islam, Mohammad Tariqul and Al Maadeed, Somaya and Zughaier, Susu M and Khan, Muhammad Salman and others},
  journal={Computers in biology and medicine},
  volume={132},
  pages={104319},
  year={2021},
  publisher={Elsevier}
}

@article{degerli2021covid,
  title={COVID-19 infection map generation and detection from chest X-ray images},
  author={Degerli, Aysen and Ahishali, Mete and Yamac, Mehmet and Kiranyaz, Serkan and Chowdhury, Muhammad EH and Hameed, Khalid and Hamid, Tahir and Mazhar, Rashid and Gabbouj, Moncef},
  journal={Health information science and systems},
  volume={9},
  number={1},
  pages={15},
  year={2021},
  publisher={Springer}
}

@article{chowdhury2020can,
  title={Can AI help in screening viral and COVID-19 pneumonia?},
  author={Chowdhury, Muhammad EH and Rahman, Tawsifur and Khandakar, Amith and Mazhar, Rashid and Kadir, Muhammad Abdul and Mahbub, Zaid Bin and Islam, Khandakar Reajul and Khan, Muhammad Salman and Iqbal, Atif and Al Emadi, Nasser and others},
  journal={Ieee Access},
  volume={8},
  pages={132665--132676},
  year={2020},
  publisher={IEEE}
}

@misc{anas_m__tahir_muhammad_e__h__chowdhury_yazan_qiblawey_amith_khandakar_tawsifur_rahman_serkan_kiranyaz_uzair_khurshid_nabil_ibtehaz_sakib_mahmud_maymouna_ezeddin_2022,
	title={COVID-QU-Ex Dataset},
	url={https://www.kaggle.com/dsv/3122958},
	DOI={10.34740/KAGGLE/DSV/3122958},
	publisher={Kaggle},
	author={Anas M. Tahir and Muhammad E. H. Chowdhury and Yazan Qiblawey and Amith Khandakar and Tawsifur Rahman and Serkan Kiranyaz and Uzair Khurshid and Nabil Ibtehaz and Sakib Mahmud and Maymouna Ezeddin},
	year={2022}
}

@article{roth2022rapid,
  title={Rapid artificial intelligence solutions in a pandemic—The COVID-19-20 Lung CT Lesion Segmentation Challenge},
  author={Roth, Holger R and Xu, Ziyue and Tor-D{\'\i}ez, Carlos and Jacob, Ramon Sanchez and Zember, Jonathan and Molto, Jose and Li, Wenqi and Xu, Sheng and Turkbey, Baris and Turkbey, Evrim and others},
  journal={Medical image analysis},
  volume={82},
  pages={102605},
  year={2022},
  publisher={Elsevier}
}

@misc{ct_images,
    title = {CT Images in COVID-19 [Data set].} ,
    author = {An, P. and Xu, S. and Harmon, S. A. and Turkbey, E. B. and Sanford, T. H. and Amalou, A. and Kassin, M. and Varble, N. and Blain, M. and Anderson, V. and Patella, F.and Carrafiello, G. and Turkbey, B. T. and Wood, B. J.},
    DOI = {https://doi.org/10.7937/TCIA.2020.GQRY-NC81},
    publisher = {The Cancer Imaging Archive},
    year = {2020}
}

@article{dorent2023crossmoda,
  title={CrossMoDA 2021 challenge: Benchmark of cross-modality domain adaptation techniques for vestibular schwannoma and cochlea segmentation},
  author={Dorent, Reuben and Kujawa, Aaron and Ivory, Marina and Bakas, Spyridon and Rieke, Nicola and Joutard, Samuel and Glocker, Ben and Cardoso, Jorge and Modat, Marc and Batmanghelich, Kayhan and others},
  journal={Medical Image Analysis},
  volume={83},
  pages={102628},
  year={2023},
  publisher={Elsevier}
}

@article{van2018automated,
  title={Automated measurement of fetal head circumference using 2D ultrasound images},
  author={van den Heuvel, Thomas LA and de Bruijn, Dagmar and de Korte, Chris L and Ginneken, Bram van},
  journal={PloS one},
  volume={13},
  number={8},
  pages={e0200412},
  year={2018},
  publisher={Public Library of Science San Francisco, CA USA}
}

@article{simpson2019large,
  title={A large annotated medical image dataset for the development and evaluation of segmentation algorithms},
  author={Simpson, Amber L and Antonelli, Michela and Bakas, Spyridon and Bilello, Michel and Farahani, Keyvan and Van Ginneken, Bram and Kopp-Schneider, Annette and Landman, Bennett A and Litjens, Geert and Menze, Bjoern and others},
  journal={arXiv preprint arXiv:1902.09063},
  year={2019}
}

@article{ahmed2022deep,
  title={Deep learning based automated detection of intraretinal cystoid fluid},
  author={Ahmed, Zeeshan and Panhwar, Shahbaz Qamar and Baqai, Attiya and Umrani, Fahim Aziz and Ahmed, Munawar and Khan, Arbaaz},
  journal={International Journal of Imaging Systems and Technology},
  volume={32},
  number={3},
  pages={902--917},
  year={2022},
  publisher={Wiley Online Library}
}

@article{hernandez2022isles,
  title={ISLES 2022: A multi-center magnetic resonance imaging stroke lesion segmentation dataset},
  author={Hernandez Petzsche, Moritz R and de la Rosa, Ezequiel and Hanning, Uta and Wiest, Roland and Valenzuela, Waldo and Reyes, Mauricio and Meyer, Maria and Liew, Sook-Lei and Kofler, Florian and Ezhov, Ivan and others},
  journal={Scientific data},
  volume={9},
  number={1},
  pages={762},
  year={2022},
  publisher={Nature Publishing Group UK London}
}

@inproceedings{jha2019kvasir,
  title={Kvasir-seg: A segmented polyp dataset},
  author={Jha, Debesh and Smedsrud, Pia H and Riegler, Michael A and Halvorsen, P{\aa}l and De Lange, Thomas and Johansen, Dag and Johansen, H{\aa}vard D},
  booktitle={International conference on multimedia modeling},
  pages={451--462},
  year={2019},
  organization={Springer}
}

@inproceedings{pogorelov2017kvasir,
  title={Kvasir: A multi-class image dataset for computer aided gastrointestinal disease detection},
  author={Pogorelov, Konstantin and Randel, Kristin Ranheim and Griwodz, Carsten and Eskeland, Sigrun Losada and de Lange, Thomas and Johansen, Dag and Spampinato, Concetto and Dang-Nguyen, Duc-Tien and Lux, Mathias and Schmidt, Peter Thelin and others},
  booktitle={Proceedings of the 8th ACM on Multimedia Systems Conference},
  pages={164--169},
  year={2017}
}

@article{antonelli2022medical,
  title={The medical segmentation decathlon},
  author={Antonelli, Michela and Reinke, Annika and Bakas, Spyridon and Farahani, Keyvan and Kopp-Schneider, Annette and Landman, Bennett A and Litjens, Geert and Menze, Bjoern and Ronneberger, Olaf and Summers, Ronald M and others},
  journal={Nature communications},
  volume={13},
  number={1},
  pages={4128},
  year={2022},
  publisher={Nature Publishing Group UK London}
}

@article{maqbool2020m2caiseg,
  title={m2caiseg: Semantic segmentation of laparoscopic images using convolutional neural networks},
  author={Maqbool, Salman and Riaz, Aqsa and Sajid, Hasan and Hasan, Osman},
  journal={arXiv preprint arXiv:2008.10134},
  year={2020}
}

@article{NeurIPS-CellSeg,
      title = {The Multi-modality Cell Segmentation Challenge: Towards Universal Solutions},
      author = {Jun Ma and Ronald Xie and Shamini Ayyadhury and Cheng Ge and Anubha Gupta and Ritu Gupta and Song Gu and Yao Zhang and Gihun Lee and Joonkee Kim and Wei Lou and Haofeng Li and Eric Upschulte and Timo Dickscheid and José Guilherme de Almeida and Yixin Wang and Lin Han and Xin Yang and Marco Labagnara and Vojislav Gligorovski and Maxime Scheder and Sahand Jamal Rahi and Carly Kempster and Alice Pollitt and Leon Espinosa and Tâm Mignot and Jan Moritz Middeke and Jan-Niklas Eckardt and Wangkai Li and Zhaoyang Li and Xiaochen Cai and Bizhe Bai and Noah F. Greenwald and David Van Valen and Erin Weisbart and Beth A. Cimini and Trevor Cheung and Oscar Brück and Gary D. Bader and Bo Wang},
      journal = {Nature Methods},
      volume={21},
      pages={1103–1113},
      year = {2024},
      doi = {https://doi.org/10.1038/s41592-024-02233-6}
  }

@misc{siim-acr-pneumothorax-segmentation,
    author = {Anna Zawacki and Carol Wu and George Shih and Julia Elliott and Mikhail Fomitchev and Mohannad Hussain and ParasLakhani and Phil Culliton and Shunxing Bao},
    title = {SIIM-ACR Pneumothorax Segmentation},
    year = {2019},
    howpublished = {\url{https://kaggle.com/competitions/siim-acr-pneumothorax-segmentation}},
    note = {Kaggle}
}

@article{fedorov2018annotated,
  title={An annotated test-retest collection of prostate multiparametric MRI},
  author={Fedorov, Andriy and Schwier, Michael and Clunie, David and Herz, Christian and Pieper, Steve and Kikinis, Ron and Tempany, Clare and Fennessy, Fiona},
  journal={Scientific data},
  volume={5},
  number={1},
  pages={1--13},
  year={2018},
  publisher={Nature Publishing Group}
}

@inproceedings{zukic2014robust,
  title={Robust detection and segmentation for diagnosis of vertebral diseases using routine MR images},
  author={Zuki{\'c}, D{\v{z}}enan and Vlas{\'a}k, Ale{\v{s}} and Egger, Jan and Ho{\v{r}}{\'\i}nek, Daniel and Nimsky, Christopher and Kolb, Andreas},
  booktitle={Computer Graphics Forum},
  volume={33},
  pages={190--204},
  year={2014},
  organization={Wiley Online Library}
}

@misc{lctsc,
    author = {Yang, J. and Sharp, G. and Veeraraghavan, H. and Van Elmpt, W. and Dekker, A. and Lustberg, T. and Gooding, M.},
    title = {Data from Lung CT Segmentation Challenge (LCTSC) (Version 3) [Data set].},
    year = {2017},
    publisher = {The Cancer Imaging Archive},
    DOI = {https://doi.org/10.7937/K9/TCIA.2017.3R3FVZ08}
}

@article{kuijf2019standardized,
  title={Standardized assessment of automatic segmentation of white matter hyperintensities and results of the WMH segmentation challenge},
  author={Kuijf, Hugo J and Biesbroek, J Matthijs and De Bresser, Jeroen and Heinen, Rutger and Andermatt, Simon and Bento, Mariana and Berseth, Matt and Belyaev, Mikhail and Cardoso, M Jorge and Casamitjana, Adria and others},
  journal={IEEE transactions on medical imaging},
  volume={38},
  number={11},
  pages={2556--2568},
  year={2019},
  publisher={IEEE}
}

@article{al2020dataset,
  title={Dataset of breast ultrasound images},
  author={Al-Dhabyani, Walid and Gomaa, Mohammed and Khaled, Hussien and Fahmy, Aly},
  journal={Data in brief},
  volume={28},
  pages={104863},
  year={2020},
  publisher={Elsevier}
}

@article{tschandl2018ham10000,
  title={The HAM10000 dataset, a large collection of multi-source dermatoscopic images of common pigmented skin lesions},
  author={Tschandl, Philipp and Rosendahl, Cliff and Kittler, Harald},
  journal={Scientific data},
  volume={5},
  number={1},
  pages={1--9},
  year={2018},
  publisher={Nature Publishing Group}
}

@inproceedings{codella2018skin,
  title={Skin lesion analysis toward melanoma detection: A challenge at the 2017 international symposium on biomedical imaging (isbi), hosted by the international skin imaging collaboration (isic)},
  author={Codella, Noel CF and Gutman, David and Celebi, M Emre and Helba, Brian and Marchetti, Michael A and Dusza, Stephen W and Kalloo, Aadi and Liopyris, Konstantinos and Mishra, Nabin and Kittler, Harald and others},
  booktitle={2018 IEEE 15th international symposium on biomedical imaging (ISBI 2018)},
  pages={168--172},
  year={2018},
  organization={IEEE}
}

@article{codella2019skin,
  title={Skin lesion analysis toward melanoma detection 2018: A challenge hosted by the international skin imaging collaboration (isic)},
  author={Codella, Noel and Rotemberg, Veronica and Tschandl, Philipp and Celebi, M Emre and Dusza, Stephen and Gutman, David and Helba, Brian and Kalloo, Aadi and Liopyris, Konstantinos and Marchetti, Michael and others},
  journal={arXiv preprint arXiv:1902.03368},
  year={2019}
}

@article{anis2025limitations,
  title={On the limitations of vision-language models in understanding image transforms},
  author={Anis, Ahmad Mustafa and Ali, Hasnain and Sarfraz, Saquib},
  journal={arXiv preprint arXiv:2503.09837},
  year={2025}
}
\bibliographystyle{abbrvnat}




\newpage
\appendix
\section{Limitations and Future Directions}
\label{sec:limitations}
In this work, we make the first step for learning task representations in a data-driven fashion and explore these representation in depth.
Yet, what we could not explore in this work as the scope would have been too broad is how we can make practical use of the learned task representations.
For example, it would be very interesting to explore if choosing a pre-training dataset for a new task based on a high task representation similarity may lead to a better transfer learning performance.
Similarly, another practical use, which we did not explore yet is the use of our task representations as a way to convey information about a task to be solved to a network.
This is a very natural exploration in the framework of in-context learning.
There, networks need to be given information about a task to be addressed, which is often done with few-shot examples, but which could naturally be done with our learned task representations as well.
We did not explore such practical uses of the \emph{TaCo} representations, as we primarily set the focus on validating that informative task representations can be learned end-to-end and that these representations convey fundamental task properties, such as, iteratively changing task definitions leading to small steps in traversing the representation space, or, an arising similarity among tasks that share fundamental properties such as sparse outputs or that they are based on geometric transformations.
While the methodology we propose is general and can be applied to learn task representations for natural images, here, we focused on the medical setting, as it offers a good test-bed with a high diversity in imaging modalities (\eg, MRI, CT, Xray, OCT, Ultrasound, \etc) as well as coarse- and fine-grained structures of interest (\eg, large Lungs in Xray, thin and small cell organelles in microscopy).
Exploring \emph{TaCo} in the natural image domain would be an exiting research direction, in this work we limit the scope to primarily focus on medical imaging.


\section{Example Images for the Utilized 39 Datasets}
\label{sec:appendix_exampleimages}

In~\Cref{tab:example_images}, we display an example image for each of the 39 utilized datasets which are contained in the MedSAM~\cite{ma2024segment} and Openorganelle~\cite{heinrich2021whole} dataset collections.
The  diverse image modalities and different shapes of objects of interest in these dataset are clearly visible.
Further, we provide additional meta information on the datasets in~\Cref{tab:dataset_license}.

\begin{table}[ht]
\small
\centering
\renewcommand{\arraystretch}{1.3}
\setlength{\extrarowheight}{0pt}
\caption{Example images and segmentations of the datasets from MedSAM~\cite{ma2024segment} and OpenOrganelle~\cite{heinrich2021whole}. Here, we highlight unseen datasets through yellow tinted tiles.}
\label{tab:example_images}
\begin{tabular}{c|c|c|c|c}
\toprule
\parbox[t]{0.17\linewidth}{\centering AMOSMR\\[0.1cm]\begin{minipage}{0.37\linewidth}\includegraphics[width=\linewidth]{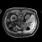}\end{minipage}\hspace{0.04\linewidth}\begin{minipage}{0.37\linewidth}\includegraphics[width=\linewidth]{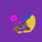}\end{minipage}\\[0.1cm]} & 

\parbox[t]{0.17\linewidth}{\centering autoPET\\[0.1cm]\begin{minipage}{0.37\linewidth}\includegraphics[width=\linewidth]{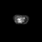}\end{minipage}\hspace{0.04\linewidth}\begin{minipage}{0.37\linewidth}\includegraphics[width=\linewidth]{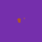}\end{minipage}\\[0.1cm]} & 

\parbox[t]{0.17\linewidth}{\centering BraTS T1\\[0.1cm]\begin{minipage}{0.37\linewidth}\includegraphics[width=\linewidth]{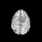}\end{minipage}\hspace{0.04\linewidth}\begin{minipage}{0.37\linewidth}\includegraphics[width=\linewidth]{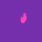}\end{minipage}\\[0.1cm]} & 

\parbox[t]{0.17\linewidth}{\centering BraTS T1CE\\[0.1cm]\begin{minipage}{0.37\linewidth}\includegraphics[width=\linewidth]{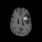}\end{minipage}\hspace{0.04\linewidth}\begin{minipage}{0.37\linewidth}\includegraphics[width=\linewidth]{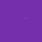}\end{minipage}\\[0.1cm]} &

\parbox[t]{0.17\linewidth}{\centering CDD-CESM\\[0.1cm]\begin{minipage}{0.37\linewidth}\includegraphics[width=\linewidth]{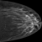}\end{minipage}\hspace{0.04\linewidth}\begin{minipage}{0.37\linewidth}\includegraphics[width=\linewidth]{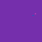}\end{minipage}\\[0.1cm]} \\

\midrule

\parbox[t]{0.17\linewidth}{\centering Cervical Cancer\\[0.1cm]\begin{minipage}{0.37\linewidth}\includegraphics[width=\linewidth]{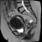}\end{minipage}\hspace{0.04\linewidth}\begin{minipage}{0.37\linewidth}\includegraphics[width=\linewidth]{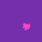}\end{minipage}\\[0.1cm]} & 

\parbox[t]{0.17\linewidth}{\centering Chest-Xray-M.-L.\\[0.1cm]\begin{minipage}{0.37\linewidth}\includegraphics[width=\linewidth]{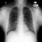}\end{minipage}\hspace{0.04\linewidth}\begin{minipage}{0.37\linewidth}\includegraphics[width=\linewidth]{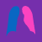}\end{minipage}\\[0.1cm]} & 

\parbox[t]{0.17\linewidth}{\centering COVID-CT-Seg.\\[0.1cm]\begin{minipage}{0.37\linewidth}\includegraphics[width=\linewidth]{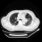}\end{minipage}\hspace{0.04\linewidth}\begin{minipage}{0.37\linewidth}\includegraphics[width=\linewidth]{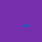}\end{minipage}\\[0.1cm]}  &

\parbox[t]{0.17\linewidth}{\centering COVID-QU-Ex-inf.\\[0.1cm]\begin{minipage}{0.37\linewidth}\includegraphics[width=\linewidth]{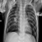}\end{minipage}\hspace{0.04\linewidth}\begin{minipage}{0.37\linewidth}\includegraphics[width=\linewidth]{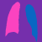}\end{minipage}\\[0.1cm]} &

\parbox[t]{0.17\linewidth}{\centering COVID-QU-Ex-L.\\[0.1cm]\begin{minipage}{0.37\linewidth}\includegraphics[width=\linewidth]{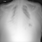}\end{minipage}\hspace{0.04\linewidth}\begin{minipage}{0.37\linewidth}\includegraphics[width=\linewidth]{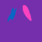}\end{minipage}\\[0.1cm]} \\

\midrule 

\parbox[t]{0.17\linewidth}{\centering COVID19-CT-Segb.\\[0.1cm]\begin{minipage}{0.37\linewidth}\includegraphics[width=\linewidth]{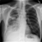}\end{minipage}\hspace{0.04\linewidth}\begin{minipage}{0.37\linewidth}\includegraphics[width=\linewidth]{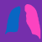}\end{minipage}\\[0.1cm]} & 

\parbox[t]{0.17\linewidth}{\centering crossmoda\\[0.1cm]\begin{minipage}{0.37\linewidth}\includegraphics[width=\linewidth]{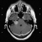}\end{minipage}\hspace{0.04\linewidth}\begin{minipage}{0.37\linewidth}\includegraphics[width=\linewidth]{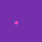}\end{minipage}\\[0.1cm]} &

\parbox[t]{0.17\linewidth}{\centering hc18\\[0.1cm]\begin{minipage}{0.37\linewidth}\includegraphics[width=\linewidth]{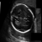}\end{minipage}\hspace{0.04\linewidth}\begin{minipage}{0.37\linewidth}\includegraphics[width=\linewidth]{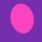}\end{minipage}\\[0.1cm]} & 

\parbox[t]{0.17\linewidth}{\centering Heart\\[0.1cm]\begin{minipage}{0.37\linewidth}\includegraphics[width=\linewidth]{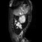}\end{minipage}\hspace{0.04\linewidth}\begin{minipage}{0.37\linewidth}\includegraphics[width=\linewidth]{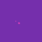}\end{minipage}\\[0.1cm]} & 

\parbox[t]{0.17\linewidth}{\centering Intraretinal-Cyst.\\[0.1cm]\begin{minipage}{0.37\linewidth}\includegraphics[width=\linewidth]{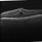}\end{minipage}\hspace{0.04\linewidth}\begin{minipage}{0.37\linewidth}\includegraphics[width=\linewidth]{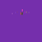}\end{minipage}\\[0.1cm]} \\

\midrule

\parbox[t]{0.17\linewidth}{\centering ISLES2022 ADC\\[0.1cm]\begin{minipage}{0.37\linewidth}\includegraphics[width=\linewidth]{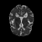}\end{minipage}\hspace{0.04\linewidth}\begin{minipage}{0.37\linewidth}\includegraphics[width=\linewidth]{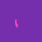}\end{minipage}\\[0.1cm]} &

\parbox[t]{0.17\linewidth}{\centering ISLES2022 DWI\\[0.1cm]\begin{minipage}{0.37\linewidth}\includegraphics[width=\linewidth]{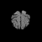}\end{minipage}\hspace{0.04\linewidth}\begin{minipage}{0.37\linewidth}\includegraphics[width=\linewidth]{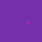}\end{minipage}\\[0.1cm]} & 

\parbox[t]{0.17\linewidth}{\centering Kvasir-SEG\\[0.1cm]\begin{minipage}{0.37\linewidth}\includegraphics[width=\linewidth]{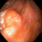}\end{minipage}\hspace{0.04\linewidth}\begin{minipage}{0.37\linewidth}\includegraphics[width=\linewidth]{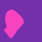}\end{minipage}\\[0.1cm]} & 

\parbox[t]{0.17\linewidth}{\centering LungLesions\\[0.1cm]\begin{minipage}{0.37\linewidth}\includegraphics[width=\linewidth]{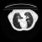}\end{minipage}\hspace{0.04\linewidth}\begin{minipage}{0.37\linewidth}\includegraphics[width=\linewidth]{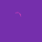}\end{minipage}\\[0.1cm]} & 

\parbox[t]{0.17\linewidth}{\centering m2caiSeg\\[0.1cm]\begin{minipage}{0.37\linewidth}\includegraphics[width=\linewidth]{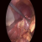}\end{minipage}\hspace{0.04\linewidth}\begin{minipage}{0.37\linewidth}\includegraphics[width=\linewidth]{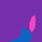}\end{minipage}\\[0.1cm]} \\

\midrule

\parbox[t]{0.17\linewidth}{\centering NeurIPS22CellSeg\\[0.1cm]\begin{minipage}{0.37\linewidth}\includegraphics[width=\linewidth]{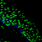}\end{minipage}\hspace{0.04\linewidth}\begin{minipage}{0.37\linewidth}\includegraphics[width=\linewidth]{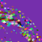}\end{minipage}\\[0.1cm]} &

\parbox[t]{0.17\linewidth}{\centering OpenOrganelle hela-2\\[0.1cm]\begin{minipage}{0.37\linewidth}\includegraphics[width=\linewidth]{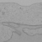}\end{minipage}\hspace{0.04\linewidth}\begin{minipage}{0.37\linewidth}\includegraphics[width=\linewidth]{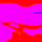}\end{minipage}\\[0.1cm]} & 

\parbox[t]{0.17\linewidth}{\centering OpenOrganelle hela-3\\[0.1cm]\begin{minipage}{0.37\linewidth}\includegraphics[width=\linewidth]{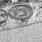}\end{minipage}\hspace{0.04\linewidth}\begin{minipage}{0.37\linewidth}\includegraphics[width=\linewidth]{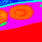}\end{minipage}\\[0.1cm]} & 

\parbox[t]{0.17\linewidth}{\centering OpenOrg. jurkat-1\\[0.1cm]\begin{minipage}{0.37\linewidth}\includegraphics[width=\linewidth]{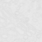}\end{minipage}\hspace{0.04\linewidth}\begin{minipage}{0.37\linewidth}\includegraphics[width=\linewidth]{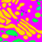}\end{minipage}\\[0.1cm]} & 

\parbox[t]{0.17\linewidth}{\centering OpenOrg. marco-2\\[0.1cm]\begin{minipage}{0.37\linewidth}\includegraphics[width=\linewidth]{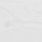}\end{minipage}\hspace{0.04\linewidth}\begin{minipage}{0.37\linewidth}\includegraphics[width=\linewidth]{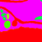}\end{minipage}\\[0.1cm]} \\

\midrule

\parbox[t]{0.17\linewidth}{\centering OpenOrg. sum159\\[0.1cm]\begin{minipage}{0.37\linewidth}\includegraphics[width=\linewidth]{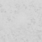}\end{minipage}\hspace{0.04\linewidth}\begin{minipage}{0.37\linewidth}\includegraphics[width=\linewidth]{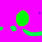}\end{minipage}\\[0.1cm]} &

\parbox[t]{0.17\linewidth}{\centering Pneumothorax M.\\[0.1cm]\begin{minipage}{0.37\linewidth}\includegraphics[width=\linewidth]{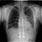}\end{minipage}\hspace{0.04\linewidth}\begin{minipage}{0.37\linewidth}\includegraphics[width=\linewidth]{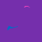}\end{minipage}\\[0.1cm]} & 

\parbox[t]{0.17\linewidth}{\centering ProstateADC\\[0.1cm]\begin{minipage}{0.37\linewidth}\includegraphics[width=\linewidth]{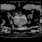}\end{minipage}\hspace{0.04\linewidth}\begin{minipage}{0.37\linewidth}\includegraphics[width=\linewidth]{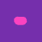}\end{minipage}\\[0.1cm]} & 

\parbox[t]{0.17\linewidth}{\centering ProstateT2\\[0.1cm]\begin{minipage}{0.37\linewidth}\includegraphics[width=\linewidth]{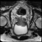}\end{minipage}\hspace{0.04\linewidth}\begin{minipage}{0.37\linewidth}\includegraphics[width=\linewidth]{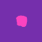}\end{minipage}\\[0.1cm]} & 

\parbox[t]{0.17\linewidth}{\centering QIN-Prostate-L.\\[0.1cm]\begin{minipage}{0.37\linewidth}\includegraphics[width=\linewidth]{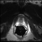}\end{minipage}\hspace{0.04\linewidth}\begin{minipage}{0.37\linewidth}\includegraphics[width=\linewidth]{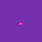}\end{minipage}\\[0.1cm]} \\

\midrule

\parbox[t]{0.17\linewidth}{\centering QIN-Prostate-P.\\[0.1cm]\begin{minipage}{0.37\linewidth}\includegraphics[width=\linewidth]{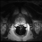}\end{minipage}\hspace{0.04\linewidth}\begin{minipage}{0.37\linewidth}\includegraphics[width=\linewidth]{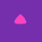}\end{minipage}\\[0.1cm]} &

\parbox[t]{0.17\linewidth}{\centering SpineMR\\[0.1cm]\begin{minipage}{0.37\linewidth}\includegraphics[width=\linewidth]{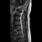}\end{minipage}\hspace{0.04\linewidth}\begin{minipage}{0.37\linewidth}\includegraphics[width=\linewidth]{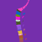}\end{minipage}\\[0.1cm]} & 

\parbox[t]{0.17\linewidth}{\centering TCIA-LCTSC\\[0.1cm]\begin{minipage}{0.37\linewidth}\includegraphics[width=\linewidth]{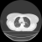}\end{minipage}\hspace{0.04\linewidth}\begin{minipage}{0.37\linewidth}\includegraphics[width=\linewidth]{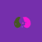}\end{minipage}\\[0.1cm]} & 

\parbox[t]{0.17\linewidth}{\centering WMH flair\\[0.1cm]\begin{minipage}{0.37\linewidth}\includegraphics[width=\linewidth]{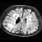}\end{minipage}\hspace{0.04\linewidth}\begin{minipage}{0.37\linewidth}\includegraphics[width=\linewidth]{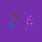}\end{minipage}\\[0.1cm]} & 

\parbox[t]{0.17\linewidth}{\centering WMH T1\\[0.1cm]\begin{minipage}{0.37\linewidth}\includegraphics[width=\linewidth]{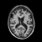}\end{minipage}\hspace{0.04\linewidth}\begin{minipage}{0.37\linewidth}\includegraphics[width=\linewidth]{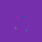}\end{minipage}\\[0.1cm]} \\

\midrule

\cellcolor{yellow!30}\parbox[t]{0.17\linewidth}{\centering BraTS flair\\[0.1cm]\begin{minipage}{0.37\linewidth}\includegraphics[width=\linewidth]{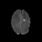}\end{minipage}\hspace{0.04\linewidth}\begin{minipage}{0.37\linewidth}\includegraphics[width=\linewidth]{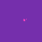}\end{minipage}\\[0.1cm]} & 

\cellcolor{yellow!30}\parbox[t]{0.17\linewidth}{\centering Breast-Ultrasound\\[0.1cm]\begin{minipage}{0.37\linewidth}\includegraphics[width=\linewidth]{figures/dataset_samples/breast-ultra_in.png}\end{minipage}\hspace{0.04\linewidth}\begin{minipage}{0.37\linewidth}\includegraphics[width=\linewidth]{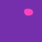}\end{minipage}\\[0.1cm]} & 

\cellcolor{yellow!30}\parbox[t]{0.17\linewidth}{\centering COVID-19-Rad.\\[0.1cm]\begin{minipage}{0.37\linewidth}\includegraphics[width=\linewidth]{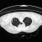}\end{minipage}\hspace{0.04\linewidth}\begin{minipage}{0.37\linewidth}\includegraphics[width=\linewidth]{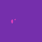}\end{minipage}\\[0.1cm]} &

\cellcolor{yellow!30}\parbox[t]{0.17\linewidth}{\centering ISIC2018\\[0.1cm]\begin{minipage}{0.37\linewidth}\includegraphics[width=\linewidth]{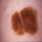}\end{minipage}\hspace{0.04\linewidth}\begin{minipage}{0.37\linewidth}\includegraphics[width=\linewidth]{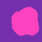}\end{minipage}\\[0.1cm]} &

\parbox[t]{0.17\linewidth}{\centering} \\
         
\bottomrule
\end{tabular}
\end{table}

{
\begin{table}
\tiny
\begin{tabular}{lll}
    \toprule
    Dataset & Link & Licence \\
    \midrule
    AMOSMR~\cite{ji2022amos} & \url{https://amos22.grand-challenge.org/Dataset/} & CC BY-NC-SA \\
    autoPET~\cite{gatidis2022whole} & \url{https://www.cancerimagingarchive.net/collection/fdg-pet-ct-lesions/}  & TCIA Restricted \\
    BraTS T1~\cite{bakas2017segmentation,bakas10segmentation,bakas2017advancing,bakas2018identifying,menze2014multimodal} & \url{http://braintumorsegmentation.org/} & Data Usage Agreement \\
    BraTS T1CE~\cite{bakas2017segmentation,bakas10segmentation,bakas2017advancing,bakas2018identifying,menze2014multimodal} & \url{http://braintumorsegmentation.org/} & Data Usage Agreement \\
    CDD-CESM~\cite{khaled2021categorized} & \url{https://www.cancerimagingarchive.net/collection/cdd-cesm/} & CC BY 4.0 \\
    Cervical Cancer~\cite{mayr2023cervical} & \url{https://www.cancerimagingarchive.net/collection/cc-tumor-heterogeneity/} & CC BY 4.0 \\
    Chest-Xray-M.-L.~\cite{jaeger2013automatic,candemir2013lung}  & \url{https://www.kaggle.com/datasets/nikhilpandey360/chest-xray-masks-and-labels} & CC0: Public Domain \\
    COVID-CT-Seg.~\cite{roth2022rapid,ct_images}  & \url{https://covid-segmentation.grand-challenge.org/Data/} & CC0: Public Domain \\
    COVID-QU-Ex-inf.~\cite{tahir2021covid,rahman2021exploring,degerli2021covid,chowdhury2020can,anas_m__tahir_muhammad_e__h__chowdhury_yazan_qiblawey_amith_khandakar_tawsifur_rahman_serkan_kiranyaz_uzair_khurshid_nabil_ibtehaz_sakib_mahmud_maymouna_ezeddin_2022}  & \url{https://www.kaggle.com/datasets/anasmohammedtahir/covidqu} & CC BY-SA 4.0\\
    COVID-QU-EX-L.~\cite{tahir2021covid,rahman2021exploring,degerli2021covid,chowdhury2020can,anas_m__tahir_muhammad_e__h__chowdhury_yazan_qiblawey_amith_khandakar_tawsifur_rahman_serkan_kiranyaz_uzair_khurshid_nabil_ibtehaz_sakib_mahmud_maymouna_ezeddin_2022}  & \url{https://www.kaggle.com/datasets/anasmohammedtahir/covidqu} & CC BY-SA 4.0 \\
    COVID19-CT-Segb.~\cite{ma2021toward}  & \url{https://github.com/JunMa11/COVID-19-CT-Seg-Benchmark} & \href{https://github.com/JunMa11/COVID-19-CT-Seg-Benchmark?tab=readme-ov-file#datasets}{multiple licenses} \\
    crossmoda~\cite{dorent2023crossmoda} & \url{https://crossmoda-challenge.ml/} & CC BY-NC-SA 4.0\\
    hc18~\cite{van2018automated}  & \url{https://hc18.grand-challenge.org/} & CC BY 4.0 \\
    Heart~\cite{simpson2019large}  & \url{http://medicaldecathlon.com/} & CC-BY-SA 4.0 \\
    Intraretinal-Cyst.~\cite{ahmed2022deep}  & \url{https://www.kaggle.com/datasets/zeeshanahmed13/intraretinal-cystoid-fluid} & CC BY-NC-SA 4.0 \\
    ISLES2022 ADC~\cite{hernandez2022isles}  & \url{https://www.isles-challenge.org/} & CC BY-SA 4.0 \\
    ISLES2022 DWI~\cite{hernandez2022isles}  & \url{https://www.isles-challenge.org/} & CC BY-SA 4.0 \\
    Kvasir-SEG~\cite{jha2019kvasir,pogorelov2017kvasir}  & \url{https://datasets.simula.no/kvasir/} & \href{https://dl.acm.org/do/10.1145/3193289/full/#sec-license}{link to license}\\
    LungLesions~\cite{antonelli2022medical}  & \url{http://medicaldecathlon.com/} & CC-BY-SA 4.0 \\
    m2caiSeg~\cite{maqbool2020m2caiseg}  & \url{https://www.kaggle.com/datasets/salmanmaq/m2caiseg} & CC BY-NC 4.0 \\
    NeurIPS22CellSeg~\cite{NeurIPS-CellSeg}  & \url{https://neurips22-cellseg.grand-challenge.org/} & CC BY-NC-ND \\
    OpenOrganelle hela-2~\cite{heinrich2021whole} & \url{https://open.quiltdata.com/b/janelia-cosem-publications/tree/heinrich-2021a/} &  CC BY 4.0 \\
    OpenOrganelle hela-3~\cite{heinrich2021whole}  & \url{https://open.quiltdata.com/b/janelia-cosem-publications/tree/heinrich-2021a/} &  CC BY 4.0 \\
    OpenOrganelle jurkat-1~\cite{heinrich2021whole}  & \url{https://open.quiltdata.com/b/janelia-cosem-publications/tree/heinrich-2021a/} &  CC BY 4.0 \\
    OpenOrganelle macro-2~\cite{heinrich2021whole}  & \url{https://open.quiltdata.com/b/janelia-cosem-publications/tree/heinrich-2021a/} &  CC BY 4.0 \\
    OpenOrganelle sum159~\cite{heinrich2021whole}  & \url{https://open.quiltdata.com/b/janelia-cosem-publications/tree/heinrich-2021a/} &  CC BY 4.0 \\
    Pneumothorax M.~\cite{siim-acr-pneumothorax-segmentation}  & \url{https://www.kaggle.com/datasets/vbookshelf/pneumothorax-chest-xray-images-and-masks}  & Unknown \\
    ProstateADC~\cite{simpson2019large}  & \url{http://medicaldecathlon.com} & CC-BY-SA 4.0 \\
    ProstateT2~\cite{simpson2019large}  & \url{http://medicaldecathlon.com} & CC-BY-SA 4.0 \\
    QIN-Prostate-L.~\cite{antonelli2022medical,fedorov2018annotated} & \url{http://doi.org/10.7937/K9/TCIA.2018.MR1CKGND} & CC BY 4.0 \\
    QIN-Prostate-P.~\cite{antonelli2022medical,fedorov2018annotated}  & \url{http://doi.org/10.7937/K9/TCIA.2018.MR1CKGND} & CC BY 4.0 \\
    SpineMR~\cite{zukic2014robust}  & \url{https://www.cg.informatik.uni-siegen.de/en/spine-segmentation-and-analysis} & Unknown \\
    TCIA-LCTSC~\cite{lctsc}  & \url{https://www.cancerimagingarchive.net/collection/lctsc/} & CC BY 3.0 \\
    WMH flair~\cite{kuijf2019standardized} & \url{https://wmh.isi.uu.nl/} & CC BY 4.0 \\
    WMH T1~\cite{kuijf2019standardized} & \url{https://wmh.isi.uu.nl/} & CC BY 4.0 \\
    BraTS flair~\cite{bakas2017segmentation,bakas10segmentation,bakas2017advancing,bakas2018identifying,menze2014multimodal} & \url{http://braintumorsegmentation.org/} & Data Usage Agreement \\
    Breast-Ultrasound~\cite{al2020dataset} & \url{https://www.kaggle.com/datasets/aryashah2k/breast-ultrasound-images-dataset} & CC0: Public Domain \\
    COVID-19-Rad.~\cite{rahman2021exploring,chowdhury2020can}  & \url{https://www.kaggle.com/datasets/tawsifurrahman/covid19-radiography-database} & \href{https://www.kaggle.com/datasets/tawsifurrahman/covid19-radiography-database/data}{multiple licenses}\\
    ISIC2018~\cite{tschandl2018ham10000,codella2018skin,codella2019skin} & \url{https://challenge.isic-archive.com/data/} & CC0: Public Domain \\
    \bottomrule
\end{tabular}
\caption{Datasets with their citation, link to the respective website and license.}
\label{tab:dataset_license}
\end{table}
}


\clearpage

\section{Examples for the Utilized 30 Tasks}
\label{sec:appendix_exampletasks}
In~\Cref{tab:sem_tasks},~\Cref{tab:trans_tasks} and~\Cref{tab:gen_tasks} , we showcase an example for each visual task that we use to learn task representations, grouped by \emph{semantic-}, \emph{generative-} and \emph{transformative} tasks in the three tables.
We can grasp the diversity of tasks, as they are distributed over the three categories.

{
\begin{table}[ht]
\small
\centering
\renewcommand{\arraystretch}{1.3}
\setlength{\extrarowheight}{0pt}
\caption{Semantic tasks from our datasets. Here, we highlight unseen tasks via yellow tinted tiles.}
\label{tab:sem_tasks}
\begin{tabular}{c|c|c|c|c}
\toprule
\multicolumn{5}{c}{\textbf{Semantic Tasks}} \\
\midrule
\parbox[t]{0.17\linewidth}{\centering Bounding Boxes\\[0.1cm]\begin{minipage}{0.48\linewidth}\includegraphics[width=\linewidth]{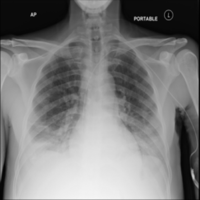}\end{minipage}\hspace{0.04\linewidth}\begin{minipage}{0.48\linewidth}\includegraphics[width=\linewidth]{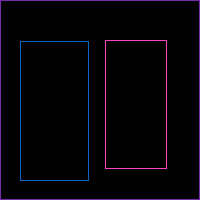}\end{minipage}\\[0.1cm]} & 
\parbox[t]{0.17\linewidth}{\centering Segmentation\\[0.1cm]\begin{minipage}{0.48\linewidth}\includegraphics[width=\linewidth]{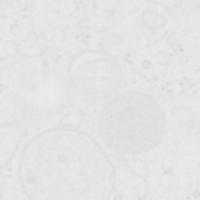}\end{minipage}\hspace{0.04\linewidth}\begin{minipage}{0.48\linewidth}\includegraphics[width=\linewidth]{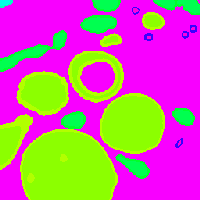}\end{minipage}\\[0.1cm]} & 
\parbox[t]{0.17\linewidth}{\centering Interactive Segm.\\[0.1cm]\begin{minipage}{0.48\linewidth}\includegraphics[width=\linewidth]{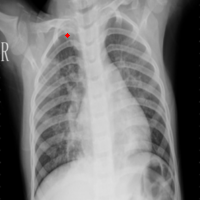}\end{minipage}\hspace{0.04\linewidth}\begin{minipage}{0.48\linewidth}\includegraphics[width=\linewidth]{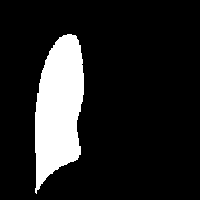}\end{minipage}\\[0.1cm]} & 
\parbox[t]{0.17\linewidth}{\centering Points\\[0.1cm]\begin{minipage}{0.48\linewidth}\includegraphics[width=\linewidth]{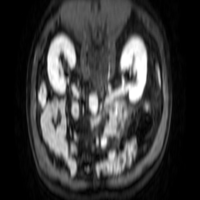}\end{minipage}\hspace{0.04\linewidth}\begin{minipage}{0.48\linewidth}\includegraphics[width=\linewidth]{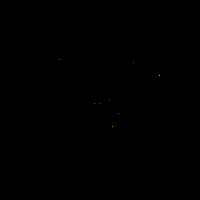}\end{minipage}\\[0.1cm]} & 
\parbox[t]{0.17\linewidth}{\centering Semantic Edges\\[0.1cm]\begin{minipage}{0.48\linewidth}\includegraphics[width=\linewidth]{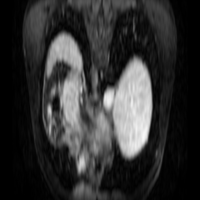}\end{minipage}\hspace{0.04\linewidth}\begin{minipage}{0.48\linewidth}\includegraphics[width=\linewidth]{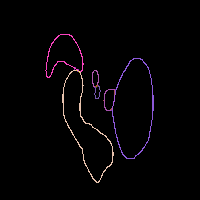}\end{minipage}\\[0.1cm]} \\
\midrule
\parbox[t]{0.17\linewidth}{\centering Skeletons\\[0.1cm]\begin{minipage}{0.48\linewidth}\includegraphics[width=\linewidth]{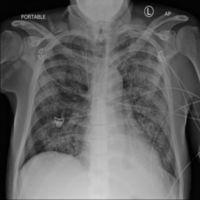}\end{minipage}\hspace{0.04\linewidth}\begin{minipage}{0.48\linewidth}\includegraphics[width=\linewidth]{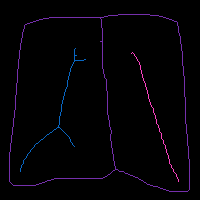}\end{minipage}\\[0.1cm]} & 
\cellcolor{yellow!30}\parbox[t]{0.17\linewidth}{\centering Semantic Hulls\\[0.1cm]\begin{minipage}{0.48\linewidth}\includegraphics[width=\linewidth]{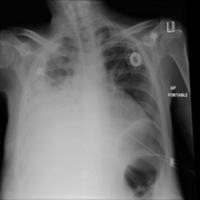}\end{minipage}\hspace{0.04\linewidth}\begin{minipage}{0.48\linewidth}\includegraphics[width=\linewidth]{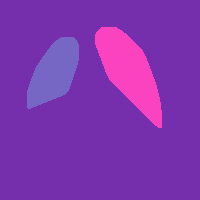}\end{minipage}\\[0.1cm]} & 
\parbox[t]{0.17\linewidth}{\centering} & 
\parbox[t]{0.17\linewidth}{\centering} & 
\parbox[t]{0.17\linewidth}{\centering} \\
\bottomrule
\end{tabular}
\end{table}
}

{
\begin{table}[ht]
\small
\centering
\renewcommand{\arraystretch}{1.3}
\setlength{\extrarowheight}{0pt}
\caption{Image transformation tasks from our datasets. Yellow tinted tiles: unseen tasks.}
\label{tab:trans_tasks}
\begin{tabular}{c|c|c|c|c}
\toprule
\multicolumn{5}{c}{\textbf{Image Transformation Tasks}} \\
\midrule
\parbox[t]{0.17\linewidth}{\centering Decrease Brightness\\[0.1cm]\begin{minipage}{0.48\linewidth}\includegraphics[width=\linewidth]{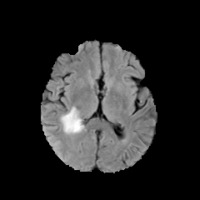
}\end{minipage}\hspace{0.04\linewidth}\begin{minipage}{0.48\linewidth}\includegraphics[width=\linewidth]{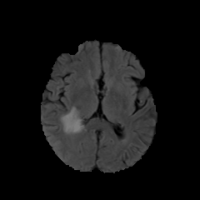
}\end{minipage}\\[0.1cm]} & 
\parbox[t]{0.17\linewidth}{\centering Distance Map\\[0.1cm]\begin{minipage}{0.48\linewidth}\includegraphics[width=\linewidth]{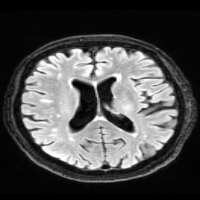
}\end{minipage}\hspace{0.04\linewidth}\begin{minipage}{0.48\linewidth}\includegraphics[width=\linewidth]{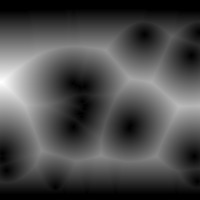
}\end{minipage}\\[0.1cm]} & 
\parbox[t]{0.17\linewidth}{\centering False Colorization\\[0.1cm]\begin{minipage}{0.48\linewidth}\includegraphics[width=\linewidth]{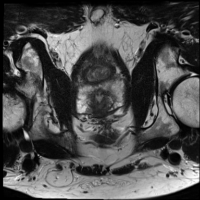
}\end{minipage}\hspace{0.04\linewidth}\begin{minipage}{0.48\linewidth}\includegraphics[width=\linewidth]{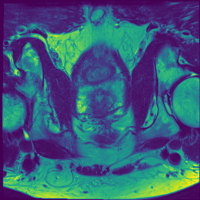
}\end{minipage}\\[0.1cm]} & 
\parbox[t]{0.17\linewidth}{\centering Geodesic And Euclid\\[0.1cm]\begin{minipage}{0.48\linewidth}\includegraphics[width=\linewidth]{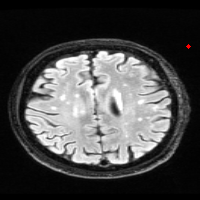
}\end{minipage}\hspace{0.04\linewidth}\begin{minipage}{0.48\linewidth}\includegraphics[width=\linewidth]{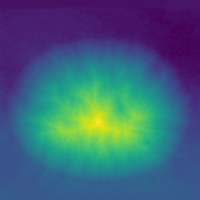
}\end{minipage}\\[0.1cm]} & 
\parbox[t]{0.17\linewidth}{\centering Geodesic\\[0.1cm]\begin{minipage}{0.48\linewidth}\includegraphics[width=\linewidth]{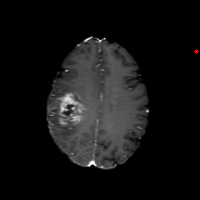
}\end{minipage}\hspace{0.04\linewidth}\begin{minipage}{0.48\linewidth}\includegraphics[width=\linewidth]{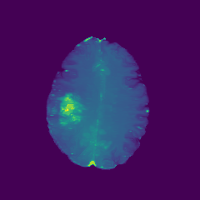
}\end{minipage}\\[0.1cm]} \\
\midrule
\parbox[t]{0.17\linewidth}{\centering Horizontal Flip\\[0.1cm]\begin{minipage}{0.48\linewidth}\includegraphics[width=\linewidth]{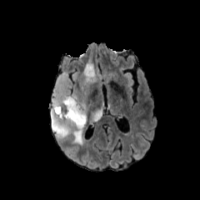
}\end{minipage}\hspace{0.04\linewidth}\begin{minipage}{0.48\linewidth}\includegraphics[width=\linewidth]{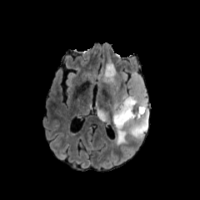
}\end{minipage}\\[0.1cm]} & 
\parbox[t]{0.17\linewidth}{\centering Identity\\[0.1cm]\begin{minipage}{0.48\linewidth}\includegraphics[width=\linewidth]{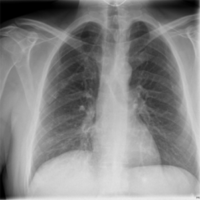
}\end{minipage}\hspace{0.04\linewidth}\begin{minipage}{0.48\linewidth}\includegraphics[width=\linewidth]{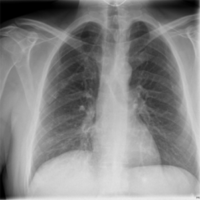
}\end{minipage}\\[0.1cm]} & 
\parbox[t]{0.17\linewidth}{\centering Increase Contrast\\[0.1cm]\begin{minipage}{0.48\linewidth}\includegraphics[width=\linewidth]{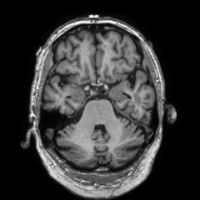
}\end{minipage}\hspace{0.04\linewidth}\begin{minipage}{0.48\linewidth}\includegraphics[width=\linewidth]{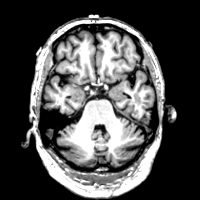
}\end{minipage}\\[0.1cm]} & 
\parbox[t]{0.17\linewidth}{\centering Rotate90\\[0.1cm]\begin{minipage}{0.48\linewidth}\includegraphics[width=\linewidth]{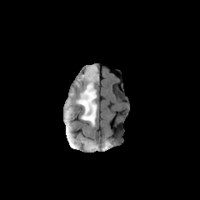
}\end{minipage}\hspace{0.04\linewidth}\begin{minipage}{0.48\linewidth}\includegraphics[width=\linewidth]{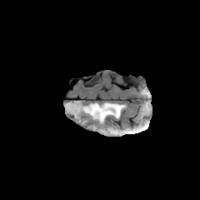
}\end{minipage}\\[0.1cm]} & 
\parbox[t]{0.17\linewidth}{\centering Rotate180\\[0.1cm]\begin{minipage}{0.48\linewidth}\includegraphics[width=\linewidth]{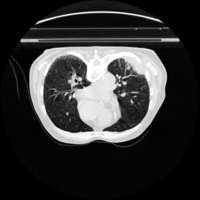
}\end{minipage}\hspace{0.04\linewidth}\begin{minipage}{0.48\linewidth}\includegraphics[width=\linewidth]{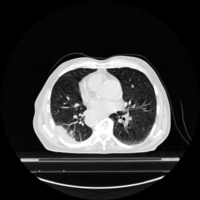
}\end{minipage}\\[0.1cm]} \\
\midrule
\parbox[t]{0.17\linewidth}{\centering Rotate270\\[0.1cm]\begin{minipage}{0.48\linewidth}\includegraphics[width=\linewidth]{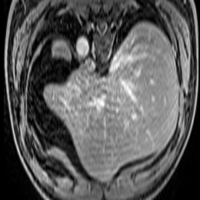
}\end{minipage}\hspace{0.04\linewidth}\begin{minipage}{0.48\linewidth}\includegraphics[width=\linewidth]{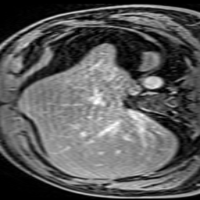
}\end{minipage}\\[0.1cm]} & 
\parbox[t]{0.17\linewidth}{\centering Zoom In\\[0.1cm]\begin{minipage}{0.48\linewidth}\includegraphics[width=\linewidth]{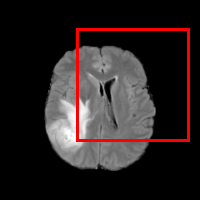
}\end{minipage}\hspace{0.04\linewidth}\begin{minipage}{0.48\linewidth}\includegraphics[width=\linewidth]{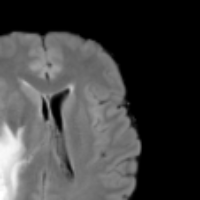
}\end{minipage}\\[0.1cm]} & 
\cellcolor{yellow!30}\parbox[t]{0.17\linewidth}{\centering Invert\\[0.1cm]\begin{minipage}{0.48\linewidth}\includegraphics[width=\linewidth]{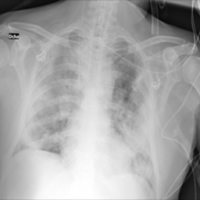
}\end{minipage}\hspace{0.04\linewidth}\begin{minipage}{0.48\linewidth}\includegraphics[width=\linewidth]{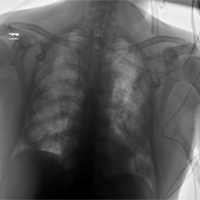
}\end{minipage}\\[0.1cm]} & 
\cellcolor{yellow!30}\parbox[t]{0.17\linewidth}{\centering Rotate45\\[0.1cm]\begin{minipage}{0.48\linewidth}\includegraphics[width=\linewidth]{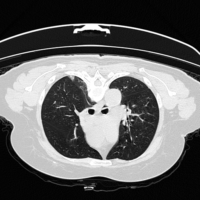
}\end{minipage}\hspace{0.04\linewidth}\begin{minipage}{0.48\linewidth}\includegraphics[width=\linewidth]{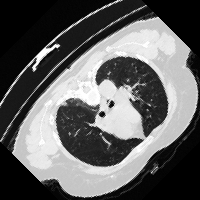
}\end{minipage}\\[0.1cm]} & 
\cellcolor{yellow!30}\parbox[t]{0.17\linewidth}{\centering Vertical Flip\\[0.03cm]\begin{minipage}{0.48\linewidth}\includegraphics[width=\linewidth]{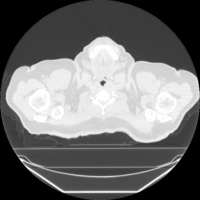
}\end{minipage}\hspace{0.04\linewidth}\begin{minipage}{0.48\linewidth}\includegraphics[width=\linewidth]{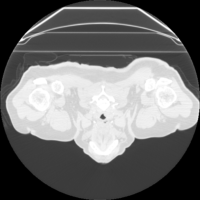
}\end{minipage}\\[0.1cm]} \\
\bottomrule
\end{tabular}
\end{table}
}

{
\begin{table}[ht]
\small
\centering
\renewcommand{\arraystretch}{1.3}
\setlength{\extrarowheight}{0pt}
\caption{Image generative tasks from our datasets. Yellow tinted tiles: unseen tasks.}
\label{tab:gen_tasks}
\begin{tabular}{c|c|c|c|c}
\toprule
\multicolumn{5}{c}{\textbf{Image Generative Tasks}} \\
\midrule
\parbox[t]{0.17\linewidth}{\centering Inpainting\\[0.1cm]\begin{minipage}{0.48\linewidth}\includegraphics[width=\linewidth]{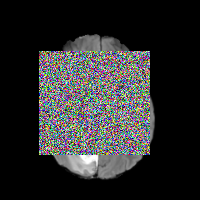
}\end{minipage}\hspace{0.04\linewidth}\begin{minipage}{0.48\linewidth}\includegraphics[width=\linewidth]{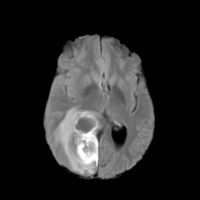
}\end{minipage}\\[0.1cm]} & 
\parbox[t]{0.17\linewidth}{\centering Inpainting(3x)\\[0.1cm]\begin{minipage}{0.48\linewidth}\includegraphics[width=\linewidth]{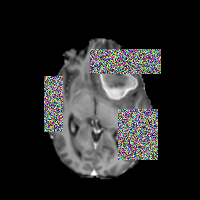
}\end{minipage}\hspace{0.04\linewidth}\begin{minipage}{0.48\linewidth}\includegraphics[width=\linewidth]{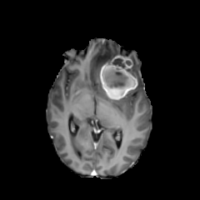
}\end{minipage}\\[0.1cm]} & 
\parbox[t]{0.17\linewidth}{\centering Image Generation\\[0.1cm]\begin{minipage}{0.48\linewidth}\includegraphics[width=\linewidth]{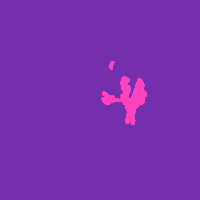
}\end{minipage}\hspace{0.04\linewidth}\begin{minipage}{0.48\linewidth}\includegraphics[width=\linewidth]{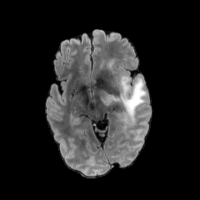
}\end{minipage}\\[0.1cm]} & 
\parbox[t]{0.17\linewidth}{\centering Salt Pepper Noise\\[0.1cm]\begin{minipage}{0.48\linewidth}\includegraphics[width=\linewidth]{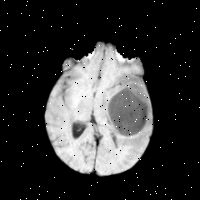
}\end{minipage}\hspace{0.04\linewidth}\begin{minipage}{0.48\linewidth}\includegraphics[width=\linewidth]{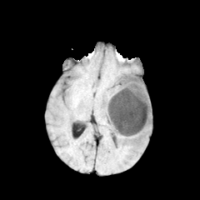
}\end{minipage}\\[0.1cm]} & 
\parbox[t]{0.17\linewidth}{\centering Super Resolution\\[0.1cm]\begin{minipage}{0.48\linewidth}\includegraphics[width=\linewidth]{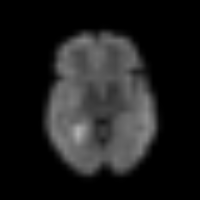
}\end{minipage}\hspace{0.04\linewidth}\begin{minipage}{0.48\linewidth}\includegraphics[width=\linewidth]{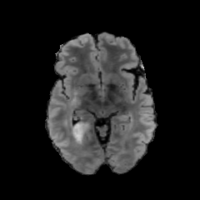
}\end{minipage}\\[0.1cm]} \\
\midrule
\cellcolor{yellow!30}\parbox[t]{0.17\linewidth}{\centering Colorization\\[0.1cm]\begin{minipage}{0.48\linewidth}\includegraphics[width=\linewidth]{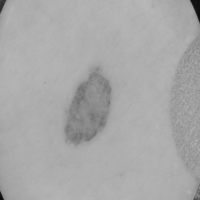
}\end{minipage}\hspace{0.04\linewidth}\begin{minipage}{0.48\linewidth}\includegraphics[width=\linewidth]{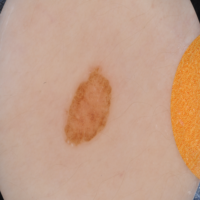
}\end{minipage}\\[0.1cm]} & 
\cellcolor{yellow!30}\parbox[t]{0.17\linewidth}{\centering Denoising\\[0.03cm]\begin{minipage}{0.48\linewidth}\includegraphics[width=\linewidth]{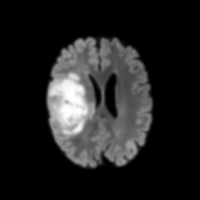
}\end{minipage}\hspace{0.04\linewidth}\begin{minipage}{0.48\linewidth}\includegraphics[width=\linewidth]{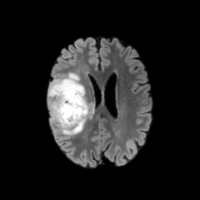
}\end{minipage}\\[0.1cm]} & 
\cellcolor{yellow!30}\parbox[t]{0.17\linewidth}{\centering Outpainting\\[0.03cm]\begin{minipage}{0.48\linewidth}\includegraphics[width=\linewidth]{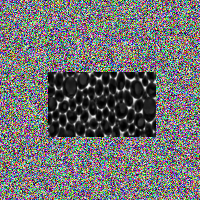
}\end{minipage}\hspace{0.04\linewidth}\begin{minipage}{0.48\linewidth}\includegraphics[width=\linewidth]{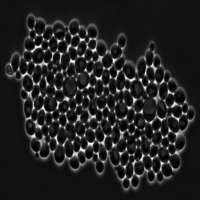
}\end{minipage}\\[0.1cm]} & 
\parbox[t]{0.17\linewidth}{\centering} &
\parbox[t]{0.17\linewidth}{\centering}\\
\bottomrule
\end{tabular}
\end{table}
}

\clearpage

\section{Hyperparamter Overview}
\label{sec:appendix_hyperparameters}

In~\Cref{tab:hyperparametertable}, we summarize the hyperparameters that we used to train the ResNet-50 architecture with different training strategies, namely SimCLR and our Task-Contrastive Learning strategy on the data, as described in the main paper.
Note that the batch size in~\Cref{tab:hyperparametertable} refers for SimCLR to $20$ inputs augmented twice, for \emph{TaCo} we sample $20$ visual tasks, for each we further sample two instances.
For \emph{TaCo + failure tasks} we sample $15$ visual tasks, for each we further sample two instances and in addition $10$ failure task instances.

\begin{table}[h]
    \centering    
    \caption{Hyperparameters for the training of SimCLR~\cite{chen2020simple} and our TaCo models.}
    \resizebox{\textwidth}{!}{%
    \begin{tabular}{lccc}
    \toprule
         & SimCLR & TaCo & TaCo + \emph{failure tasks} \\
         \midrule
         epochs &3 & 3&3 \\
         batch size& $2 \cdot 20 = \textbf{40}$& $2 \cdot 20 = \textbf{40}$& $15 \cdot 2 + 10 = \textbf{40}$ \\
         learning rate& $0.001$&$0.001$ &$0.001$ \\
         \emph{failure tasks} start iteration & -- & -- & $50.000$\\
         $\#$ \emph{failure tasks}& -- & -- & 10\\
         weight decay& $0.0001$& $0.0001$& $0.0001$\\
         momentum& $0.9$& $0.9$& $0.9$\\
         temperature $\tau$& $0.07$& $0.07$& $0.07$\\
         architecture& ResNet-50&ResNet-50 &ResNet-50 \\
         input channels&3 & 6& 6\\
         GPUs& 2$\times$NVIDIA 2080 Ti& 2$\times$NVIDIA 2080 Ti& 2$\times$NVIDIA 2080 Ti\\
         \bottomrule
    \end{tabular}
    }
    \label{tab:hyperparametertable}
\end{table}

\section{Additional Qualitative Results: t-SNE Plots}
\label{sec:appendix_tsne}
In this section, we display additional t-SNE plots to underline our insights in the main paper.

In~\Cref{fig:tsne_seen_seen_classes}, we only display embeddings of seen tasks and seen datasets, \ie, visual tasks present during training.
TaCo's learned representation space leads to nice clusters for most of the tasks, while CLIP often seems to form small \emph{instance clusters}, where an image with different task information is embedded similarly.
This showcases the focus of baselines to group by image instances and their dataset affiliation, rather than by the visual tasks, as indicative in the second row for CLIP.

In~\Cref{fig:tsne_unseen_unseen_classes}, only the visual task instances from the unseen tasks and unseen datasets are displayed.
Due to the small number of datasets and the strong difference in the dataset modalities, all methods form distinct clusters which show the different datasets (second row), yet, the task-based colorization in the first row leads to cleaner clusters for our \emph{TaCo} variants, showing that they are well suited to embed unseen/new task information.

\begin{figure*}[t]
    \centering
    \begin{tabular}{ccc}
          $\phantom{----}$ CLIP & $\phantom{------.--}$ TaCo & $\phantom{-------}$ TaCo + \emph{failure tasks}$\phantom{----}$ 
    \end{tabular}
    
    \rotatebox[origin=lb]{90}{$\phantom{---}$colored by tasks}\includegraphics[trim={0 0 8cm 0},clip,width=0.25\textwidth,height=0.25\textwidth]{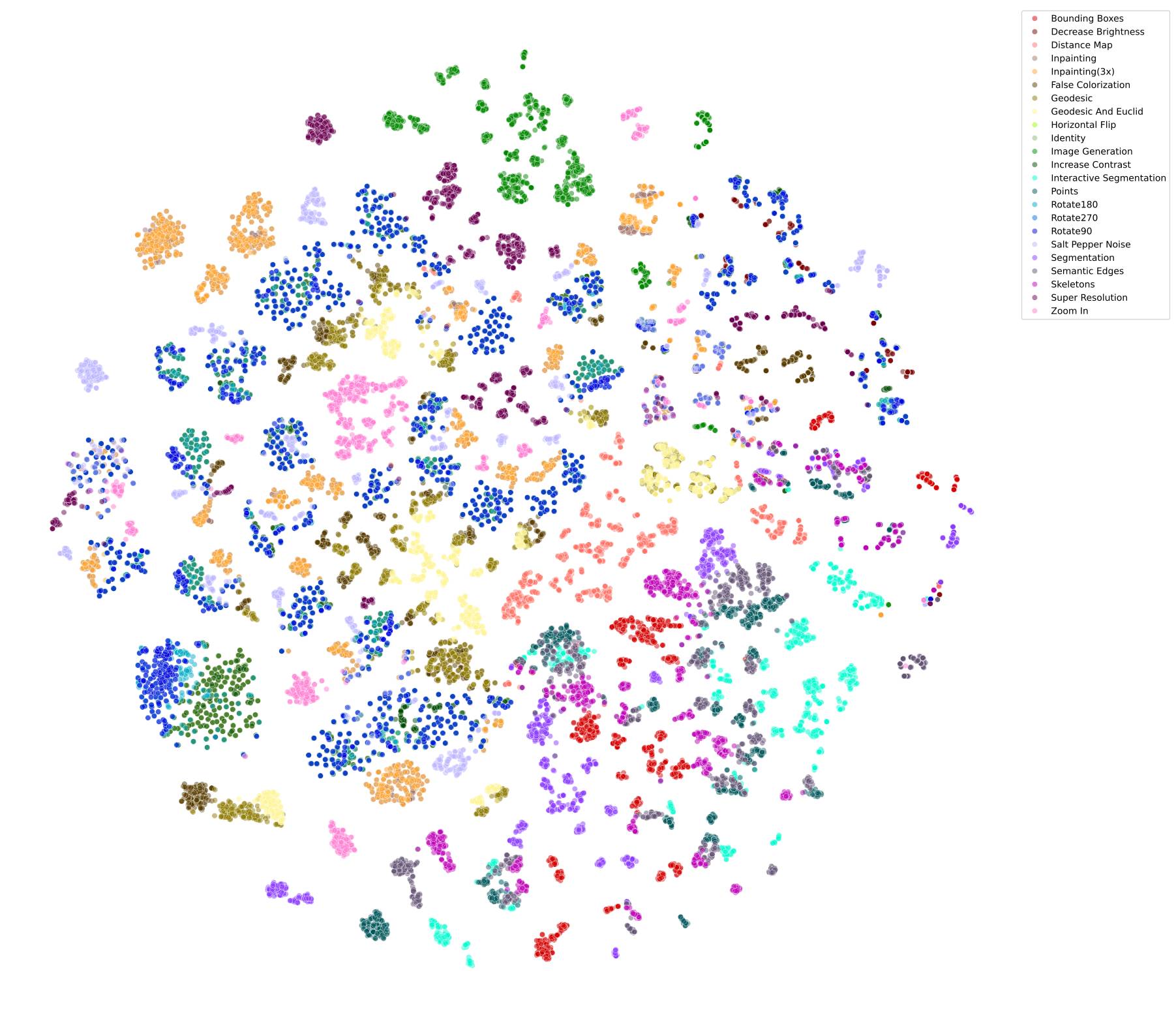}
    \includegraphics[trim={0 0 8cm 0},clip,width=0.25\textwidth,height=0.25\textwidth]{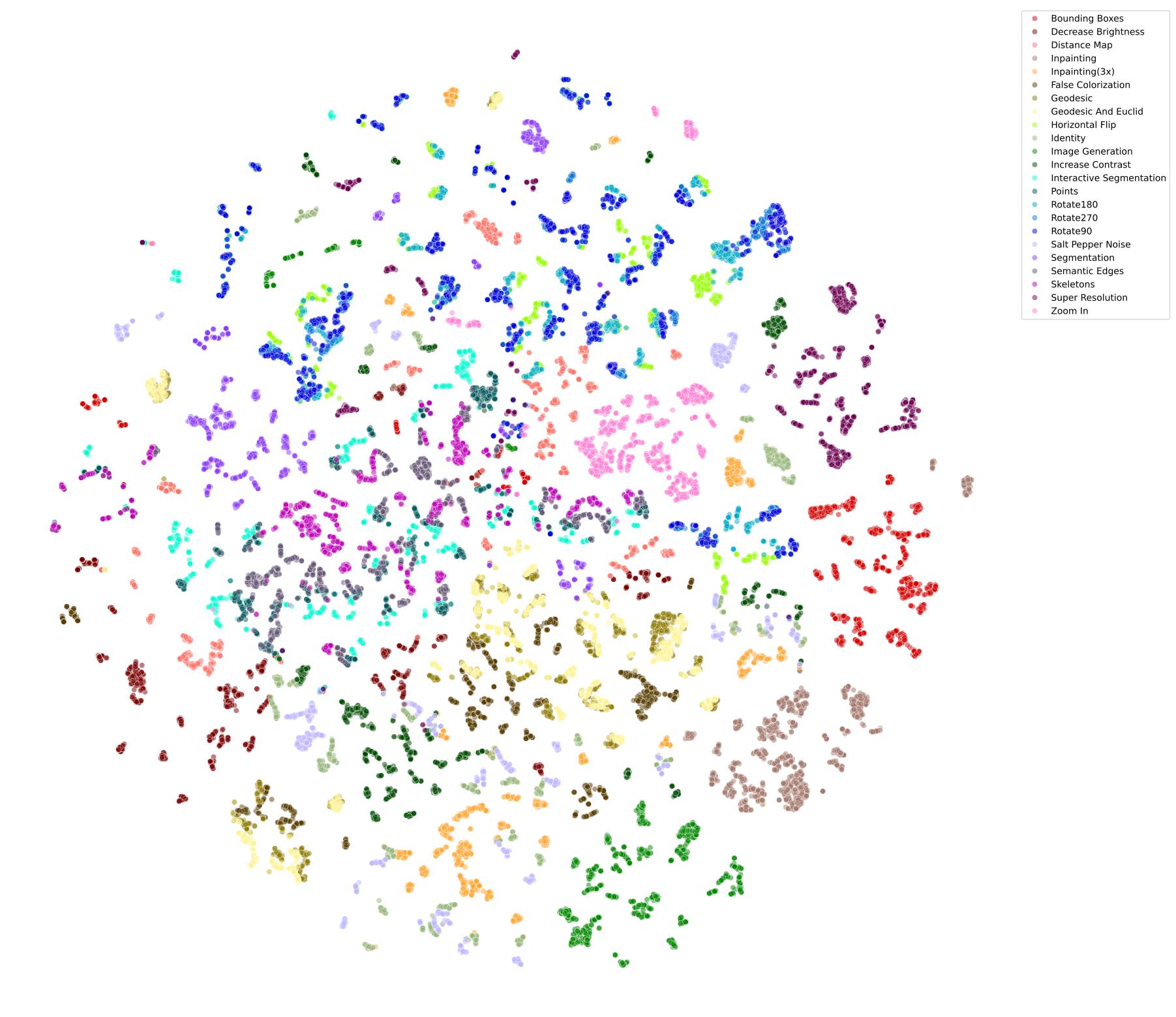}
    \includegraphics[trim={0 0 8cm 0},clip,width=0.25\textwidth,height=0.25\textwidth]{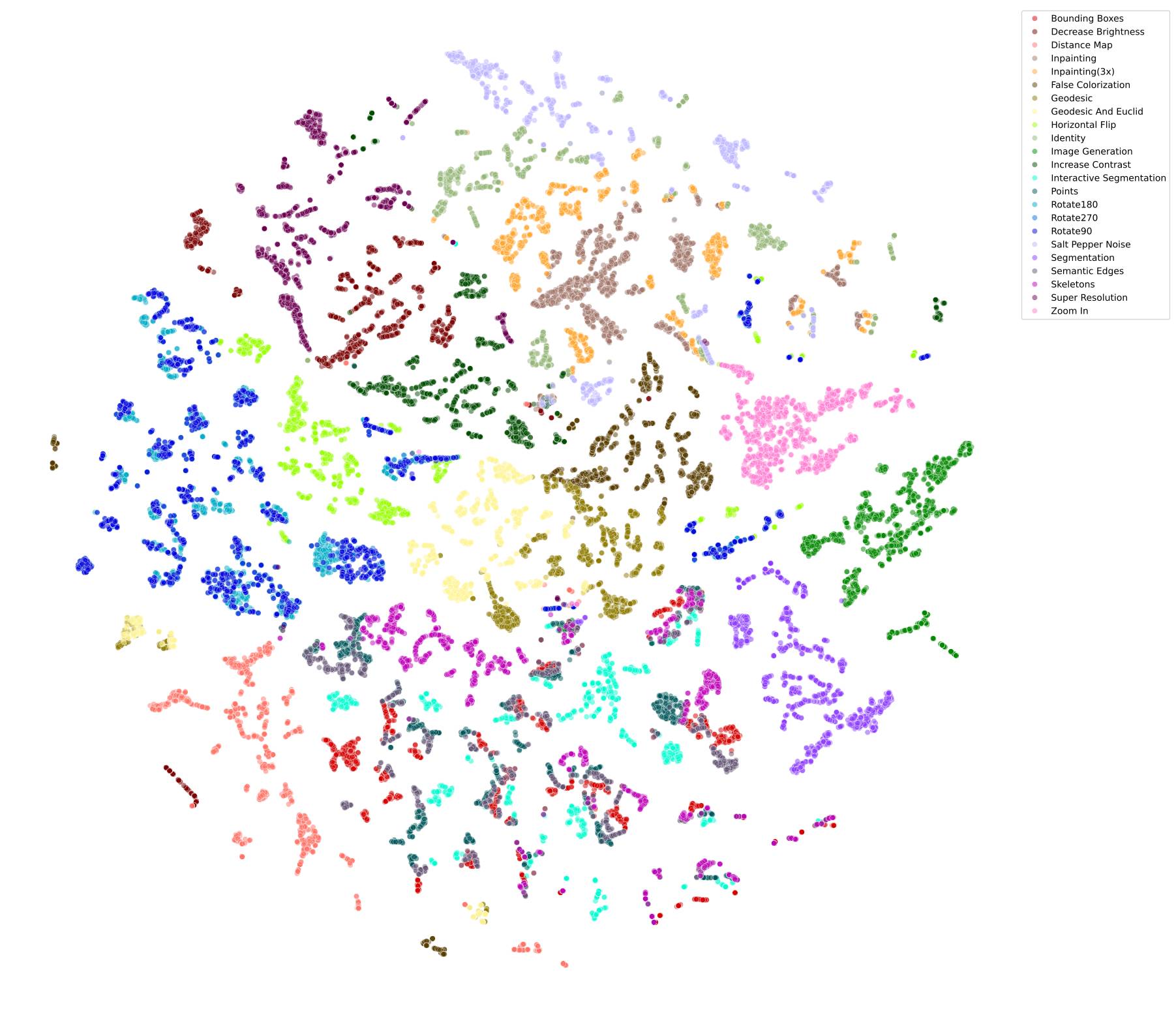}
    \includegraphics[trim={38.5cm 23cm 0 0},clip,height=0.25\textwidth]{figures/pdf_tsne_plots/SupCL/final_SupCL_sampled/seen_tasks_seen_datasets/tsne_classes_no_symbols_SupCL_final_SupCL_sampled_seen_tasks_seen_datasets_all.jpg}
        
    \rotatebox[origin=lb]{90}{$\phantom{---}$colored by datasets}\includegraphics[trim={0 0 9cm 0},clip,width=0.25\textwidth,height=0.25\textwidth]{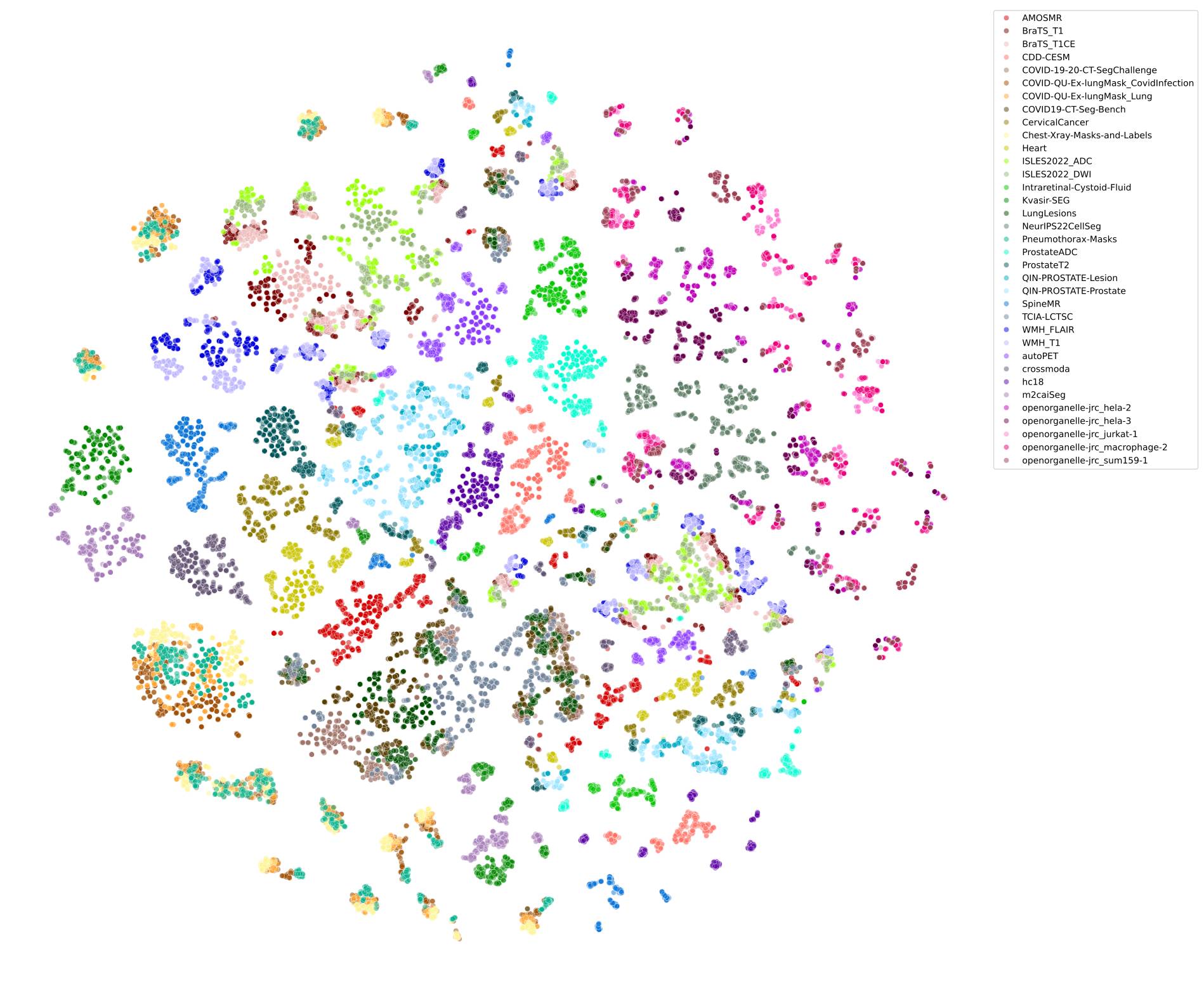}
    \includegraphics[trim={0 0 9cm 0},clip,width=0.25\textwidth,height=0.25\textwidth]{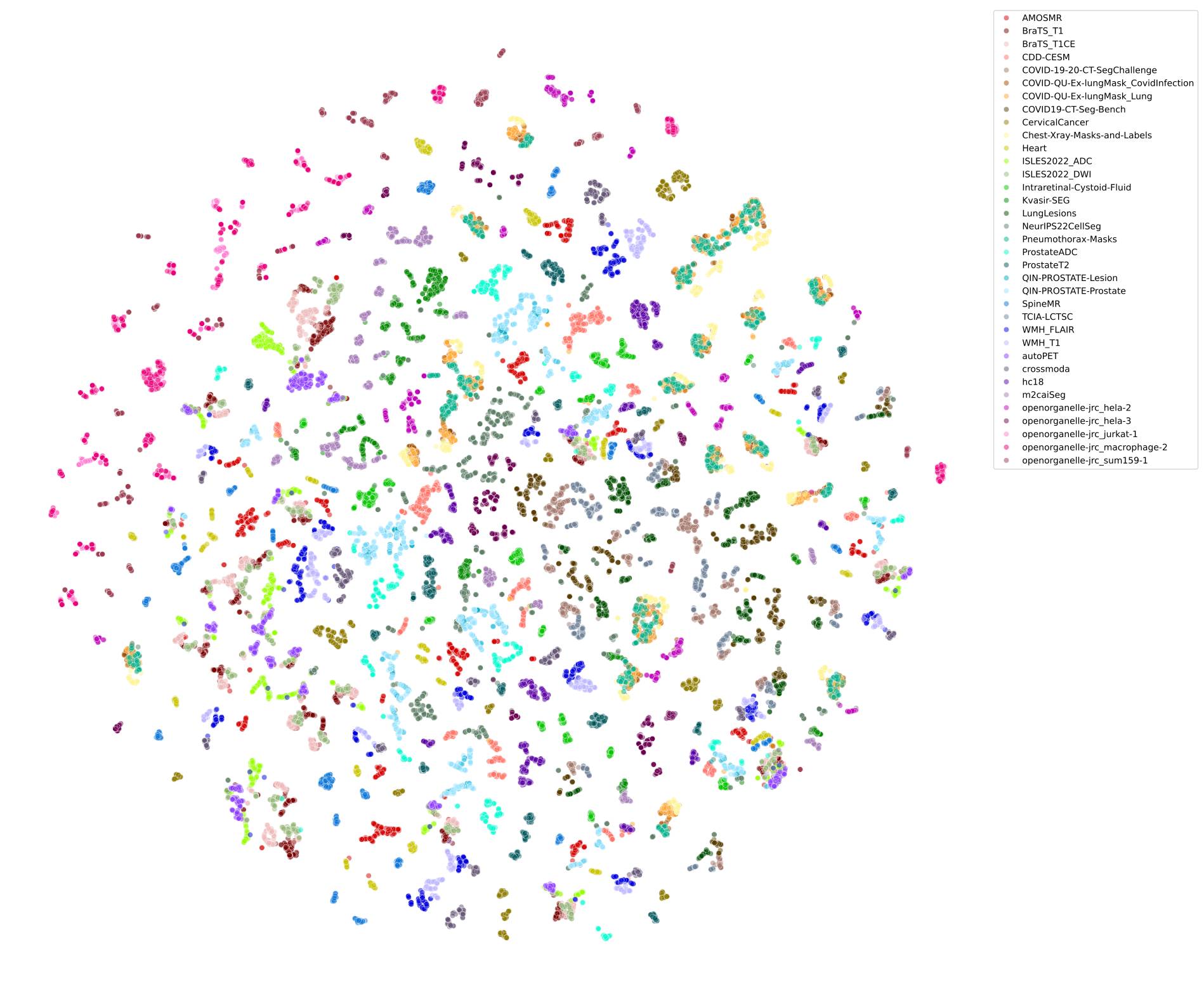}
    \includegraphics[trim={0 0 9cm 0},clip,width=0.25\textwidth,height=0.25\textwidth]{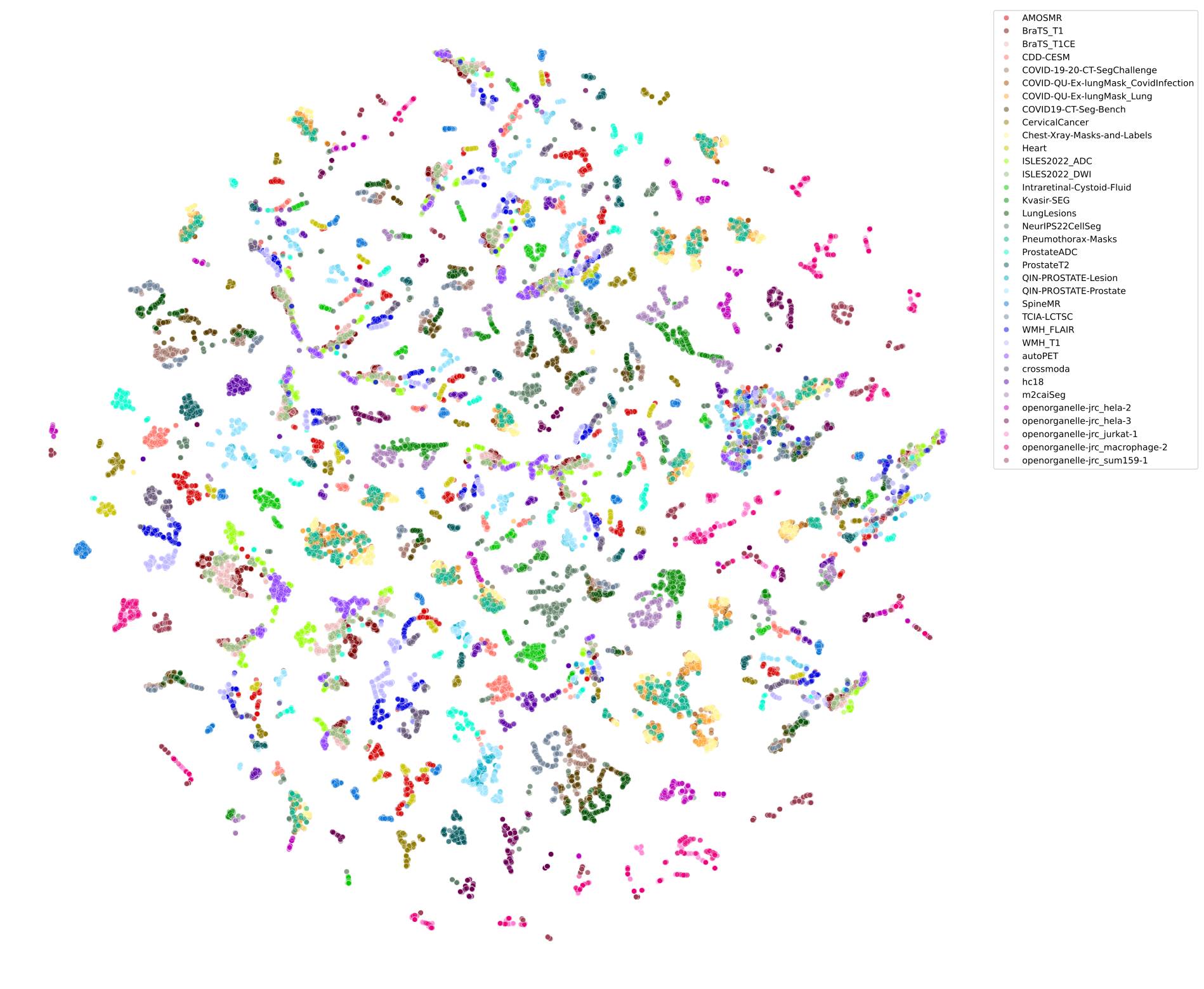}
    \includegraphics[trim={38cm 15cm 0 0},clip,height=0.25\textwidth]{figures/pdf_tsne_plots/SupCL/final_SupCL_sampled/seen_tasks_seen_datasets/tsne_datasets_no_symbols_SupCL_final_SupCL_sampled_seen_tasks_seen_datasets_all.jpg}
    \caption{Learned high dimensional representations from the \emph{seen tasks} and \emph{seen datasets} scenario projected via t-SNE. We showcase the embedding space of CLIP, \emph{TaCo} and \emph{TaCo + failure tasks}. Coloring is based on tasks in the first row and datasets in the second row.}
    \label{fig:tsne_seen_seen_classes}
\end{figure*}

\begin{figure*}[ht]
    \centering
    \begin{tabular}{ccc}
          $\phantom{----}$ CLIP & $\phantom{------.--}$ TaCo & $\phantom{-------}$ TaCo + \emph{failure tasks}$\phantom{----}$ 
    \end{tabular}
    
    \rotatebox[origin=lb]{90}{$\phantom{---}$colored by  tasks}\includegraphics[trim={0 0 6cm 0},clip,width=0.25\textwidth,height=0.25\textwidth]{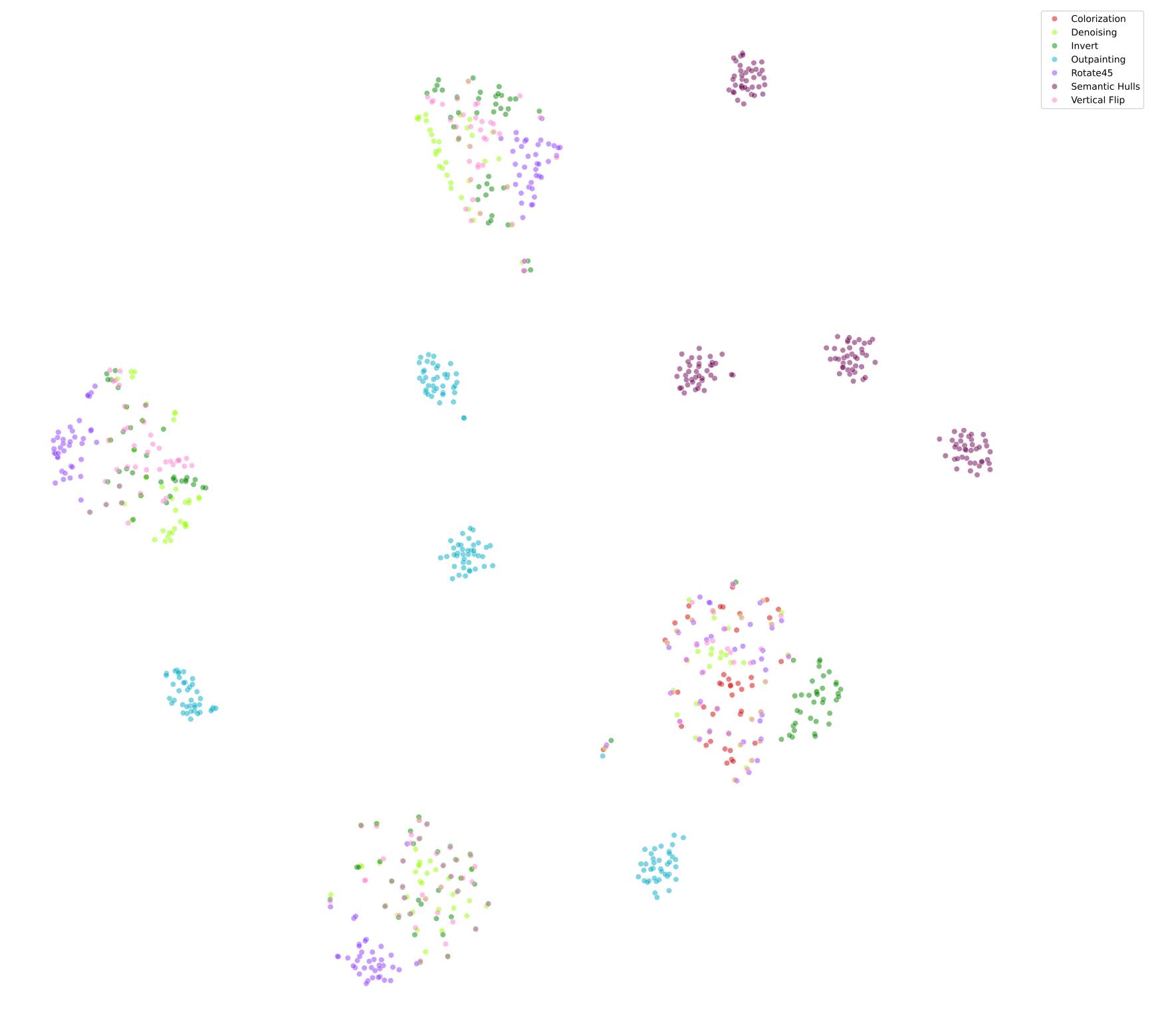}
    \includegraphics[trim={0 0 6cm 0},clip,width=0.25\textwidth,height=0.25\textwidth]{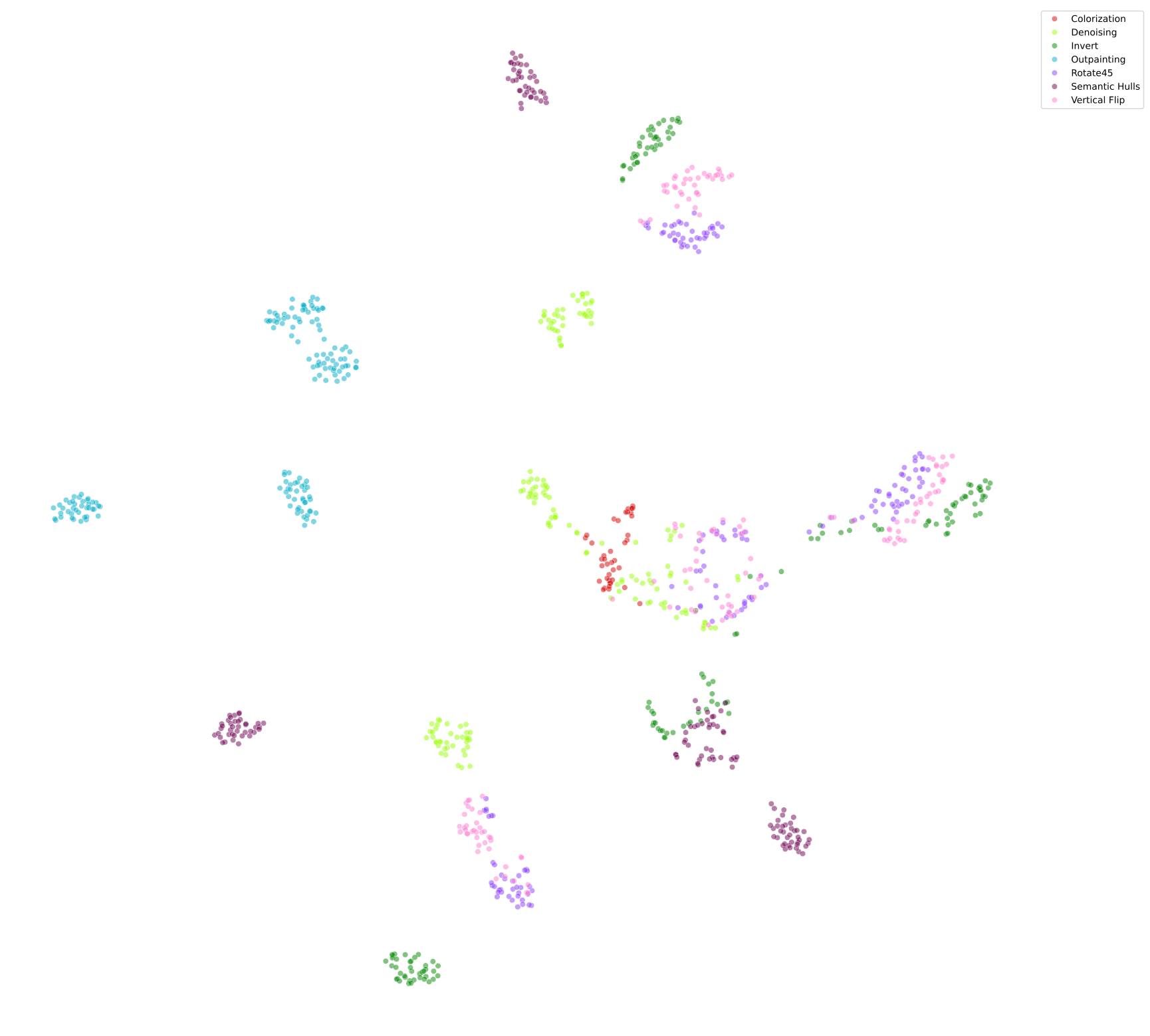}
    \includegraphics[trim={0 0 6cm 0},clip,width=0.25\textwidth,height=0.25\textwidth]{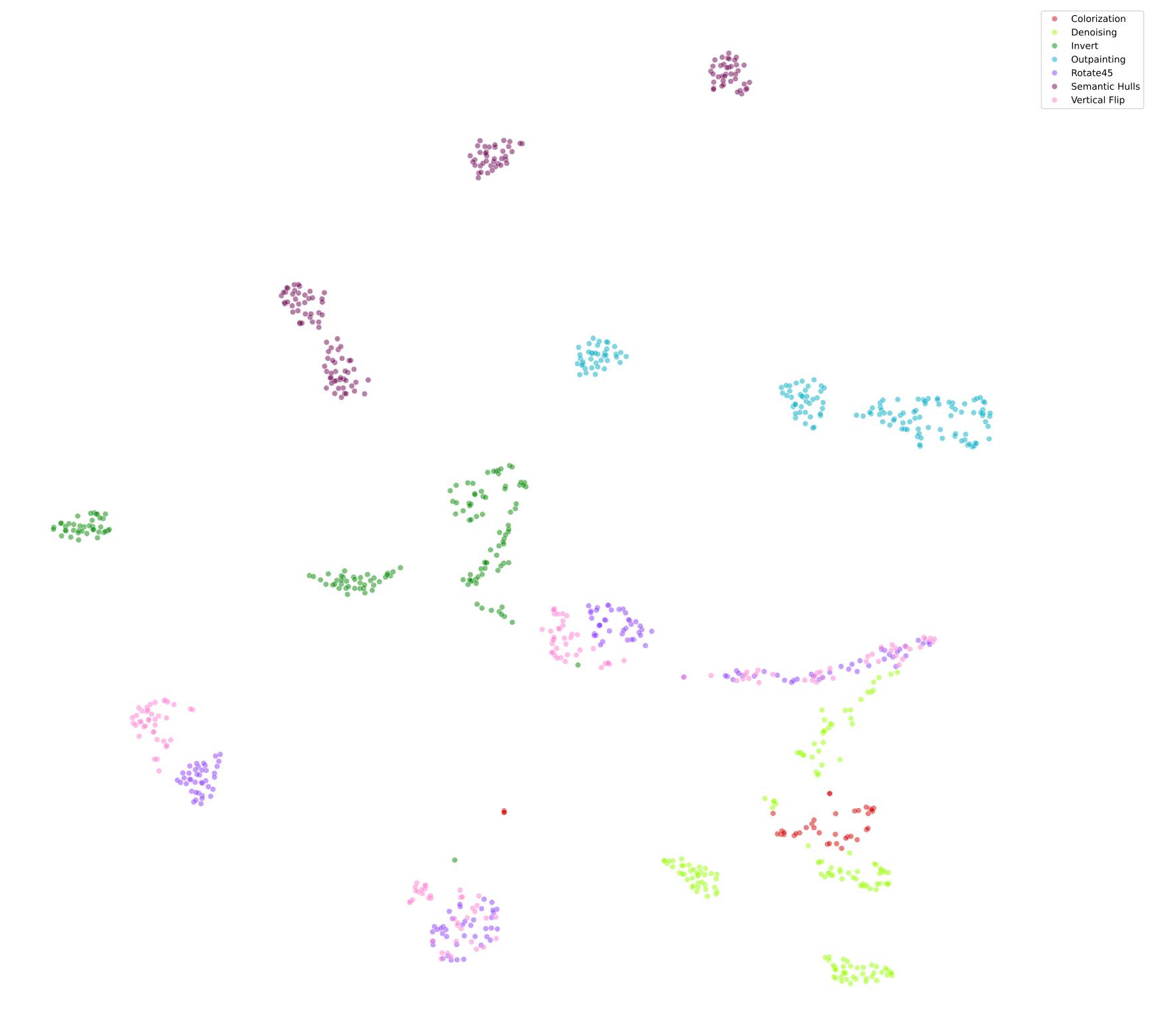}
    \includegraphics[trim={38.5cm 23cm 0 0},clip,height=0.25\textwidth]{figures/pdf_tsne_plots/SupCL/final_SupCL_sampled/unseen_tasks_unseen_datasets/tsne_classes_no_symbols_SupCL_final_SupCL_sampled_unseen_tasks_unseen_datasets_all.jpg}
    \begin{tabular}{c}
        \midrule
    \end{tabular}
    \rotatebox[origin=lb]{90}{$\phantom{---}$colored by datasets}\includegraphics[trim={0 0 9cm 0},clip,width=0.25\textwidth,height=0.25\textwidth]{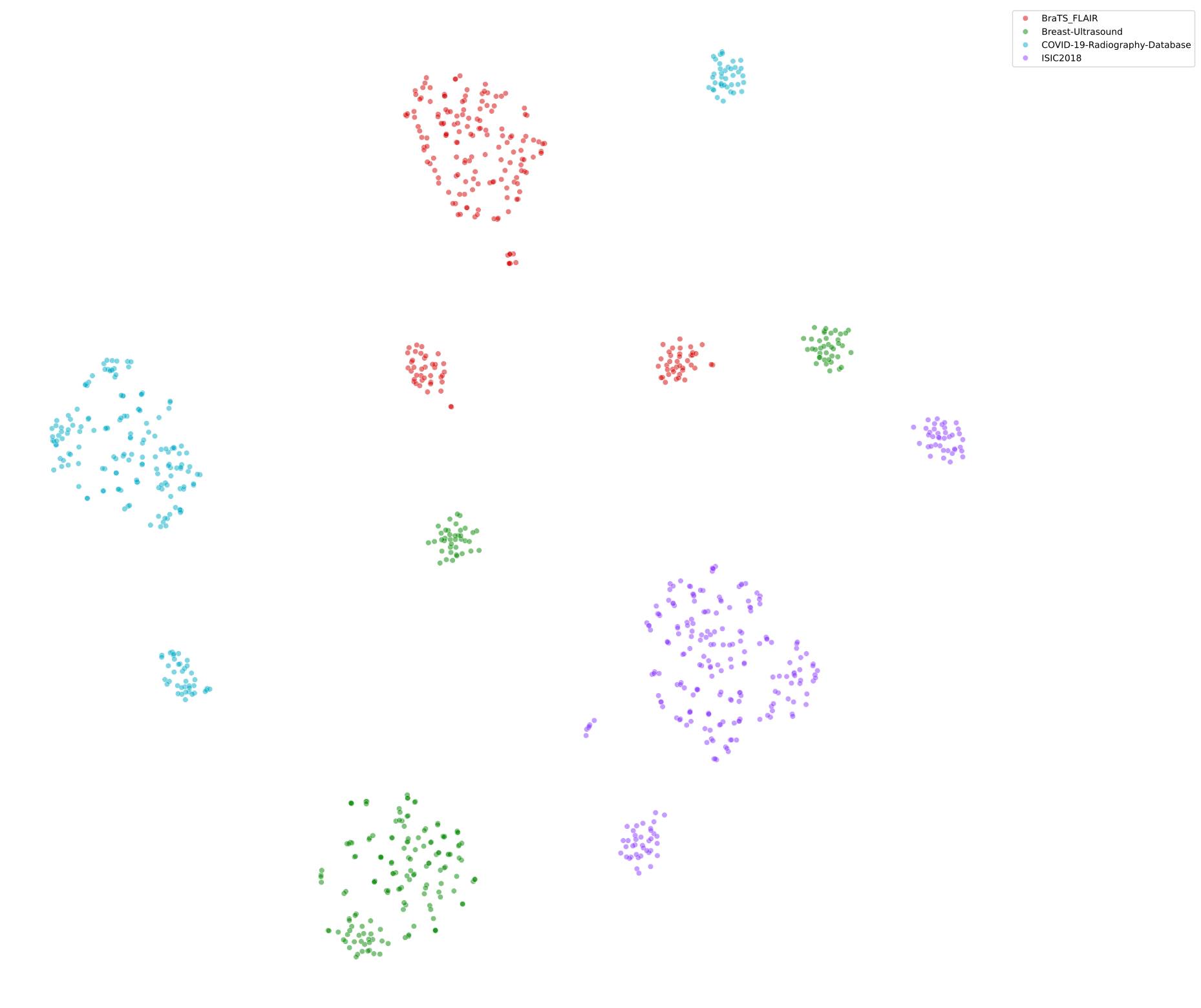}
    \includegraphics[trim={0 0 9cm 0},clip,width=0.25\textwidth,height=0.25\textwidth]{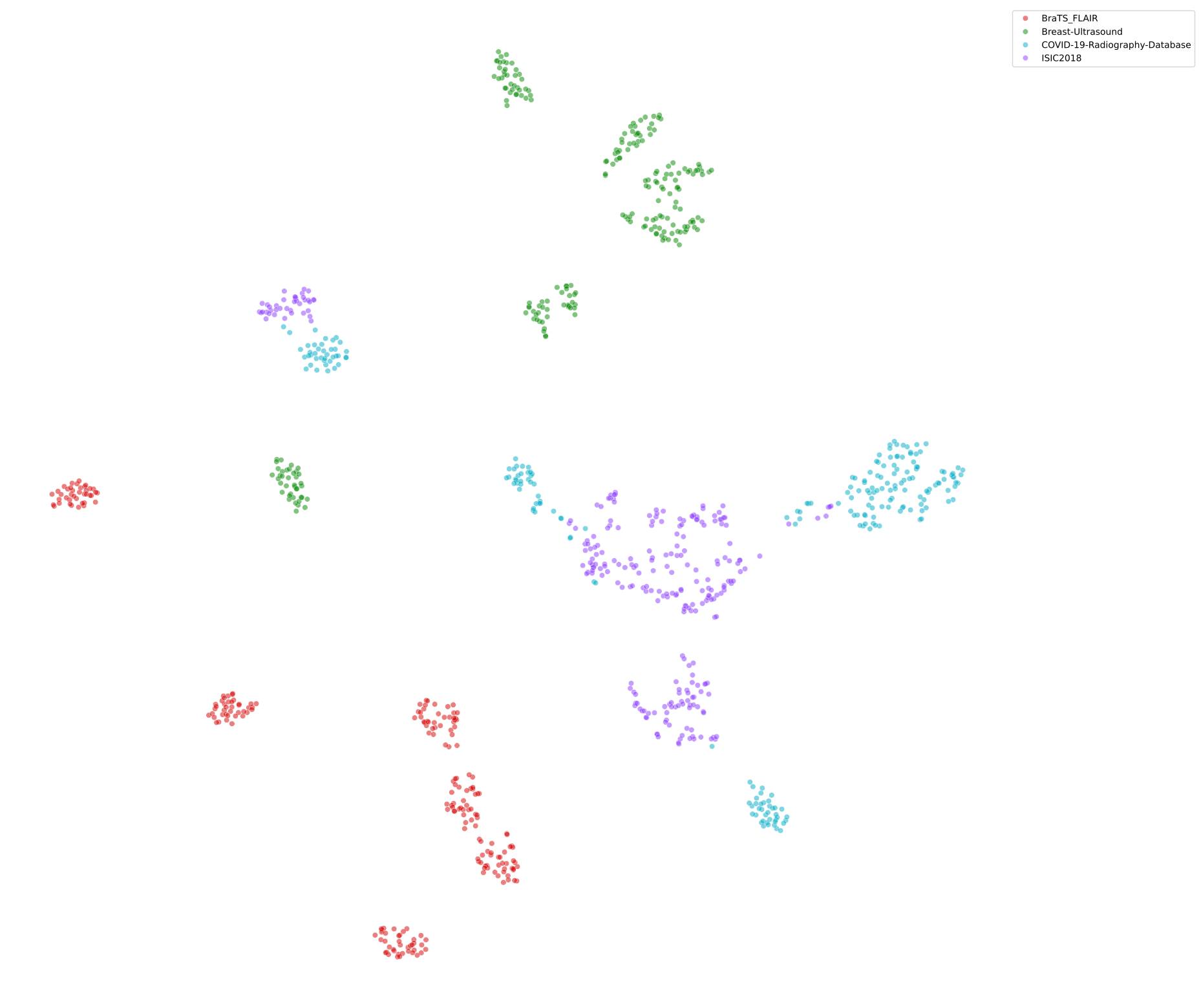}
    \includegraphics[trim={0 0 9cm 0},clip,width=0.25\textwidth,height=0.25\textwidth]{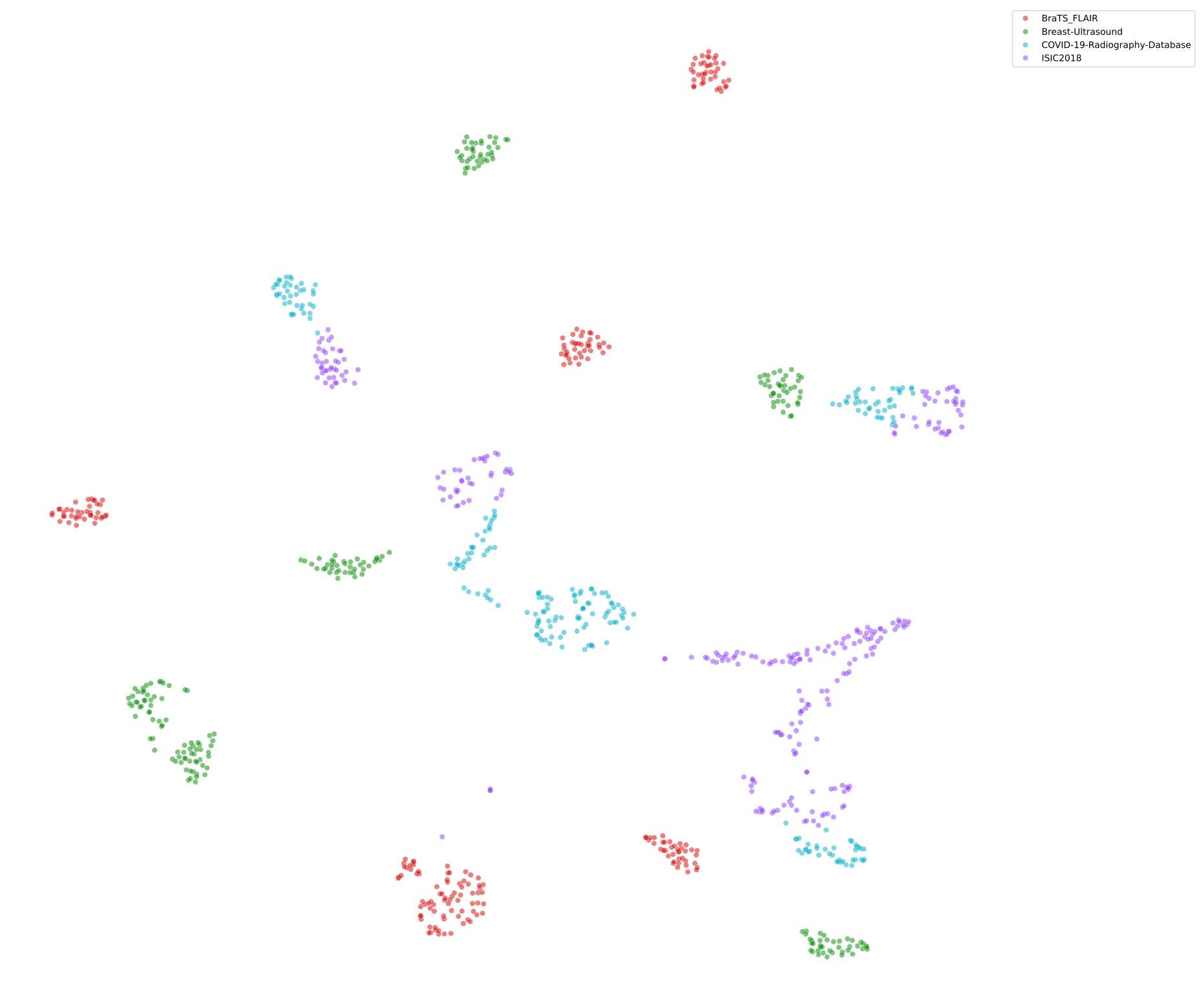}
    \includegraphics[trim={38cm 15cm 0 0},clip,height=0.25\textwidth]{figures/pdf_tsne_plots/SupCL/final_SupCL_sampled/unseen_tasks_unseen_datasets/tsne_datasets_no_symbols_SupCL_final_SupCL_sampled_unseen_tasks_unseen_datasets_all.jpg}
    \caption{Learned high dimensional representations from the \emph{unseen tasks} and \emph{unseen datasets} scenario  projected via t-SNE. We showcase the embedding space of CLIP, \emph{TaCo} and \emph{TaCo + failure tasks}. Coloring is based on tasks in the first row and datasets in the second row.}
    \label{fig:tsne_unseen_unseen_classes}
\end{figure*}


\section{Code Credits and Implementation}
\label{sec:code}

In this paper we make use of a broad set of open source code repositories which we base our implementation on.
We utilize Pytorch~\cite{NEURIPS2019_9015} as deep learning framework.
Further, we implemented our Task-Contrastive Learning strategy by utilizing the repository for Supervised Contrastive Learning~\cite{khosla2020supervised} (\url{https://github.com/HobbitLong/SupContrast}).
The data balancing implementation is based on the Imbalanced Sampler project~\footnote{\url{https://github.com/ufoym/imbalanced-dataset-sampler/}}.
We further made use of the Pandas library~\cite{mckinney2010data}, sklearn~\cite{scikit-learn} and GeodisTK~\footnote{\url{https://github.com/taigw/GeodisTK}} for computing the distance map tasks.

\end{document}